\documentclass[10pt]{article} 
\usepackage[accepted]{tmlr}
\usepackage{amssymb} 

\usepackage{amsmath,amsfonts,bm}

\def\eqref#1{equation~\ref{#1}}

\def\1{\bm{1}}

\DeclareMathAlphabet{\mathsfit}{\encodingdefault}{\sfdefault}{m}{sl}
\SetMathAlphabet{\mathsfit}{bold}{\encodingdefault}{\sfdefault}{bx}{n}

\usepackage[hidelinks]{hyperref}
\usepackage{url}
\usepackage{graphicx}
\usepackage{caption}
\usepackage{subcaption}
\usepackage{booktabs}
\usepackage{multirow}
\usepackage{amsmath}
\usepackage[most]{tcolorbox}
\usepackage{xcolor}
\usepackage{wrapfig} 
\usepackage{makecell}

\author{\name Angeliki Dimitriou \email angelikidim@ails.ece.ntua.gr \\
      \addr Artificial Intelligence and Learning Systems (AILS) laboratory\\
      National Technical University of Athens 
      \AND
      \name Nikolaos Chaidos \email nchaidos@ails.ece.ntua.gr \\
      \addr Artificial Intelligence and Learning Systems (AILS) laboratory\\
      National Technical University of Athens 
      \AND
      \name Maria Lymperaiou \email marialymp@ails.ece.ntua.gr \\
      \addr Artificial Intelligence and Learning Systems (AILS) laboratory\\
      National Technical University of Athens
      \AND
      \name Giorgos Filandrianos \email geofila@ails.ece.ntua.gr \\
      \addr Artificial Intelligence and Learning Systems (AILS) laboratory\\
      National Technical University of Athens
      \AND
      \name Giorgos Stamou \email gstam@cs.ntua.gr \\
      \addr Artificial Intelligence and Learning Systems (AILS) laboratory\\
      National Technical University of Athens}

\def\month{09}  
\def\year{2026} 
\def\openreview{\url{https://openreview.net/forum?id=DmgXSefvC0}} 

\begin{document}

\title{U-CECE: A Universal Multi-Resolution Framework for Conceptual Counterfactual Explanations}

\maketitle

\begin{abstract}%
As AI models grow more complex, explainability is essential for building trust, yet concept-based counterfactual methods still face a trade-off between expressivity and efficiency. Representing underlying concepts as atomic sets is fast but misses relational context, whereas full graph representations are more faithful but require solving the NP-hard Graph Edit Distance (GED) problem. We propose U-CECE, a unified, model-agnostic multi-resolution framework for conceptual counterfactual explanations that adapts to data regime and compute budget. U-CECE spans three levels of expressivity: atomic concepts for broad explanations, relational sets-of-sets for simple interactions, and structural graphs for full semantic structure. At the structural level, both a precision-oriented transductive mode based on supervised Graph Neural Networks (GNNs) and a scalable inductive mode based on unsupervised graph autoencoders (GAEs) are supported. Experiments on the structurally divergent CUB and Visual Genome datasets characterize the efficiency-expressivity trade-off across levels, while human surveys and LVLM-based evaluation show that, on CUB, the retrieved structural counterfactuals are frequently judged semantically equivalent to, and often preferred over, reference deterministic GED-based explanations.
\end{abstract}

\section{Introduction}
\label{sec:intro}

The rapid advancements of artificial intelligence applications have widely sparked the interest of researchers and commercial users, boasting beyond human-level successes in Computer Vision \citep{CHANG2026111934},  Multimodality \citep{Yuan2025ASO}, Natural Language Processing \citep{zheng-etal-2025-automation} and other fields \citep{gao2026dreamdojogeneralistrobotworld}. At the same time, the inscrutable nature of these black-box systems raises concerns regarding their fairness, hidden risks and rationality in decision-making \citep{fairness}. 
Building human trust in these systems requires Explainable AI (XAI) to scrutinize opaque model reasoning
\citep{ALI2023101805}.

The ever-increasing complexity of contemporary models requires transitioning from \textit{ante-hoc} to \textit{post-hoc} explainability techniques: even though inherent interpretability of neural models was deemed as a necessary pre-requisite in critical domains, promoting ante-hoc explainability as the right direction \citep{Rudin2019}, the sheer adoption and reliance on complex systems such as Large Language Models (LLMs) underscores the need for alternatives. Post-hoc XAI accepts the difficulty of interpreting decisions after the fact to maintain high performance, both in micro- and macro-levels. More specifically, mechanistic interpretability covers techniques satisfying micro-level explanations, suggesting a bottom-up approach of opening the black box \citep{bereska_mechanistic_2024}. Ultimately, it aims to reverse engineer a primarily black-box model by offering deep and causal insights through localization and tracing information across the network, setting the starting point for model editing in cases where compliance is mandatory \citep{guo2024mechanisticunlearningrobustknowledge}. Nevertheless, the need for white-box access, as required in the mechanistic case, is often far from realistic, e.g. when utilizing proprietary models. On the other side of the spectrum lies concept-based explainability, with the majority of works accepting the black-box nature of models. An approximation of the model is performed exploiting high-level conceptual cues that lie close to human perception in a top-down fashion \citep{poeta2023conceptbasedexplainableartificialintelligence}.

In the latter case, manipulation of model inputs allows the reporting of measurable changes in the output, therefore delineating relationships between input concepts and model decisions. Throughout this process, we are able to explain black-box models without requesting \textit{any} knowledge regarding their inner structure and workings, while providing explanations in human-understandable terms, alleviating the need for expertise when interpreting AI systems in general. 
Nevertheless, input-output probing should be performed in a well-defined and informative way, since the search space of input-output relationships can be large -even infinite- in the greedy case. For example, in cases of tabular input data, numerical features can be slightly altered (e.g. change a decimal digit), or receive major changes (e.g. increase or decrease the value of a feature by orders of magnitude) and everything in between. This possibility is applicable to all features and data samples, resulting in infinite combinations in practice, without any guarantee that meaningful patterns would occur from each combination. Thus, a heuristic strategy should apply some restrictions on such manipulations, instructing causal pathways between restricted input manipulations leading to observable outcomes.

The idea of minimal input changes inducing measurable output changes has been extensively studied under the term \textit{counterfactual explanations} \citep{Guidotti2024}. By probing how a model responds to interventions on its input, counterfactuals expose cause-effect relations within the model, aligned with human reasoning \citep{stepin2021survey}. For concept-based explanations, a counterfactual question would explore alternative scenarios to the observed one, by answering \textit{which concept X should be changed to X' so that a model's outcome Y changes to Y', while the transition from X to X' is minimal?}. Under this formulation, minimality admits multiple interpretations, with exact minimality rarely attainable in practice. Methods such as ours strive for minimality and quantify it rather than guarantee it. Since concepts can be directly linked to semantics, imposing minimal semantic distances forms a natural strategy for crafting \textit{counterfactual interventions}.



Despite the success of conceptual approaches in literature \citep{filandrianos2022conceptual, dervakos2023choose, dimitriou2023structure}, existing methods often force a trade-off between expressivity and computational efficiency. While simple sets of atomic concepts are lightweight, they fail to capture the relational context of a complex scene
 - such as the difference between a "helmet hanging on a bike and a person riding the bike" versus a "person wearing a helmet riding a bike". 
Conversely, explicit graph-based representations can offer high fidelity in many cases but suffer from the NP-hard complexity of requiring the solution of the Graph Edit Distance (GED) problem for accurate counterfactual retrieval.

In this paper, we propose \textit{\textbf{U-CECE}: A \textbf{U}niversal \textbf{C}ounterfactual \textbf{E}xplanations via \textbf{C}onceptual \textbf{E}dits} framework. 
Universal is used in a representational sense: a single pipeline whose downstream stages remain structurally unchanged across resolutions and across input modalities, given a symbolic concept layer.
Unlike prior disjointed methods, U-CECE provides a unified architecture that can be adaptively scaled in terms of complexity based on the specific data regime and application requirements. Our framework integrates four distinct modules into a cohesive hierarchy: 
\begin{itemize}
    \item Level 1 (\textbf{U-CECE-Atomic}): A baseline  utilizing atomic concept sets for rapid, broad explanations.
    \item Level 2 (\textbf{U-CECE-Relational}): A "sets-of-sets" approach that captures simple object interactions via "rolled-up" concepts.
    \item Level 3 (\textbf{U-CECE-Structural}): The highest fidelity tier, representing scenes as full concept graphs. This tier can adaptively fork into two operational modes: (a) \textbf{Transductive} Mode: A precision-focused path for limited quantity of data that utilizes supervised Graph Neural Networks (GNNs) to approximate GED, and (b) \textbf{Inductive Mode}: A scalable path when training data is abundant that leverages unsupervised retrieval through Graph Autoencoders (GAEs) for real-time inference.
\end{itemize}

Our current paper marks several key contributions: 
\textbf{I)}
    The introduction of a  model-agnostic framework that harmonizes 
    atomic, relational, and structural
    conceptual representations into a single unified multi-resolution pipeline,
    evaluated under one common protocol and enabling, for the first time, a controlled comparison of what each level of expressivity costs and recovers on identical data.
\textbf{II)} An adaptive retrieval strategy that trades off fidelity and scalability through transductive or inductive  retrieval, making  conceptual counterfactuals practical under different data and compute regimes, 
and providing the first systematic evaluation of unsupervised inductive retrieval for this task
\textbf{III)} Strong empirical evidence that approximate conceptual counterfactuals preserve semantic faithfulness: 
human and Large Vision-Language Model (LVLM)-based studies show that retrieved counterfactuals are semantically aligned with, and often preferred over, reference GED-based counterfactuals, on the CUB dataset.
U-CECE is proposed as an adaptive framework, not a state-of-the-art retriever and we accordingly evaluate its internal consistency across resolutions, the minimality and cost of its edits, its computational profile, and its alignment with human judgment with experiments covering two visual datasets with curated conceptual annotations and two classifiers.
\section{Background on Counterfactual Explanations}
\label{sec:counterfactuals}

The journey toward transparent AI decision-making has addressed major challenges over the years, answering the question of how a model arrives at a particular response based on input evidence. The rather static nature of a \textit{"How does  instance X influence model decision Y"} query, targeting the invocation of statistical associations in data, can be revised to allow for dynamic monitoring of input-output relationships, providing a deeper understanding of cause-effect trajectories in model behavior.
A \textit{what if} reframing indeed addresses how an intervention $X \rightarrow X'$ on the input data would stimulate a modification in Y, or reversely, how we could force an alternative outcome $Y'$ to happen given $X$. By adjoining a vast amount of such interventions, we are able to form an "imaginary" reality within which retrospective queries such as \textit{"would Y remain Y if X' had happened instead of X?"} or \textit{"would some Y' have occurred if X' had happened instead of X?"} are compared to the existing reality, where Y appears if X happens. This comparison ultimately constitutes counterfactual explanations, perfectly aligning with black-box systems due to their observational nature \citep{book-of-why}.
Formally, considering a model M for which $M(X)=Y$, a counterfactual explanation comprises an instance X' for which $M(X')=Y'\neq Y$, while the distance $d(X, X')$ is ideally minimal.

\paragraph{Low-level Methods.} Counterfactual explanations have already populated the spectrum of explainability literature.
With a focus on visual classifiers, which constitute our experimental basis, there have been several endeavors operating in low-level pixel spaces: edits on specific areas of the image are proposed in order to alter a classifier's predictions. 
Generating counterfactual regions on images requires respecting the surrounding areas in order to contribute to plausible counterfactual instances \citep{Chang2018ExplainingIC}, Similarly, \cite{pmlr-v97-goyal19a} propose minimal alterations of image regions so that an image is classified as an alternative distractor class, defining minimality as the number of regions modified. 
Getting closer to the semantic edits case, minimizing the loss between an initial image and an image with manipulated attributes can instruct counterfactual generation with advanced actionability \citep{Liu2019GenerativeCI}. With the advent of more potent generative models, such as diffusion models \citep{diffusion}, visual quality and feature faithfulness of generated counterfactuals reach new levels, while the guidance of synthesizing edits becomes more accurate, thus approaching conceptual desiderata \citep{diff-counterfactuals, pmlr-v177-sanchez22a, dime, Deja2023LearningDR,Farid2023LatentDC, Weng2023FastDC, ace, Sobieski2024RethinkingVC, Komanduri2024CausalDA, Motzkus2024CoLaDCEC}.

\paragraph{Concept-Based Methods.} Moving beyond pixel-level manipulations, recent literature has shifted toward conceptual counterfactuals, where perturbations are defined in a high-level, human-understandable feature space. Unlike pixel edits, which can lead to biased or cryptic outputs, conceptual edits modify symbolic attributes while maintaining semantic coherence.
While approaches such as Concept Activation Vectors (CAVs)~\citep{kim2018interpretability,abid2022meaningfully}, clustering of embeddings~\citep{ghorbani2019towards}, or integrating explicit intermediate concepts through concept bottleneck models (CBMs)\citep{koh2020concept} provide ways to define concepts in neural latent spaces, these methods are limited when it comes to structured reasoning about hierarchical and relational knowledge. 
Most concept-based methods treat concepts as isolated variables rather than interconnected entities. The Conceptual Edits as Counterfactual Explanations (CECE) framework \citep{filandrianos2022conceptual} addressed this by treating an image as a set of atomic concepts and defining distance via a hierarchical taxonomy. This work was later extended by the Semantic Counterfactuals (SC) approach \citep{dervakos2023choose}, which introduced rolled-up concepts to capture relationships within a set-of-sets structure. The same semantic framework paired with a generation module was able to identify semantic gaps between humans and neural classifiers, establishing conceptual counterfactuals as a means of faithfully interpreting opaque models for classification \citep{spanos2025vcece}.
However, when used as stand-alone modules, these advancements often under-represent the structural complexity of real-world scenes, as they cannot fully model the intricate dependencies between objects.

\paragraph{Graph Retrieval for Counterfactuals.} To achieve full data expressivity while preserving conceptual desiderata, counterfactual discovery has been reframed as a structural retrieval problem using scene graphs. The Scene Graph Counterfactual Explanations (SGCE) framework \citep{dimitriou2023structure} represents a significant departure from set-based conceptual models. In SGCE, images are represented as scene graphs where nodes represent objects and edges represent their pairwise relationships. This allows for more descriptive, accurate, and human-aligned conceptual explanations by capturing the significance of semantic edges in the presence of intricate relationships. The utilization of scene graphs necessitates the use of Graph Edit Distance (GED), which quantifies the minimum cost of transforming one conceptual graph into another via node and edge edits. However, because the exact computation of GED is an NP-hard problem, its application in black-box conceptual interpretability requires efficient approximation techniques. The challenge lies in bypassing the exhaustive search of graph similarity for all input pairs, which is an integral but computationally expensive part of the counterfactual computation process. 
The acceleration of GED has been pursued in prior literature via efficient approximations, such as reductions to linear assignment problems \citep{jonker1988shortest, riesen2009approximate, fankhauser2011speeding} or redefinitions of the edit costs \citep{serratosa2014fast}, as well as graph kernels \citep{ged-kernels}. 
More recent works leverage Graph Neural Networks (GNNs) to obtain lower-dimensional representations, employing graph matching networks for structural similarity, and neural frameworks for learning complex graph distance functions through local or global contexts \citep{bai2019unsupervised, li2019graph, ranjan2022greed, cheng2025computing}. 
With graph embeddings at hand, counterfactual graph matching becomes equivalent to finding the closest graph embedding pairs, as showcased in SGCE, which adopted the siamese GNN component from \cite{bai2019unsupervised}. 
Another branch in recent literature delves specifically into scene graph similarity, which is highly relevant for this study. Supervised methods, such as  IRSGS \citep{yoon2021image} or Hi-SIGIR \citep{wang2023hi}, typically rely on similarity labels derived from image captions or sentence embeddings to train GNNs. However, these are often limited by the high cost of label generation and the inherent inconsistency of caption-based similarities. In contrast, unsupervised techniques based on  Graph Autoencoders (GAEs) eliminate the need for labeled training data, while significantly accelerating the training burden \citep{graph-edits}, even though default GAE baselines fall behind their supervised counterparts in terms of retrieval performance. In a similar research path, the SCENIR framework \citep{chaidos2025scenirvisualsemanticclarity} introduces a competitive GAE-driven architecture that achieves state-of-the-art concept-aware image retrieval, while advocating for GED as a deterministic and robust reference measure for evaluation. 
By integrating different parts of these diverse strategies, the U-CECE framework provides high structural expressivity while maintaining the flexibility to choose between high-precision supervised matching or scalable inductive retrieval.

\section{The U-CECE Framework: Multi-Resolution Semantics}
\label{sec:framework}

\begin{figure}[t!]
    \centering
    \includegraphics[width=0.78\linewidth]{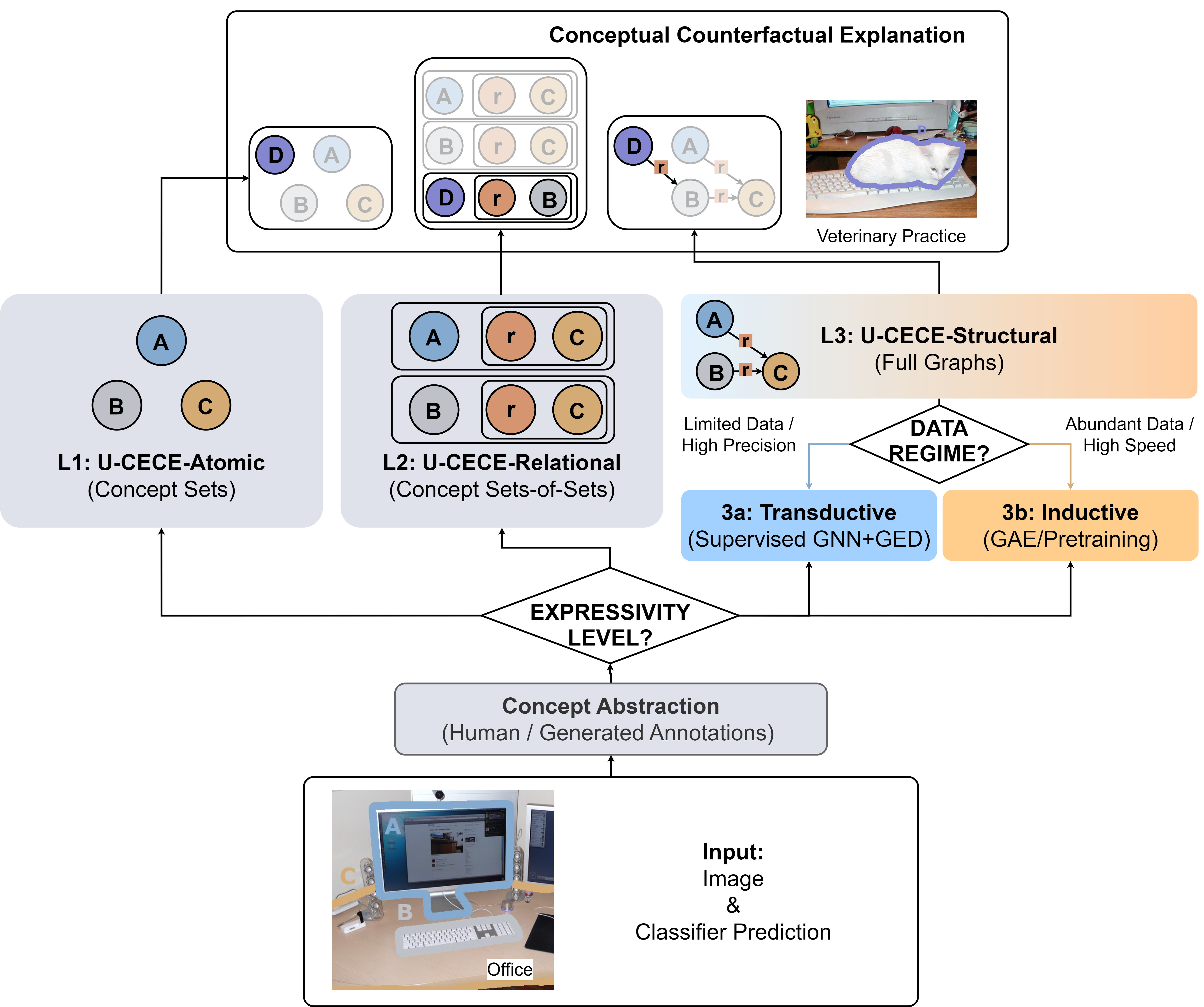}
    \caption{
    \textbf{The U-CECE Framework}  for Multi-Resolution Conceptual Counterfactuals. The pipeline begins with Concept Abstraction, mapping raw inputs to a symbolic vocabulary ($\mathsf{Screen (A)}$, $\mathsf{Keyboard (B)}$, $\mathsf{Table (C)}$, $\mathsf{Cat (D)}$) and their relations ($\mathsf{on (r)}$). Segmentation visualizations are for clarity, not to imply a specific abstraction pipeline. The central Expressivity Level decision directs queries through three increasingly complex representations: Atomic Sets (L1), Sets-of-Sets (L2), or Full Scene Graphs (L3). In the high-fidelity L3 path, U-CECE forks into different engines based on the Data Regime: Supervised Transductive GNNs (3a) for high-precision matching on limited data, or Unsupervised Inductive GNNs (3b) for scalable retrieval on abundant data. All paths converge to provide a Conceptual Counterfactual Explanation - such as flipping an "Office" prediction to "Veterinary Practice" by introducing a $\mathsf{cat}$ $\mathsf{on}$ a $\mathsf{keyboard}$.
    }
    \label{fig:framework-comparison}
\end{figure}

In the following sections, we introduce U-CECE, a unified framework designed to provide semantic-aware counterfactual explanations across varying levels of relational expressivity. This multi-resolution approach allows the framework to adapt to the underlying complexity of data samples and the specific requirements of the downstream application. The architectural pipeline, illustrated in Figure \ref{fig:framework-comparison}, commences with the essential step of \textit{concept abstraction}, which converts raw inputs into human-understandable terms. We then define the formal symbolic hierarchies and taxonomies used to quantify semantic distances. Finally, we detail the three distinct levels of data expressivity - Atomic, Relational, and Structural - and the specialized U-CECE components developed to navigate the unique computational and semantic challenges of each.

\subsection {Concept Abstraction}

By design, U-CECE operates as a fundamentally modality-agnostic framework. 
It initiates by mapping raw, domain-specific inputs, whether they be pixels, audio waveforms, or text tokens, into a unified, high-level symbolic space required for conceptual reasoning. For instance, considering the visual desk setup in the Input Image (Figure \ref{fig:framework-comparison}), raw pixel features are abstracted into a distinct conceptual vocabulary, yielding symbolic terms such as $\mathsf{screen (A)}$, $\mathsf{keyboard (B)}$, and $\mathsf{table (C)}$. Crucially, this process also captures the structural relationships that define the scene, such as $\mathsf{screen\,\text{--}on\text{--}table}$ or $\mathsf{keyboard\,\text{--}next to\text{--}computer}$. 
As illustrated at the base of our architecture, this abstraction layer functions as a common interface. By translating modalities into a standardized, interconnected graph of concepts, the core downstream steps of the framework (which enable multi-resolution expressivity and counterfactual retrieval) remain completely structurally unchanged, regardless of the original input domain.

To populate the conceptual layer of U-CECE, semantic metadata can be sourced through either existing annotated datasets or automated extraction pipelines 
regardless of modality.
Focusing our analysis on visual data,
in many research settings, such as our own, concepts are already available through human-curated datasets where annotations, captions, or metadata explicitly link pixels with semantics. Notable examples include Visual Genome \citep{krishna2017visual}, which provides dense scene graph annotations encompassing objects, attributes, and relationships, or COCO \citep{lin2014microsoft}, which offers captions and annotated bounding boxes. 
Alternatively, in the absence of manual ground-truth labels, a common approach in the visual domain is Scene Graph Generation (SGG), where an image is parsed automatically into objects, attributes, and pairwise relations. 
In this setting, detectors identify objects (e.g., $\mathsf{dog}$, $\mathsf{person}$, $\mathsf{bike}$), while relational models capture interactions (e.g., $\mathsf{person\text{--}rides\text{--}bike}$, $\mathsf{dog\text{--}next\text{ }to\text{--}person}$). Recent SGG research has advanced along several directions: improving object and relation detection with transformer-based architectures, mitigating dataset biases through debiasing and causal inference techniques \citep{im2024egtr, tang2020unbiased, zellers2018neural}, and enhancing generalization with open-vocabulary or weakly supervised methods \citep{lu2016visual, zareian2021open, dong2022stacked}. 

Careful consideration of granularity, domain specificity, and the intended audience ensures that these concepts serve as a faithful and interpretable basis for counterfactual reasoning. This is the essence of “choosing your data wisely” \citep{dervakos2023choose}: the selected concept vocabulary directly shapes both the fidelity and the usefulness of explanations. Building on this, once these concepts are curated and extracted, organizing them into expressive structures ranging from atomic sets to full graphs is essential for enabling the effective multi-resolution explanations provided by the U-CECE framework.

\subsection{Representation Hierarchy: From Atoms to Graphs}

The U-CECE framework, being a conceptual explainability approach, aims to shift the focus from low-level, often unintuitive, features to high-level, human-understandable semantic concepts. 
While deep learning architectures typically rely on pixel-level correlational patterns (e.g., pixel brightness, color), we follow the foundational principle of Conceptual Edits as Counterfactual Explanations (CECE) \citep{filandrianos2022conceptual}: Instead of identifying pixels to be perturbed, we identify the minimal set of semantic edits (e.g., "adding a $\mathsf{helmet}$" or "removing a $\mathsf{laptop}$" instead of "decreasing the pixel intensity in this region") required to flip a model's decision.
This way, U-CECE enables structured reasoning across multiple dimensions, harnessing hierarchical taxonomies, and extending them to relational structures for higher expressivity, to provide an effective pipeline for navigating the trade-offs between symbolic abstraction and computational efficiency. Ultimately, U-CECE ensures that the minimal counterfactual edit is found not in a raw feature space or a disjointed set, but within a semantically coherent and organizationally sound knowledge representation.

To enable multi-resolution counterfactuals, we must first formalize how concepts are structured and related within our framework. U-CECE treats the representation of a data sample not as a static label, but as a symbolic entity that can scale in complexity from a simple category to a dense relational graph.

\subsubsection{Symbolic Atoms and Taxonomies}
At the most fundamental level, we define the building blocks of our semantic space. Formally, a \emph{concept} $C$ within our framework is defined as any of the following:
\begin{itemize}
    \item An \textbf{atomic concept} $A$ (e.g., $\mathsf{cat}$, $\mathsf{laptop}$), representing a basic category.
    \item The \textbf{top concept} $\top$, representing the universal domain.
    \item The \textbf{bottom concept} $\perp$, representing the empty or impossible category.
\end{itemize}

These concepts are organized into \emph{taxonomies}, i.e. formal hierarchies termed \textbf{TBoxes}. For example, the axiom $\mathsf{cat} \sqsubseteq \mathsf{animal}$ indicates that ``every $\mathsf{cat}$ is an $\mathsf{animal}$.'' Visualized as a directed acyclic graph, these hierarchies allow for automated semantic reasoning: recognizing a $\mathsf{cat}$ inherently implies the presence of more generic concepts like $\mathsf{LivingThing}$ through transitivity. This hierarchical grounding is essential for Level 1 (Atomic) counterfactuals, where the distance of an edit is determined by the path through this taxonomy.

\subsubsection{Relational Enriched Concepts (Roles)}
Beyond simple inclusion, concepts participate in directed relationships, formally termed \emph{roles} (e.g., $\mathsf{person \text{--} wears \text{--} helmet}$). To maintain computational tractability while increasing expressivity, we adopt the "rolling up" strategy of the SC framework \citep{dervakos2023choose}. By incorporating outgoing relationships directly into node labels, we produce enriched concepts of the form $\exists r.C$ (e.g. $\exists \mathsf{wears.helmet}$), interpreted as ``something that wears a $\mathsf{helmet}$.'' This  allows a \textbf{set-of-sets} representation, effectively preserving the relational structure within a symbolic format, bridging the gap between atomic sets and full graphs.

\subsubsection{Structural Scene Graphs}
The final step in our formalization incorporates full interactions through \textbf{graph-structured representations}. 
While sets and "rolled-up" concepts are computationally efficient, they can obscure critical distinctions in complex scenes. 
Consider the visual domain, where a scene might contain concepts such as $\mathsf{man}$, $\mathsf{on.bike}$, and $\mathsf{wearing.helmet}$. Representing these only as an unordered label set obscures important distinctions - for example, whether there are multiple riders, and if the helmets belong to the men on bikes or to other men standing nearby. Such ambiguities are not trivial; they define the difference between safe and unsafe scenarios in high-stakes domains.
To resolve this, we leverage the graph-based formalism established by the SGCE framework \citep{dimitriou2023structure}. Each data sample (image $I$) is represented as a  concept graph $G = (V, E)$, where nodes $V$ correspond to objects (with associated feature vectors $X$) and edges $E$ capture their explicit pairwise relationships, typically encoded in an adjacency matrix $A$. In this structural view, the underlying hierarchical semantics of the TBox remain intact, but they are now embedded within a topology that is both richer and more faithful to the underlying data. This structural fidelity forms the basis for our Level 3 expressivity tier, where counterfactuals are discovered through graph matching.

\subsection{Expressivity Levels}
\label{sec:expressivity}

The U-CECE framework is built on the principle that not all data samples or application domains require the same level of relational detail. To this end, we formalize three tiers of expressivity, each defining "minimality" and "distance" according to the complexity of the underlying symbolic representation.

\subsubsection{Level 1 (Atomic)}

\label{sec:concept-distance}

At the most fundamental resolution, U-CECE-Atomic leverages the principles of the CECE framework, representing an image as a set of atomic concepts. The "minimality" of a counterfactual is governed by the \emph{conceptual distance} ($d_c$) within a structured semantic space.
Rather than treating concepts as isolated labels, they are embedded in a hierarchical structure, denoted $TBox(V_T, E_T)$. Nodes $v_T$ correspond to concepts (e.g., $\mathsf{cat}$, $\mathsf{mammal}$), while edges $E_T$ encode semantic inclusion  relationships (e.g., every $\mathsf{cat}$ is also a $\mathsf{mammal}$). 
The distance $d_c(A,B)$ between two concepts $A, B$  is defined as the length of the shortest path between them:
\begin{equation}
\label{eq:concept-distance}
    d_c(A,B) = \min_{p \in P(A,B)} \sum_{(u,v) \in p} w(u,v)
\end{equation}
where \( P(A,B) \) is the set of all paths connecting \( A \) and \( B \) in the \( TBox \) and $w(u,v)$ is the weight or cost of the transition between concepts. Typically, this cost defaults to 1 but can be weighted to reflect domain-specific actionability. 
From a computational perspective, \( P(A,B) \) are  computed using Breadth First Search (BFS), unless weighted edges are introduced, in which cases the  Dijkstra algorithm is executed instead. If no path exists (disconnected components), \( d_c(A,B) = \infty \), indicating that the corresponding edit is infeasible. 


A graph-based notion of distance captures semantic relatedness (concepts lying closer in the hierarchy are cheaper to substitute, while also being semantically closer), and provides a principled way to assign costs to \textit{edit operations} for counterfactuals. In our paradigm, the following three edit operations can be implemented:
\begin{itemize}
    \item \textbf{Replacement (\textit{R}):} Substituting concept $A$ with $B$ has a cost equal to their distance $d_c(A,B)$.
    \item \textbf{Deletion (\textit{D}):} Removing concept $A$ is treated as replacing it with a generic placeholder, with cost proportional to its distance from the top-level category $\top$ ($d_c(A,\top)$).
    \item \textbf{Insertion (\textit{I}):} Introducing a new concept $B$ is modeled symmetrically, as the distance from the generic placeholder $\perp$ to $B$ ($d_c(\perp,B)$).
\end{itemize}


As illustrated in Figure \ref{fig:framework-comparison}, this deterministic approach allows  U-CECE-Atomic  to retrieve counterfactuals by treating the abstracted concepts as an unordered set. For our office scene example (initially containing $A$, $B$, and $C$), the L1 path identifies the minimal edit necessary to flip the classification to "Veterinary Practice". While for illustrative simplicity the optimal counterfactual image remains the same across all expressivity levels in this instance, from an explanatory standpoint, the mere addition of concept $D$ ($\mathsf{cat}$) to the existing set suffices at this resolution. Overall, the L1 tier provides a high-speed, transparent explanation, identifying $D$ as the key semantic component required to change the model's prediction in the illustrated case.


\subsubsection{Level 2 (Relational)}
\label{sec:relational_distance}


Moving beyond isolated atoms, Level 2 (U-CECE-Relational) adopts the SC approach to capture simple object interactions. In this tier, images are treated as \emph{exemplars} $\mathcal{E}$ - formulated as collections of  "rolled-up" concepts (i.e., source objects paired with their relational targets). Because this algorithm builds its understanding of the scene exclusively through valid edge connections, any isolated atomic concepts (objects without structural relationships) are excluded from the exemplar. Since each exemplar is essentially a set of concepts, distances are computed hierarchically generalizing the atomic metric to a \emph{set-edit distance}.  

\paragraph{Set-level edit distance.}
Given two finite sets $A = \{a_1,\dots,a_m\}$ and $B = \{b_1,\dots,b_n\}$,
the set-edit distance is computed via a minimum-cost bipartite matching:

\begin{equation}
    d_{\text{set}}(A,B) \;=\; \min_{M \in \mathcal{M}(A,B)} \;
\sum_{(a,b)\in M} d_r(a,b)
\end{equation}
where: $\mathcal{M}(A,B)$ is the set of all full matchings between $A$ and $B$ (allowing dummy matches with a fixed penalty $\delta$ if $|A| \neq |B|$). 
The distance $d_r(\cdot,\cdot)$ evaluates the transformation of the entire relational unit.
While for atomic concepts $C, D$, their distance is given by the metric
$d_c(C,D)$ defined in Section \ref{sec:concept-distance},
for rolled-up concepts of the form $\exists r.C$ and $\exists s.D$, we re-define it to $d_r(\exists r.C, \exists s.D)$ and treat them
as two-element sets $\{r, C\}$ and $\{s, D\}$. Their distance is then computed as a set-edit
distance between these trivial sets:
\begin{equation}
    d_r(\exists r.C, \exists s.D) \;=\; d_{\text{set}}(\{r, C\}, \{s, D\})
\end{equation}
In the paradigm discussed within this paper, exemplars are essentially images within a dataset modeled as a "bag of relational facts". The distance between two images $\mathcal{I}_1, \mathcal{I}_2$ is then:
\begin{equation}
    d_{\mathcal{E}}(\mathcal{I}_1, \mathcal{I}_2) = d_{\text{set}}(\mathcal{I}_1, \mathcal{I}_2)
\end{equation}
where unmatched elements incur a penalty $\delta$. Level 2 provides a crucial middle ground: it preserves the semantics of interactions (the "on-ness" or "near-ness") without the computational complexity of full graphs.

As shown in the middle path of Figure \ref{fig:framework-comparison}, U-CECE-Relational moves beyond the mere presence of concepts to explain their structural context. For our example, the L2 path explicitly identifies the relational transition. The resulting explanation highlights the addition of the rolled-up concept $\exists \mathsf{on}.\mathsf{keyboard}$ $(\mathsf{r}.\mathsf{B})$ which refers to the added $\mathsf{cat}$ concept.  
While this level provides a significantly more nuanced explanation than a simple label, it serves as a semantic hint, establishing that the relationship exists within the scene's set of features.
However, this strict edge-centric focus means the metric inherently penalizes structural shifts while remaining blind to unconnected background objects. For example, if an isolated but highly relevant object (e.g. $\mathsf{doctor}$) were present in the scene without any explicit edges linking it to other entities, the relational metric would entirely ignore its presence. Consequently, while L2 successfully captures relational state changes, it leaves the explicit, point-to-point topological binding to the higher-resolution structural tier.



\subsubsection{Level 3 (Structural)}
\label{sec:structural_distance}
At the highest resolution, U-CECE-Structural represents scenes as full conceptually-rich graphs $G=(V, E)$. This tier captures the complete topology of the data, ensuring that counterfactuals are sensitive to the specific relationships between multiple instances of the same concept, effectively addressing L2 restrictions.

Finding the minimal counterfactual here equates to solving the \textbf{Graph Edit Distance (GED)} problem \citep{sanfeliu1983distance}. 
Graph matching stands as a natural extension of placing concepts on a bipartite graph and then calculating the perfect matching; instead of finding the \textit{sets of concepts} to be added, deleted or replaced, we need to adapt and extend this formulation by discovering \textit{matchings between graphs}. In that case, edits correspond to finding the closest subgraphs between a query graph and its candidates and then defining what needs to be changed in the form of neighbor-informed \textit{concept-role-concept} triplets.

Specifically, by assuming a query image $I_q$ and a set of candidates $I^i_c, i=1, 2,..., N$, their corresponding scene graphs are denoted as $G_q$ and $G^i_c, i=1, 2,..., N$ respectively. If $I_q$ belongs to a class $A$, a counterfactual image should be any $I^i_c \neq I_q$ that belongs to some class $B^i \neq A$. This translates in finding the closest scene graph $G^i_c \in B^i$ to the query $G_q \in A$, which culminates in the GED problem. By considering a set of graph edits $\{e^i_1, e^i_2, ... e^i_m\}\in \{R, D, I\}$ of non-negative costs $\{c_{e^i_1}, c_{e^i_2}, ... c_{e^i_m}\}$ needed to transit from $G_q$ to some other counterfactual scene graph $G^i_c$, GED is mathematically formulated as:
\begin{equation}
GED(G_q, G^i_c)=\min_{e_1, e_2, ... e_m} \sum_{j=1}^m c_{e^i_j} \quad \text{such that } G_q \in A, G^i_c \in B^i, A \neq B^i
\end{equation}
GED is an NP-hard problem which given $N-1$ graph candidates, needs to be computed $N-1$ times between $G_q$ and all $G^i_c$ in order to conclude to the counterfactual graph $G^k_c \in B^k$, posing an even greater overhead. 
The calculation of $G^k_c$ is mathematically termed in the following equation:
\begin{equation}
   G^k_c = \arg \min_{G^i_c \in B^i, i=1, 2, ..., N-1}GED(G_q, G^i_c)
\end{equation}
Due to its deterministic characteristics, GED proves well-suited for counterfactual graph retrieval, though computational optimizations are necessary to improve its efficiency. U-CECE employs two distinct retrieval modes: a high-precision, computationally expensive path and a less precise path that operates inductively. 
Both engines utilize an identical retrieval pipeline, differing only in the methodology for computing graph embeddings and the underlying training regime. The matching procedure of the embeddings between $G_q$ and all $G^i_c$ during inference yields the closest graph $G^k_c \in B^k$ to the query $G_q$. Then GED will have to be calculated only \textit{once}. 
After that, $G^k_c$ comprises the counterfactual graph of $G_q$ and $GED(G_q, G^k_c)$ provides a measure of conceptual distance 
to the retrieved instance (not a newly generated image like in  \cite{spanos2025vcece}). 
Consequently, the \textit{R, D, I} graph edits $\{e_1, e_2, ... e_m\}$  required to transform $G_q\rightarrow G^k_c$ comprise a \textit{local} counterfactual explanation for these instances, describing \textit{what needs to be altered} in $G_q$ to become indistinguishable from $G^k_c$, which is equivalent to outlining the conceptual cues 
- such as a specific relationship between a $\mathsf{cat}$ and a $\mathsf{keyboard}$ - 
that discriminate the closest instances from the two classes. 
This process should be repeated for all the $N$ scene graphs in our test subset, so that we obtain $N$ counterfactual scene graph pairs, by accepting $N$ \textit{GED} operations in total. 

As shown in the final path of Figure \ref{fig:framework-comparison}, U-CECE-Structural provides the most complete explanation. While L1 identifies the object ($\mathsf{cat}$) and L2 identifies the interaction ($\mathsf{on-keyboard}$), only the L3 tier explicitly binds these elements together into a cohesive graph, resolving any ambiguity found in the lower tiers. Furthermore, by formulating the explanation as a subgraph, this approach captures unconnected objects, successfully incorporating any isolated nodes that were invisible to the relational metric. This ensures that the explanation reflects not just a general state of the room, but a precise structural configuration: it is the $\mathsf{cat}$ itself that is on the $\mathsf{keyboard}$. From a human perspective, this yields a more intuitive counterfactual. While a $\mathsf{cat}$ simply sitting in the corner of a room might not be enough to override the ``Office'' classification, a $\mathsf{cat}$ specifically on the $\mathsf{keyboard}$ serves as a strong semantic cue that the space is not a functional workplace, shifting the context to one that prioritizes the presence of animals over professional tasks.

All in all, the three tiers are unified by their distance functions belonging to a single edit algebra. The atomic distance $d_c$ is a minimum-cost matching over concept sets, i.e. GED restricted to node operations on edge-free graphs, while the relational distance $d_r$ applies the same matching to rolled-up units. 
GED is therefore not a criterion native to L3, but the generalisation that contains the atomic metric as a strict special case and the relational metric as a decomposition, making a single reference meaningful across expressivities.

\section{U-CECE-Structural Retrieval Modes}
\label{sec:structural}


While GED provides a mathematically rigorous foundation for counterfactual matching, its computational complexity necessitates neural approximations to remain viable for large-scale applications \citep{dimitriou2023structure}. To this end, we introduce the two approximate scene graph retrieval engines that comprise our L3 mode.
Depending on the availability of reference labels and the required inference speed, U-CECE selects between a supervised precision-focused engine or an unsupervised efficiency-focused one. This modularity ensures that U-CECE remains applicable across diverse data regimes -whether utilizing abstracted concepts that stem from human-curated annotations or are automatically generated.

\subsection{Transductive Mode: The Precision Path}
To implement this structural tier focused on precision, we utilize a transductive, supervised engine (U-CECE-Transductive) that leverages the deterministic strengths of GED. Here, transductive refers to a retrieval setting in which the model is trained with respect to a fixed candidate graph set from the target dataset, rather than being optimized to generalize to unseen graph collections. This mode is designed for scenarios where the "minimality" of an explanation is essential, ensuring that the retrieved counterfactual is the closest possible semantic neighbor within the available data.
The core philosophy of the transductive engine is to "learn" the GED metric directly. By training a model specifically on the dataset of interest, we can accelerate the counterfactual search between each $G_q$ to all other $G^i_c$ during inference.


\paragraph{Supervised GED Approximation.} The training pipeline considers a GNN ($f_{\theta}$) tasked to map given semantic graphs $G$ into an embedding space $h_G=f_{\theta}(G)\in \mathbb{R}^d$, such that geometric proximity in the embedding space reflects semantic similarity as quantified via GED.
In technical terms, upon constructing the supervision pairs comprising randomly sampled graphs ($G^{i_1}, G^{i_2}$), a Siamese GNN \citep{krivosheev2020siamesegraphneuralnetworks}, comprising two identical node embedding units, is trained to produce a fixed-dimensional vector $h_G\in \mathbb{R}^d$ after performing $L$ rounds of message passing. Initial node features $ \mathbf{h}_v^{(0)} = \textbf{x}_v$ are extracted using pre-trained Glove \citep{glove} embeddings. Considering \( \mathcal{N}(v) \) as the neighborhood of node \( v \), \( \mathbf{W}^{(\ell)} \) as learnable weight matrices, and \( \sigma \) as a nonlinear activation function (e.g., ReLU), for each GNN layer $l=1, ..., L$, each node embedding $\textbf{h}^l_v$ is updated as:
\begin{equation}
\mathbf{h}_v^{\ell} = \sigma \left( \mathbf{W}^{\ell} \cdot \text{AGGREGATE} \left( \left\{ \mathbf{h}_u^{(\ell-1)} \mid u \in \mathcal{N}(v) \cup \{v\} \right\} \right) \right),
\end{equation}
After $L$ layers, global graph embeddings are obtained via pooling operations:
\begin{equation}
h_G = \text{POOL}\left( \left\{ \mathbf{h}_v^{L} \right\}_{v \in V} \right)
\end{equation}
To enhance structural consistency in the embedding space,
we define a GED-driven regression loss function $\mathcal{L}$ for the graph pair ($G^{i_1}, G^{i_2}$), which enforces the graph embeddings $h_{G^{i_1}}, h_{G^{i_2}}$ to approximate their reference semantic distance, by penalizing discrepancies between embedding distances and the GED values:
\begin{equation}
 \mathcal{L}_{} = (||h_{G^{i_1}} - h_{G^{i_2}}|| - GED(G^{i_1}, G^{i_2}))^2
\end{equation}
By incorporating geometric learning principles like Multi-Dimensional Scaling (MDS) \citep{mds, bai2019unsupervised},
we aim to ensure that the latent space preserves the global distance topology of the TBox-driven graph edits.

\paragraph{Transductive Training.} The training procedure is transductive, i.e. the model is optimized directly on a subset of the graphs it will be tasked to explain.
The training procedure is enforced on a predefined random graph subset of cardinality $p<N$ (empirically we set $p\approx N/2$).
To maintain efficiency for the transductive engine, the supervision signal (GED) is computed for only $N/2$ graph combinations within the dataset. As demonstrated by \citet{dimitriou2023structure}, this significantly accelerates the training process in the supervised setting without sacrificing substantial retrieval accuracy.

\paragraph{Counterfactual Retrieval.} 
After training, counterfactual retrieval is transformed from an NP-hard search into a high-speed vector comparison.
Ultimately, by considering a graph $G_q \in A$ as the query, having a graph embedding $h_{G_q}\in \mathbb{R}^d$, its counterfactual graph $G^k_c \in B^k \neq A$, corresponding to the graph embedding $h_{G^k_c}\in \mathbb{R}^d$, is retrieved by maximizing the cosine similarity between their embeddings:
\begin{equation} \label{eq:graph-retrieval}
G^k_c = \arg\max_{G^i_c \in B^{i} \ne A} 
\frac{h_{G_q}^\top h_{G^k_c}}{\| h_{G_q} \| \cdot \| h_{G^k_c} \|}, i=1,...,N.
\end{equation}


\subsection{Inductive Mode: The Efficiency Path}
\label{sec:unsupervised-bg}
The computational overhead of generating pairwise GED supervision signals in the transductive case (even with a reduced number of pairs) poses challenges for large-scale or real-time applications. To address this, we introduce U-CECE-Inductive, a lightweight alternative that circumvents GED-related operations during training.
Inductive refers to a model trained to learn reusable graph representations that can generalize beyond the specific graph pairs or candidate set seen during training. 
This path utilizes
unsupervised Graph Autoencoders (GAEs), which learn to encode a given scene graph into a low-dimensional latent representation and then reconstruct it from that representation. By training on individual graphs rather than pairs, these frameworks maintain a training complexity that scales linearly 
$\mathcal{O}(N)$ 
with dataset size, as opposed to the quadratic 
$\mathcal{O}(N^2)$ 
scaling of supervised metric learning, resulting in significant training time reduction.

In our experiments, we explore a spectrum of unsupervised architectures, where each method introduces specific mechanisms to enhance the quality of the learned latent embeddings.

\paragraph{Graph Autoencoders} at their foundational formulation
 comprise an encoder $f_{\theta}(G)=\mathbf{Z}$ that maps a graph $G$ with feature matrix $\mathbf{X}$ and adjacency matrix $\mathbf{A}$ to a latent embedding matrix $\mathbf{Z}$, and a decoder tasked to reconstruct the adjacency matrix $\hat{\mathbf{A}}$ via an embedding inner-product  \citep{kipf2016variationalgraphautoencoders}. The Variational Graph Autoencoder (VGAE) extends this by learning a probabilistic distribution (of mean $\mu$ and variance $\sigma$) rather than deterministic embeddings. The VGAE objective function balances reconstruction accuracy with a regularization term ensuring the latent distribution approximates a Gaussian prior:
\begin{equation}
    \mathcal{L}_\text{VGAE} = \mathcal{L}_\text{edge\_recon} +
    \mathcal{L}_\text{KL}
\end{equation}
\noindent where $\mathcal{L}_\text{edge\_recon} = \mathbb{E}[\log p(\mathbf{A}|\mathbf{Z})]$ maximizes the likelihood of the reconstructed graph structure (implemented as sampled Binary Cross-Entropy loss between edge predictions and ground truth adjacency matrix), and $\mathcal{L}_{\text{KL}}$ represents the Kullback-Leibler divergence between the learned distribution $q(\mathbf{Z}|\mathbf{X},\mathbf{A})$ and the prior $p(\mathbf{Z})$.



\paragraph{Adversarially Regularized Variational Graph Autoencoders (ARVGA)} \citep{arvga}
integrate principles from Generative Adversarial Networks (GANs) to further enforce a robust latent space distribution. They employ a discriminator network $\mathcal{D}$ alongside the VGAE generator $\mathcal{G}$ (encoder) to distinguish between the learned latent codes and samples from a prior distribution (typically Gaussian). This adversarial training scheme involves a min-max game where the generator tries to fool the discriminator, adding a regularization term that encourages the embeddings to match the prior more effectively than KL divergence alone. The objective combines the standard VGAE loss with this adversarial mechanism:
\begin{equation}
    \mathcal{L}_{\text{ARVGA}} = \mathcal{L}_{\text{VGAE}} + \mathcal{L}_{\text{adv}}
\end{equation}
\noindent where $\mathcal{L}_{\text{adv}}$ denotes the standard GAN minimax loss, in which the generator tries to fool the discriminator into classifying the embeddings as samples drawn from the prior $p_z$.



\paragraph{Graph Feature Autoencoders (GFA)} \citep{feat_gae} further boost node information learning by incorporating a dedicated feature decoder, contrary to the topological reconstruction focus (predicting edges) of 
 standard GAEs. Instead of solely reconstructing $\mathbf{A}$, GFA explicitly reconstructs the node feature matrix $\mathbf{X}$, ensuring that semantic attributes are preserved in the latent space. The loss function is modified to include feature reconstruction error:
\begin{equation}
    \mathcal{L}_\text{GFA} = \mathcal{L}_\text{VGAE} + \mathcal{L}_\text{feat\_recon}
\end{equation}
\noindent where $\mathcal{L}_\text{feat\_recon}$ is implemented as a Mean Squared Error loss between  ground truth  $\mathbf{X}$ and  predicted $\hat{\mathbf{X}}$.

\paragraph{SCENIR}
is a state-of-the-art scene graph retrieval framework that synthesizes the aforementioned components into a unified architecture \citep{chaidos2025scenirvisualsemanticclarity}. It employs a split-encoder design (independent GNN branches for $\mu$ and $\sigma$ parameters) to better capture distinct structural and uncertainty features. Furthermore, it replaces the standard inner-product decoder with parallel MLP-based decoders for both edges and features to prevent over-smoothing and capture complex relations. Its training objective combines all previous aspects: reconstruction of features and edges, adversarial regularization, and variational inference:
\begin{equation} \label{eq:scenir-loss}
    \mathcal{L}_\text{SCENIR} = \lambda_{1}(\mathcal{L}_\text{feat\_recon} + \mathcal{L}_\text{edge\_recon}) + \lambda_{2}\mathcal{L}_\text{adv} + \lambda_{3}\mathcal{L}_\text{KL}
\end{equation}
The defining advantage of SCENIR is its natural alignment with inductive settings, though it remains versatile enough to also support  transductive  scenarios. While transductive models are "dataset-locked"(trained and evaluated on a fixed set of instances), our inductive engine learns a general transformation function that can generalize to unseen queries at inference time without re-training.
Regardless of the specific architecture used, global graph embeddings are obtained via pooling operations on the latent node embeddings, allowing for efficient nearest-neighbor retrieval in the latent space, approximating the retrieval results of  GED.

\subsection{Performance and Complexity Analysis}
\label{sec:complexity_analysis}
The selection between the \textbf{Transductive} and \textbf{Inductive} engines involves a fundamental trade-off between topological precision and computational scalability. While both seek to approximate GED via latent space retrieval instead of direct GED learning, their operational overhead is governed by distinct complexity classes.
\paragraph{Training Complexity and Scaling.}
The primary distinction between the two modes lies in the construction of the latent manifold. The \textbf{Transductive Mode} relies on supervised metric learning, where the model is optimized to minimize the regression loss between embedding distances and reference GED. For a dataset of $N$ scene graphs, the absolute computational cost encompasses both the GNN forward passes and the generation of the pairwise supervision signal: $\mathcal{O}(N \cdot \text{cost}(\text{GNN}) + N^2 \cdot \text{cost}(\text{GED}))$. Because there are $\binom{N}{2} \in \mathcal{O}(N^2)$ potential pairwise combinations, the quadratic supervision bottleneck strictly dominates the linear message-passing operations. Even with empirical $N/2$ optimization, the effective training complexity reduces to:
\begin{equation}
    T_{\text{train(trans)}} \propto \mathcal{O}(N^2 \cdot \text{cost}(\text{GED}))
\end{equation}
where $\text{cost}(\text{GED})$ represents the computational overhead of a single graph matching operation. While exact solvers based on $A^*$ search scale exponentially ($\mathcal{O}(2^{|V|})$), practical implementations of the Transductive engine, 
such as the one deployed in this paper, utilize the Volgenant-Jonker (VJ) algorithm \citep{jonker1988shortest}
to reduce the matching to a Linear Assignment Problem (LAP). This substitution brings the cost down to $\mathcal{O}(|V|^3)$, where $|V|$ is the number of vertices in the larger graph. However, even with this polynomial reduction, the $\mathcal{O}(N^2)$ scaling of the pairwise training regime remains a significant bottleneck, effectively bounding the Transductive engine to high-fidelity, sparse data regimes.

In contrast, the \textbf{Inductive Mode} leverages self-supervision via reconstruction. Because the loss function is computed per graph instance rather than per pair, the $\mathcal{O}(N^2)$ supervision bottleneck is entirely eliminated. The training complexity scales strictly \textit{linearly} with the dataset size, bounded only by the efficiency of the underlying GNN backbone:
\begin{equation}
    T_{\text{train(induc)}} \propto \mathcal{O}(N \cdot \text{cost}(\text{GNN}))
\end{equation}
Here, $\text{cost}(\text{GNN})$ varies depending on the specific message-passing architecture utilized (e.g., GCN, GIN, GAT). For a network with $L$ layers and an embedding dimensionality of $d$, the cost generally scales as $\mathcal{O}(L \cdot (|V| d^2 + |E| d))$, accounting for both the node-wise feature transformations and the edge-wise neighborhood aggregations (including attention coefficient computations in GAT). In large-scale datasets, the total number of graphs $N$ vastly exceeds the structural parameters of any individual scene graph ($N \gg |V|, |E|$), making $\text{cost}(\text{GNN})$ a dataset-independent constant factor. By confining the complexity strictly to this linear $O(N)$ scaling, the inductive mode enables the processing of vast data distributions that remain computationally inaccessible to transductive supervised methods.

\paragraph{Inference and Retrieval Efficiency.}
At inference time, both engines utilize an identical retrieval pipeline, transforming the NP-hard structural matching bottleneck into a vector similarity task. Given a query graph $G_q$, the complexity of retrieving the counterfactual $G^k_c$ involves a forward pass through the GNN $f_\theta$ followed by a nearest-neighbor search across the $N-1$ candidates:
\begin{equation}
    T_{\text{inf}} \propto \mathcal{O}(\text{cost}(\text{GNN}{q}) + N \cdot d)
\end{equation}
where $\text{cost}(\text{GNN}{q})$ represents the linear message-passing cost for the single query graph, and $d$ is the latent dimensionality.
However, the \textbf{Inductive Mode} provides superior operational flexibility. While the transductive model is effectively ``dataset-locked'' ($\mathcal{G}_{\text{train}} \approx \mathcal{G}_{\text{test}}$), the inductive engine learns a generalizable mapping $f: \mathcal{G} \to \mathbb{R}^d$. This allows for the immediate generation of counterfactuals for novel, out-of-distribution scene graphs without the need for the $\mathcal{O}(N^2)$ retraining cycle required by transductive supervised signals.


\paragraph{Precision vs. Topological Generalization.}
The \textit{Precision Path} is optimized to favor the minimality of the semantic edit, making it the preferable choice for applications where preserving fidelity to the original structure is critical and arbitrary deviations are undesirable. Conversely, the \textit{Efficiency Path} captures the broader manifold of the scene graph distribution. Rather than relying on granular, potentially noisy input edges, this inductive approach leverages the learned global topology to infer structural relationships probabilistically, effectively attaining generalizability. 
\section{Experimental Analysis}
\label{sec:experiments}

U-CECE is evaluated as an adaptive framework rather than as a competitor to existing retrieval systems. The engines from each tier have been validated against state-of-the-art methods in the works that introduced them, and this paper does not rest on outperforming an external system. Our evaluation is therefore designed to prioritize the framework's internal consistency across expressivities, the minimality and cost of the resulting edits, the computational profile of each engine, and alignment with human understanding. Specifically, it is structured across the following primary axes:
i) \textit{Performance Across Expressivity Modes:} a traditional benchmarking of the different expressivity levels using standard local counterfactual metrics to establish a baseline of efficacy; ii) \textit{Structural Engine Optimization:} an exhaustive evaluation of various GNN architectures for the Transductive and Inductive engines at the highest expressivity tier, including an assessment of inductive capabilities; 
iii) \textit{Efficiency Analysis: } an assessment of computational overhead and retrieval latency across all U-CECE levels, with a specific focus on the structural engine; 
iv) \textit{Human-Centric Validation:} an evaluation of explanation quality through manual inspection of qualitative examples and a systematic human survey to validate the semantic faithfulness of our most accurate structural engine against human visual intuition; and v) \textit{Neural Interpretability:} an exploration of counterfactual interpretability from the perspective of Large Vision-Language Models (LVLMs) as a Judge by recreating our human survey with neural-based annotators to measure the alignment between machine and human perception.


\subsection{Datasets}
Conceptual counterfactuals are inherently versatile and applicable across a broad spectrum of data domains. Previous research has established their efficacy on standard XAI benchmarks or automatically generated datasets \citep{dervakos2023choose, dimitriou2023structure}. In this study, our primary objective is to rigorously evaluate the distinct expressivity levels and retrieval modes of  U-CECE. To this end, we focus on real-world \textit{image} datasets characterized by high-quality annotations at both the concept and graph levels. These controlled datasets provide the necessary structural richness to be employed horizontally throughout our experimental analysis, ensuring a consistent and robust evaluation of our transductive and inductive engines.

All tiers of expressivity are evaluated on the exact same subset of Visual Genome (VG) \citep{krishna2017visual} images, varying only in the scope of the structural information extracted. Specifically, to serve L1 expressivity, concept abstraction is restricted to isolated object annotations. For L2 expressivity and higher, the representations incorporate full relational data. Furthermore, both L1 and L2 leverage the WordNet synsets provided by VG to enrich their respective elements (utilizing synsets exclusively for objects in L1, and for both objects and relations in L2). For these experiments, we adopt the dataset split introduced by SGCE \citep{dimitriou2023structure}, which treats edge density as a decisive factor for the selected explanation algorithm. Specifically, the VG-DENSE split contains 500 highly interconnected graphs with fewer isolated nodes, while the VG-RANDOM split contains 500 arbitrarily selected graphs. 
We further use the Caltech-UCSD Birds (CUB) dataset \citep{wah2011caltech}, which offers fine-grained concept-level annotations of bird parts and their visual attributes. While graph-level annotations are not directly provided, they can be straightforwardly constructed and connected to WordNet  \citep{dimitriou2023structure}, resulting in 422 tree-shaped graphs of fairly small depth whose size depends on the number of visible parts in each image. 

The roles of these datasets are strategically partitioned across our evaluation axes. CUB, a widely adopted benchmark in visual XAI, serves as the primary benchmark here as well, being used horizontally throughout the entire experimental section; its fine-grained attributes and manageable graph structures make it uniquely suited for evaluating the complete pipeline, ranging from standard quantitative metrics and GNN optimization to the human perception surveys and neural interpretability studies. In contrast, VG is utilized specifically to investigate the impact of structural expressivity. By leveraging VG-DENSE and VG-RANDOM, where edge density serves as a proxy for relational importance, we report on performance across our retrieval engines and provide qualitative insights into how U-CECE handles complex (edge-dense) scenes.

\subsection{Model Details and Training Protocols}

To evaluate our retrieval engines, we utilize several GNN architectures across two distinct training protocols. For U-CECE-Transductive, we leverage GCN \citep{gcn}, GAT \citep{gat}, and GIN \citep{gin} within the Siamese similarity framework of SGCE. 
We use GED as the supervision signal for SGCE, intentionally omitting alternative semantic similarity baselines (e.g. CLIP-based similarities on image captions) due to their inherent instability demonstrated in previous work \citep{chaidos2025scenirvisualsemanticclarity}. Such approaches produce rankings highly sensitive to the choice of sentence embedder, rendering the use of a deterministic, mathematically rigorous metric like GED necessary. 
For the unsupervised U-CECE-Inductive engine, we employ GAE-based variations (Section \ref{sec:unsupervised-bg} --GFA, ARVGA, SCENIR). We conduct thorough ablation studies across these variants for both engines to identify the optimal configurations for counterfactual retrieval. Unsupervised GNNs are evaluated in both transductive and inductive settings. While the transductive models are trained and evaluated directly on specific benchmark subsets (consisting of 500 VG graphs and 422 CUB graphs), the inductive capability is assessed by pretraining on larger sets of both datasets (containing ~28k graphs for VG / ~11k graphs for CUB), strictly excluding the evaluation samples. By anchoring all experiments to the same 500 VG and 422 CUB test graphs, we ensure a consistent and fair comparison between the specialized representations of the transductive models and the generalized latent spaces of the inductive engine. Details are provided in Appendix \ref{app:train}. 

As classifiers under explanation, we utilize the setup laid out by  \cite{dervakos2023choose, dimitriou2023structure}, i.e. a ResNet-50 for CUB to ensure experimental alignment and a Places365 model \citep{places} to better capture the scene-centric nature of VG, which object-focused ImageNet classifiers are less equipped to represent. For datasets with ground-truth classification labels (CUB) the target class for the counterfactual explanation is selected as the category most frequently confused with the source class \citep{vandenhende2022making}. In scenarios where labels are unavailable (VG), the target class is instead defined as the category of the single highest-ranked instance that belongs to any class other than the source one.

To ensure a high-fidelity and internally consistent evaluation across all expressivity tiers of  U-CECE, we  implement several strategic refinements to the underlying experimental components. To provide a more robust representation of scene semantics, the classifiers are optimized to assign improved labels. The mathematical protocol for calculating structural edits - particularly the averaging of edge-level operations - is standardized to provide a more precise and stable measure of topological change across datasets. For the L1 and L2 expressivity modes, we utilize a more accurate WordNet synset initialization to provide a more rigorous semantic grounding. By ensuring that all submodules operate on a consistent, high-quality baseline, the resulting evaluation more accurately captures the inherent performance of the unified framework. Consequently, while the reported metrics may deviate from those in previous publications due to these optimizations, this standardized approach ultimately maintains empirical integrity.

\subsection{Evaluation Metrics}

To evaluate the U-CECE retrieval engines, we utilize ranking metrics to measure the alignment between GNN-based rankings and the reference GED.
All GED values are computed with the suboptimal bipartite heuristic \citep{jonker1988shortest, fankhauser2011speeding}, implemented in DGL \citep{wang2019deep}, following SGCE. Node and edge operation costs are computed as the shortest path between concepts in the WordNet hierarchy. The reference is therefore approximate but fully deterministic and independent of the retrieval engines under evaluation, which is the property our ranking metrics require.
We report Precision@k ($P@k$) and Normalized Discounted Cumulative Gain ($nDCG@k$) for $k \in \{1, 2, 4\}$. While counterfactuals primarily prioritize the top-ranked result ($k=1$) to ensure structural minimality, the inclusion of $k=2$ and $k=4$ during ablation studies provides a more comprehensive overview of the quality and stability of different engine variants.
In the default case, all valid retrieved items are considered equally relevant. However, since the top-ranked instance returned by the GED search represents the reference counterfactual, we also employ \textit{binary} metric counterparts. This stricter evaluation measures the engine’s precision in recovering the reference instance for structural minimality. All aforementioned metrics are bounded in the range $[0, 1]$, where higher values indicate better retrieval accuracy and closer alignment with the reference ranking.

More tailored to counterfactuals, we analyze edit-level metrics based on the number of node and edge operations ($R, D, I$). These counts are averaged across the dataset for the top-ranked retrieval ($k=1$) and denoted as $\bar{e}_{node}$, $\bar{e}_{edge}$, and $\bar{e}_{total}$. These results are paired with the mean GED ($\overline{\text{GED}}$) of the top-ranked retrievals to offer a view into the cost of the operations as well. These structural metrics have a theoretical lower bound of 0 (no edits needed), with lower values being preferred as they signify greater structural minimality and a more concise transition from the source to the counterfactual state. By quantifying these transitions, we gain insight into the logical complexity of the explanations and the effectiveness of the U-CECE engines in approximating the foundational graph-based reasoning.

When comparing the retrieval power across the U-CECE expressivity tiers (L1-Atomic, L2-Relational, and L3-Structural), we strictly evaluate the first-ranked proposed counterfactual ($k=1$). In this context, the focus shifts from the performance of the retriever in isolation to a deterministic measure of how closely the final generated explanation approximates the GED-based counterfactual. To this end, in these comparative experiments, we report the average edit counts and costs, and only $P@1$.
The reported numbers quantify how closely each tier recovers a counterfactual selected under a common structural reference; they are not an absolute measure of explanation quality, which we address separately through qualitative inspection and human evaluation. As noted in Section \ref{sec:structural_distance}, the reference subsumes the atomic and relational metrics rather than competing with them.

\subsection{Results}

\subsubsection{Expressivity Level Mode Performance}

In Table \ref{tab:expressivity}, we evaluate the performance of the three U-CECE expressivity tiers in a strict top-1 counterfactual retrieval task. Across all datasets, $P@1$ scores appear relatively low (ranging from $0.078$ to $0.248$). This is indicative of the challenging nature of the task: retrieving the single reference instance from a candidate pool of hundreds of potentially conceptually similar images. This is particularly evident in the CUB dataset, where inter-class variation is minimal. Bird species often share a restricted subset of attributes and differ primarily in the visibility or orientation of specific parts. Consequently, while $P@1$ is conservative, the retrieval quality largely improves as the rank threshold $k$ increases (Appendix \ref{app:expressivity-more}).

For the CUB dataset, we observe that L1  actually outperforms the L3 engine in $P@1$. This finding is consistent with the nature of CUB graphs, which are essentially tree-shaped structures of low depth representing bird parts without explicit spatial orientation. In such a domain, a coarser explanation often approximates the reference more effectively. 
Interestingly, while the L2 engine achieves the lowest edge edit count ($\bar{e}_{edge} = 2.9$), its associated cost ($\overline{GED} = 219.638$) is the highest in the cohort. This discrepancy suggests a form of local structural disruption: since the L2 tier groups relations into "triplet sets" and ignores the broader neighboring topology, it may suggest fewer edits that are semantically "expensive" or nonsensical, such as swapping a functional part (e.g., a beak) for a purely visual attribute (e.g., a solid pattern).

\begin{wraptable}{r}{0.58\textwidth}
\centering \scriptsize
\renewcommand{\arraystretch}{1.7}
\caption{Comparison of the three expressivity U-CECE tiers across CUB, VG-DENSE, and VG-RANDOM datasets. For U-CECE-Structural, the best-performing variant and architecture are selected. \textbf{Bold} for best result.}
\label{tab:expressivity}
\begin{tabular}{clccccc}
\toprule
& Level & $P@1 \uparrow$ & $\bar{e}_{node} \downarrow$ & $\bar{e}_{edge} \downarrow$ & $\bar{e}_{total} \downarrow$ & $\overline{GED} \downarrow$ \\ 
\midrule
        \multirow{3}{*}{\rotatebox[origin=c]{90}{\textbf{CUB}}} 
        & Atomic     & \textbf{0.175} & \textbf{5.750} & 3.475 & 9.225 & \textbf{210.594} \\
        & Relational & 0.137 & 6.175 & \textbf{2.900} & \textbf{9.075} & 219.638 \\
        & Structural & 0.138 & 7.506 & 6.038 & 13.544 & 217.131 \\ 
        \midrule
        \multirow{3}{*}{\rotatebox[origin=c]{90}{\textbf{\shortstack{VG \\ DENSE}}}} 
        & Atomic     & 0.212 & \textbf{4.564} & 11.236 & 15.800 & 109.170 \\
        & Relational & 0.166 & 5.018 & 10.702 & 15.720 & 113.532 \\
        & Structural & \textbf{0.248} & 4.942 & \textbf{10.312} & \textbf{15.254} & \textbf{105.194} \\ 
        \midrule
        \multirow{3}{*}{\rotatebox[origin=c]{90}{\textbf{\shortstack{VG \\ RANDOM}}}} 
        & Atomic     & \textbf{0.216} & \textbf{11.176} & 12.216 & \textbf{23.392} & 160.982 \\
        & Relational & 0.078 & 13.150 & 11.868 & 25.018 & 179.868 \\
        & Structural & 0.200 & 12.392 & \textbf{11.840} & 24.232 & \textbf{159.356} \\ 
        \bottomrule
\end{tabular}
\end{wraptable}
The merits of L3 expressivity are most prominent in the VG-DENSE split. Here, the Structural engine achieves a $P@1$ of $0.248$, representing a $3.6$ and $8.2$ points closer agreement with the reference over the L1 and L2 tiers, respectively. In scenarios where graphs are highly interconnected, higher structural expressivity is required to navigate the complex relational dynamics and retrieve an accurate counterfactual. Furthermore, the L3 engine achieves the lowest total edit count $\bar{e}_{total}$ and the lowest  $\overline{GED}$ cost, proving its efficiency in dense environments. In contrast, VG-RANDOM  exhibits trends more similar to CUB. When scene graphs are arbitrarily selected and potentially sparser, the performance advantage of L3 over L1 diminishes. This reinforces our intuition that the necessity for structural detail is a function of relational density; when object interactions are sparse, a set-based or atomic view of concepts is often sufficient for retrieval.

\begin{figure*}[t!]
    \centering
    \includegraphics[width=\textwidth]{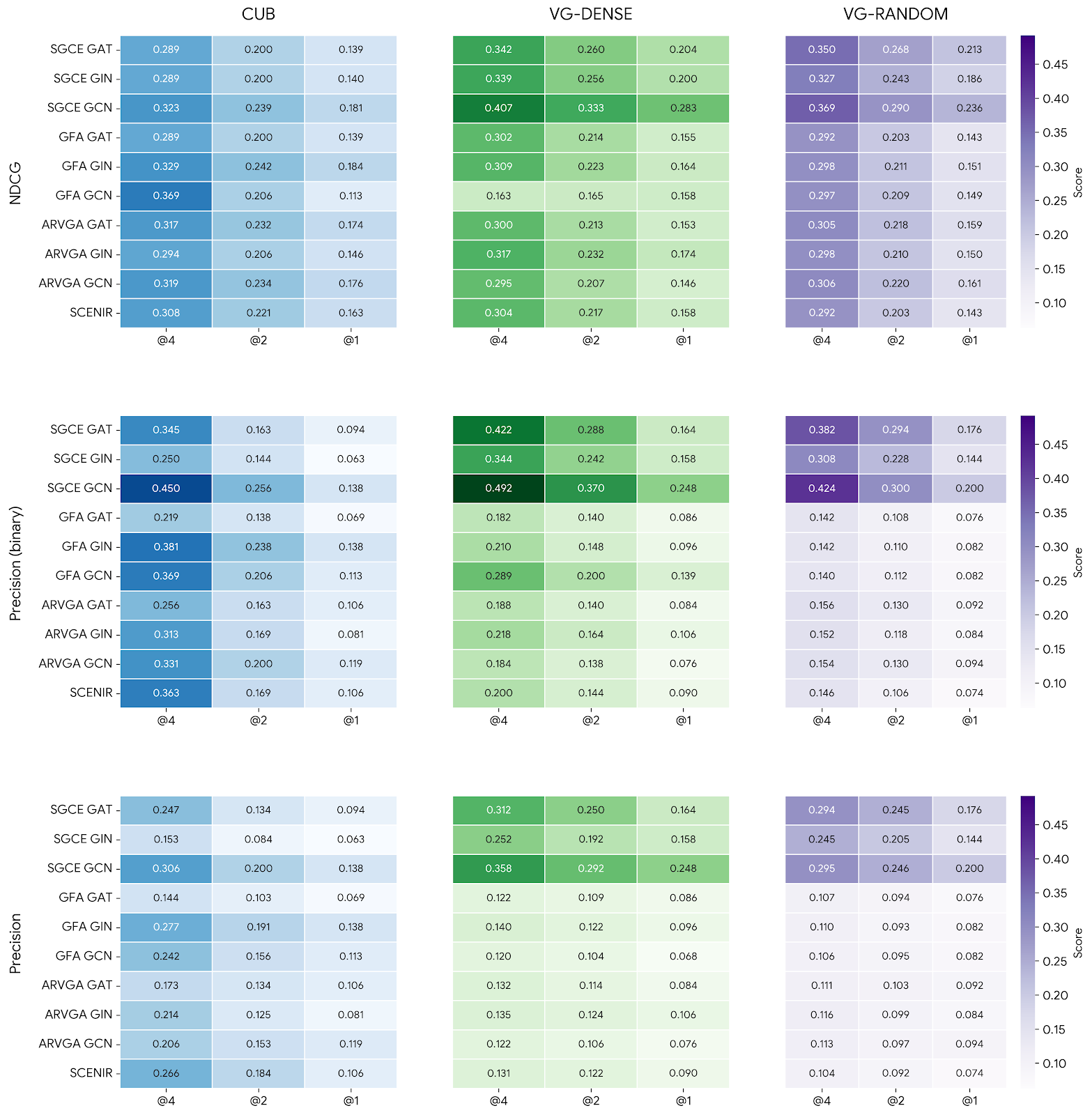}
    \caption{
        Retrieval performance comparison of underlying GNN components of the U-CECE-Structural (Transductive) mode, including  SGCE, GFA, ARVGA, and SCENIR architectures. Ablations on the GNN variant are presented where applicable. 
        Color intensity in heatmaps is normalized globally across all metrics and datasets to facilitate direct comparisons. Distinct color hues represent the different datasets.
    }
    \label{fig:fine_grained_heatmaps}
\end{figure*}

The L3 engine emerges as a robust and versatile retriever; even in domains where it does not achieve the highest precision, it remains highly competitive while providing the most verbose and topologically faithful explanations. The L1 engine is remarkably consistent and cheap, making it an excellent candidate for datasets with low relational importance. Finally, the L2 engine shows its greatest potential in dense settings but can suffer from high semantic costs. Ultimately, these quantitative metrics confirm that \textit{no single expressivity tier is universally superior}. 
Instead, the choice of engine should be informed by the underlying data complexity. 
This pattern also speaks to the neutrality of our evaluation reference: the structural tier leads on one of three splits, precisely the split our hypothesis predicts.
To further comprehend the subjective utility of these levels, we provide a qualitative inspection (Section \ref{sec:qualitative}), exploring how such numeric differences translate into human-understandable explanations.

\subsubsection{Structural Retrieval Engine Trade-offs}

Having established the efficacy of the structural L3 tier, we now provide a fine-grained analysis of the architectural backbones and training paradigms within the U-CECE-Structural engine. This section evaluates the trade-offs between supervised and unsupervised regimes, specifically comparing various GNN backbones across the transductive and inductive setups. To establish a fair comparison, we first utilize a standardized transductive protocol for all GNN-based variations. We perform systematic ablations on the message-passing backbones, interchanging GCN, GAT, and GIN, within the supervised SGCE framework (U-CECE-Transductive) and the unsupervised GFA, ARVGA, SCENIR autoencoders (U-CECE-Inductive). SCENIR is implemented with GIN to ensure best retrieval performance \citep{chaidos2025scenirvisualsemanticclarity}.

Figure \ref{fig:fine_grained_heatmaps} illustrates
the retrieval performance for $k \in \{1, 2, 4\}$. Across all heatmaps, we observe a monotonic improvement in retrieval fidelity as $k$ increases. The results confirm the stronger performance of the supervised SGCE models in approximating the reference GED rankings: specifically, SGCE-GCN achieves the most robust performance on the VG subsets, where its ability to model complex relational densities is most pronounced. Interestingly, the performance deltas between architectures are more significant in the P@k and Binary P@k metrics, whereas nDCG@k exhibits a smoother distribution across variants. This suggests that while the exact recovery of top-ranked instances varies by backbone, the overall rank-order stability remains relatively consistent across different supervised GNNs. In contrast, unsupervised models generally underperform in the transductive setting, as they lack the explicit similarity-driven signal used by SGCE. While there is no definitive winner between GFA, ARVGA and SCENIR globally, the models that utilize a feature decoder (GFA, SCENIR) exhibit competitive performance on CUB, particularly for higher $k$ values, suggesting that the concentrated distribution of node and edge attribute semantics in CUB is more readily captured by feature-reconstruction backbones than the more relational, structure-heavy scene graphs of VG.

\begin{table*}[t]
\centering
\caption{Comparison of edit metrics and $\overline{GED}$ across CUB, VG-DENSE, and VG-RANDOM datasets for all U-CECE-Structural models in the transductive setting.}
\label{tab:final-comparison}
\renewcommand{\arraystretch}{1.4} 
\resizebox{\textwidth}{!}{ 
\begin{tabular}{clcccccccccccc}
\toprule
& & \multicolumn{4}{c}{\textbf{CUB}} & \multicolumn{4}{c}{\textbf{VG-DENSE}} & \multicolumn{4}{c}{\textbf{VG-RANDOM}} \\
\cmidrule(lr){3-6} \cmidrule(lr){7-10} \cmidrule(lr){11-14}
& \footnotesize Model & \footnotesize $\bar{e}_{node} \downarrow$ & \footnotesize $\bar{e}_{edge} \downarrow$ & \footnotesize $\bar{e}_{total} \downarrow$ & \footnotesize $\overline{GED} \downarrow$ 
        & \footnotesize $\bar{e}_{node} \downarrow$ & \footnotesize $\bar{e}_{edge} \downarrow$ & \footnotesize $\bar{e}_{total} \downarrow$ & \footnotesize $\overline{GED} \downarrow$
        & \footnotesize $\bar{e}_{node} \downarrow$ & \footnotesize $\bar{e}_{edge} \downarrow$ & \footnotesize $\bar{e}_{total} \downarrow$ & \footnotesize $\overline{GED} \downarrow$ \\
\midrule

\multirow{3}{*}{SGCE}
& GAT & 7.281 & 5.506 & 12.788 & 224.71 & 5.278 & 10.892 & 16.170 & 108.64 & 12.810 & 11.978 & 24.788 & 159.36 \\
& GIN & 9.732 & 10.181 & 19.913 & 237.63 & 5.106 & 10.766 & 15.872 & 116.17 & 12.792 & 11.394 & 24.186 & 173.72 \\
& GCN & 7.506 & 6.038 & 13.544 & \textbf{217.13} & 4.942 & \textbf{10.312} & \textbf{15.254} & \textbf{105.19} & 12.392 & 11.840 & 24.232 & \textbf{159.32} \\
\midrule
\hline

\multirow{3}{*}{GFA}
& GAT & 8.775 & 5.194 & 13.969 & 243.44 & 5.070 & 11.564 & 16.634 & 131.58 & 12.988 & 12.412 & 25.400 & 198.78 \\
& GIN & 6.488 & 4.650 & 11.138 & 217.37 & \textbf{4.856} & 10.786 & 15.642 & 129.86 & 12.916 & 12.218 & 25.134 & 200.97 \\
& GCN & 6.456 & \textbf{3.738} & \textbf{10.194} & 220.84 & 5.030 & 11.432 & 16.462 & 132.05 & 12.696 & 12.338 & 25.034 & 201.04 \\
\midrule

\multirow{3}{*}{ARVGA}
& GAT & 8.050 & 4.900 & 12.950 & 238.03 & 5.062 & 11.482 & 16.544 & 133.48 & 12.508 & 12.192 & 24.700 & 196.95 \\
& GIN & \textbf{6.444} & 3.919 & 10.363 & 228.74 & 4.894 & 10.912 & 15.806 & 128.94 & \textbf{12.234} & \textbf{11.794} & \textbf{24.028} & 198.65 \\
& GCN & 6.625 & 4.106 & 10.731 & 223.14 & 5.048 & 11.316 & 16.364 & 132.29 & 12.464 & 12.088 & 24.552 & 198.46 \\
\midrule
 SCENIR & GIN & 6.369 & 4.7 & 11.069 & 223.91  & 5.028 & 11.546 & 16.574 & 131.56   & 12.646 & 12.352 & 24.998 & 197.68  \\
\bottomrule
\end{tabular}
}
\end{table*}

While the retrieval metrics establish ranking efficacy, the edit metrics reveal a divergence between topological minimality and semantic relevance (Table \ref{tab:final-comparison}). In VG-DENSE, the supervised SGCE-GCN variant remains the dominant performer, achieving the lowest total edit count ($\bar{e}_{total} = 15.254$) and the lowest semantic cost ($\overline{GED} = 105.19$). 
For SGCE-GIN, similarly to the retrieval metrics, there is a significant performance drop specifically on CUB ($\bar{e}_{total} = 19.913$, $\overline{GED} = 237.63$); despite extensive optimization, we hypothesize this stems from GIN's rigid focus on structural isomorphism, which causes it to overfit to the uniform anatomical topologies of the bird graphs rather than exploiting the critical semantic variance found within the continuous attributes.
Furthermore, in the CUB and VG-RANDOM subsets, we observe a distinct behavior: unsupervised models such as GFA-GCN and ARVGA-GIN frequently achieve lower raw edit counts, nearly 25\% lower than their supervised counterparts on CUB, yet maintain significantly higher $\overline{GED}$ costs, confirming that while unsupervised autoencoders often find "shorter" topological paths, these transitions frequently result in semantically noisier shortcuts that fail to respect the learned semantic manifold. Consequently, these results reinforce our argument that raw edit numbers are an insufficient standalone metric for conceptual counterfactuals and must be evaluated alongside $\overline{GED}$ to accurately measure the logical complexity and structural fidelity of the retrieved explanations.

\begin{table*}[t!]
\centering
\caption{U-CECE-Inductive performance across various GNN backbones  (SCENIR, ARVGA, GFA).
In \underline{underlined} instances, the inductive model strictly outperforms its transductive counterpart.}
\label{tab:inductive-detailed-summary}
\renewcommand{\arraystretch}{1.3}
\resizebox{\textwidth}{!}{ 
\begin{tabular}{clccccccccccccc}
\toprule
 & \multirow{2}{*}{\textbf{Model}} & \multicolumn{3}{c}{\textbf{nDCG ($\uparrow$)}} & \multicolumn{3}{c}{\textbf{Precision (binary) ($\uparrow$)}} & \multicolumn{3}{c}{\textbf{Precision ($\uparrow$)}} & \multicolumn{4}{c}{\textbf{Edits \& GED ($\downarrow$)}} \\
\cmidrule(lr){3-5} \cmidrule(lr){6-8} \cmidrule(lr){9-11} \cmidrule(lr){12-15}
 & & @4 & @2 & @1 & @4 & @2 & @1 & @4 & @2 & @1 & $\bar{e}_{node}$ & $\bar{e}_{edge}$ & $\bar{e}_{total}$ & $\overline{GED}$ \\
\midrule

\multirow{7}{*}{\rotatebox[origin=c]{90}{\textsc{Cub}}} 
 
 & GFA-GAT & 0.279 & 0.189 & 0.128 & \underline{0.263} & \underline{0.163} & 0.063 & \underline{0.194} & \underline{0.116} & 0.063 & \underline{7.175} & \underline{4.538} & \underline{11.713} & \underline{230.469} \\
 & GFA-GIN & 0.296 & 0.208 & 0.149 & 0.288 & 0.169 & 0.081 & 0.192 & 0.131 & 0.081 & 7.219 & \underline{4.369} & 11.588 & 230.888 \\
 & GFA-GCN & 0.313 & \underline{0.227} & \underline{0.169} & 0.35 & 0.206 & \underline{0.125} & \textbf{\underline{0.270}} & \underline{0.178} & \underline{0.125} & 6.563 & \textbf{4.106} & 10.669 & \textbf{221} \\
 & ARVGA-GAT & 0.304 & 0.217 & 0.158 & \underline{0.313} & \underline{0.169} & 0.1 & \underline{0.223} & \underline{0.175} & 0.1 & \underline{6.694} & 5.481 & \underline{12.175} & \underline{228.806} \\
 & ARVGA-GIN & \underline{0.3} & \underline{0.212} & \underline{0.153} & 0.313 & 0.156 & \underline{0.1} & \underline{0.258} & \underline{0.175} & \underline{0.1} & \underline{6.313} & 4.756 & 11.069 & \underline{224.694} \\
 & ARVGA-GCN & \textbf{\underline{0.325}} & \textbf{\underline{0.241}} & \textbf{\underline{0.184}} & \textbf{\underline{0.356}} & 0.2 & \textbf{\underline{0.138}} & \underline{0.267} & \textbf{\underline{0.188}} & \textbf{\underline{0.138}} & \underline{6.488} & 5.137 & 11.625 & \underline{221.531} \\
 & SCENIR & \underline{0.32} & \underline{0.235} & \underline{0.177} & 0.338 & \textbf{\underline{0.225}} & \underline{0.125} & 0.25 & 0.159 & \underline{0.125} & \textbf{\underline{6.2}} & \underline{4.169} & \textbf{\underline{10.369}} & \underline{222.719} \\
\midrule

\multirow{7}{*}{\rotatebox[origin=c]{90}{\textsc{VG-Dense}}} 
 
 & GFA-GAT & 0.293 & 0.204 & 0.144 & \underline{0.19} & 0.128 & 0.076 & \underline{0.131} & 0.107 & 0.076 & 5.1 & 11.752 & 16.852 & 132.6 \\
 & GFA-GIN & \textbf{0.307} & \textbf{0.22} & \textbf{0.162} & \underline{0.214} & \textbf{\underline{0.154}} & \textbf{0.096} & \underline{0.145} & \underline{0.128} & \textbf{0.096} & \textbf{4.972} & \textbf{11.468} & \textbf{16.44} & \textbf{130.116} \\
 & GFA-GCN & \underline{0.294} & \underline{0.205} & 0.145 & 0.158 & 0.114 & 0.074 & \underline{0.122} & \underline{0.107} & \underline{0.074} & 5.196 & 11.986 & 17.182 & 134.4 \\
 & ARVGA-GAT & \underline{0.304} & \underline{0.217} & \underline{0.158} & \underline{0.19} & \underline{0.146} & \underline{0.088} & 0.13 & \underline{0.116} & 0.088 & 5.088 & 11.638 & 16.726 & \underline{132.212} \\
 & ARVGA-GIN & \textbf{0.307} & \textbf{0.22} & \textbf{0.162} & \textbf{0.218} & \textbf{0.154} & \textbf{0.096} & \textbf{\underline{0.148}} & \textbf{\underline{0.133}} & \textbf{0.096} & 5.014 & 11.604 & 16.618 & 130.316 \\
 & ARVGA-GCN & 0.293 & 0.205 & 0.145 & 0.164 & 0.116 & 0.074 & \underline{0.128} & \underline{0.108} & 0.074 & 5.196 & 11.926 & 17.122 & \underline{132.092} \\
 & SCENIR & 0.285 & 0.196 & 0.135 & 0.182 & 0.116 & 0.062 & 0.113 & 0.092 & 0.062 & 5.194 & 11.852 & 17.046 & 132.2 \\
\midrule

\multirow{7}{*}{\rotatebox[origin=c]{90}{\textsc{VG-Random}}} 
 
 & GFA-GAT & 0.289 & 0.2 & 0.14 & \underline{0.144} & 0.104 & 0.076 & \underline{0.12} & \underline{0.097} & 0.076 & 13.136 & 12.76 & 25.896 & \underline{197.496} \\
 & GFA-GIN & 0.296 & 0.208 & 0.149 & \textbf{\underline{0.158}} & \underline{0.112} & \underline{0.084} & \underline{0.121} & \underline{0.096} & \underline{0.084} & \underline{12.732} & 12.414 & 25.146 & \underline{196.546} \\
 & GFA-GCN & 0.294 & 0.206 & 0.146 & \underline{0.15} & \underline{0.114} & 0.08 & \underline{0.114} & \underline{0.097} & 0.08 & 13.104 & 12.682 & 25.786 & \textbf{\underline{196.1}} \\
 & ARVGA-GAT & \textbf{0.303} & \textbf{0.216} & \textbf{0.156} & 0.15 & \textbf{0.12} & \textbf{0.092} & \textbf{\underline{0.123}} & \textbf{0.102} & \textbf{0.092} & 12.884 & 12.632 & 25.516 & 196.97 \\
 & ARVGA-GIN & 0.295 & 0.207 & 0.147 & 0.154 & 0.114 & 0.082 & \underline{0.12} & \underline{0.101} & 0.082 & \textbf{12.554} & 12.312 & \textbf{24.866} & \underline{197.367} \\
 & ARVGA-GCN & 0.290 & 0.201 & 0.141 & 0.144 & 0.11 & 0.076 & \underline{0.115} & 0.093 & 0.076 & 13.32 & 12.88 & 26.2 & 199.406 \\
 & SCENIR & 0.29 & 0.201 & 0.14 & 0.128 & 0.098 & 0.07 & 0.1 & 0.088 & 0.07 & 13.06 & \textbf{\underline{12.188}} & 25.248 & 199.166 \\
\bottomrule
\end{tabular}
}
\end{table*}

Finally, we evaluate the  performance of  U-CECE-Inductive, utilizing all unsupervised GNN models that can intuitively be adapted to the inductive setup. As detailed in our methodology, the inductive regime offers a significant computational advantage: by pretraining on a massive unlabeled corpus, counterfactual identification is reduced to a single inference pass, eliminating the need for labels or task-specific training.
To this end, a detailed  evaluation of the U-CECE-Inductive framework across all unsupervised autoencoder architectures and their respective GNN backbones is presented in Table \ref{tab:inductive-detailed-summary}. 
Comparing these strict zero-shot inductive results with the previously detailed transductive baselines reveals a nuanced and encouraging trend: despite having never been updated on the target test topologies, the inductive models maintain highly competitive global ranking fidelity. For example, on the CUB dataset, the inductive ARVGA-GCN and SCENIR models achieve nDCG@4 scores of 0.325 and 0.320 respectively, effectively matching or slightly outperforming the transductive ones. This suggests that large-scale unsupervised pretraining allows the engines to learn a more global semantic manifold that generalizes well to entirely unseen queries.

However, removing the advantage of dataset-locked supervision exposes a natural trade-off regarding exact retrieval. 
In dense environments, domain-specific transductive supervision retains its edge in Top-1 precision.
Because the transductive engine maps explicit graph-pair distances during training, it is better equipped to isolate exact target instances within complex local neighborhoods. In contrast, the inductive models trade a few percentage points of peak $P@1$ for scalability. 
Yet, expanding the retrieval window slightly reveals some dominance for both variants of $P@k$ when $k>1$. While inductive models may occasionally miss the $@1$ ranking, they are more consistent at populating the top with more similar instances (better $P@k$), indicated by the density of underlined values across datasets and models.
The high top-$k$ precision pairs with the inductive framework's superior $\overline{GED}$ to highlight that without being forced to overfit to specific reference pairs, the generalized inductive space improves in taking noisy shortcuts compared to transductive autoencoders. Consequently, for the more structurally uniform CUB dataset, the inductive models are even able to yield counterfactuals with superior structural minimality (i.e. SCENIR with 10.369 $\overline{e}_{total}$ compared to the transductive 11.069). 


The effect of topological density on zero-shot generalization is highlighted by the performance deltas. For instance, SCENIR suffers a Top-1 precision penalty of -0.028 in VG-DENSE, but this penalty shrinks to -0.004 on VG-RANDOM. Inductive GFA-GIN and ARVGA-GIN not only almost eliminate the precision penalty on VG-RANDOM, but surpass their transductive counterparts in $\overline{GED}$ by 4.42 and 1.28 points respectively.
This performance profile can be directly attributed to the nature of the pretraining data; since the vast majority of the full VG dataset consists of sparse scene graphs, the inductive latent space is naturally biased toward such topological motifs.


While the L1 and L2 expressivity tiers do not require training and are inherently "inductive," they lack an optimized continuous latent space, making their exact graph-matching inference significantly slower. Thus, U-CECE-Inductive provides a unique and highly desirable combination: the high-expressivity and semantic richness of the L3 tier, coupled with rapid inference and generalization to entirely unseen graph topologies, bypassing the supervised training complexity bottleneck without compromising much explanatory integrity.


\subsubsection{Efficiency Analysis}

To assess computational efficiency, we evaluate the aggregate retrieval latency across all U-CECE engines. We define retrieval time as the total duration required to rank \textit{all} query images against the entire database and extract their counterfactuals. Naturally, the computational overhead varies inherently across branches. L1 and L2 expressivities require practically no preprocessing, whereas L3 expressivity relies on a pre-trained model. Consequently, for L3, retrieval time encompasses performing inference across all samples, followed by a conditional ranking based on classifier predictions using cosine similarity. Furthermore, the supervised variants of U-CECE-Structural necessitate computationally expensive GED label calculations for roughly half of all pairwise graph combinations before training, which is highly relevant to consider transductively. To ensure a fair comparison on the exact \textit{same} data, all models are benchmarked in a transductive setting. Evaluating in an inductive setting would introduce discrepancies strictly in training times, owing to the different graph structures and the different volume of data present in the inductive training set. Finally, to establish consistent and conservative time estimates, we report the performance of the GCN variant for the supervised GNNs and employ SCENIR for the unsupervised structural engine. As the most technically complex autoencoder, SCENIR effectively serves as an upper bound for unsupervised retrieval times.


\begin{wraptable}{r}{0.65\textwidth} 
\centering
\small 
\caption{Offline overhead (GED+training) and aggregate retrieval times across CUB, VG-DENSE, and VG-RANDOM datasets (sec).}
\begin{tabular}{lcccccc}
\toprule
& \multicolumn{2}{c}{\textbf{CUB}} & \multicolumn{2}{c}{\textbf{VG-DENSE}} & \multicolumn{2}{c}{\textbf{VG-RANDOM}} \\ \cmidrule(lr){2-3} \cmidrule(lr){4-5} \cmidrule(lr){6-7}
\textbf{Engine} & \textbf{Offline} & \textbf{Retr} & \textbf{Offline} & \textbf{Retr} & \textbf{Offline} & \textbf{Retr} \\ 
\midrule
L1 & - & 147 & - & 82& - & 179\\
L2 & - & 10654 & - & 4344 & - &4011 \\\hline
L3-super & \makecell{46220+\\32691} & 0.10 & \makecell{13982+\\12059} & 0.10 & \makecell{18787+\\16271} & 0.14 \\
L3-unsuper & 0+25.18& 0.52 & 0+27.26& 0.64 &0+29.10 & 0.59 \\ 
\bottomrule
\end{tabular}
\label{tab:efficiency_wrap}
\end{wraptable}

Table \ref{tab:efficiency_wrap} shows that L1 and L2 engines incur substantially higher retrieval times, with L1 being the smaller of the two as it encodes no relational information, evidence that relations impose a considerable computational overhead. L3 expressivities, by contrast, retrieve in under one second across all datasets. The supervised variant carries a significant offline cost, on the order of hours for GED computation and training; the unsupervised variants avoid this entirely, requiring no GED and training linearly in the number of graphs rather than over tens of thousands of graph pairs. At inference, however, L3-supervised is the more efficient of the two, owing to its single-layer GCN encoder. Across datasets, VG-DENSE is relation-rich but sparse in nodes, yielding the lowest L1 time; CUB's larger node counts combined with relations inflate L2 and L3 times, while the VG variants remain closer together. Trainable L3 models exhibit comparable retrieval times throughout, as they operate in latent space. SGCE's lower retrieval times are attributed to more efficient batching. In terms of GPU cost, L3-supervised (SGCE) has ~600K parameters while L3-unsupervised (SCENIR) has ~650K parameters at training time but uses roughly half at inference (encoder-only). Both their memory footprints range around 1GB, with SGCE's increasing during training due to graph pair formatting, not the architecture itself. 
These measurements characterise the cost regimes at the scales we evaluate rather than tracing the asymptotic behaviour of the Equations in Section \ref{sec:complexity_analysis} directly; they are intended to give a concrete reference point for deployment decisions at comparable dataset sizes.

\subsubsection{Qualitative Analysis}
\label{sec:qualitative}

\begin{figure*}[ht!]
    \centering
    \includegraphics[width=0.92\textwidth]{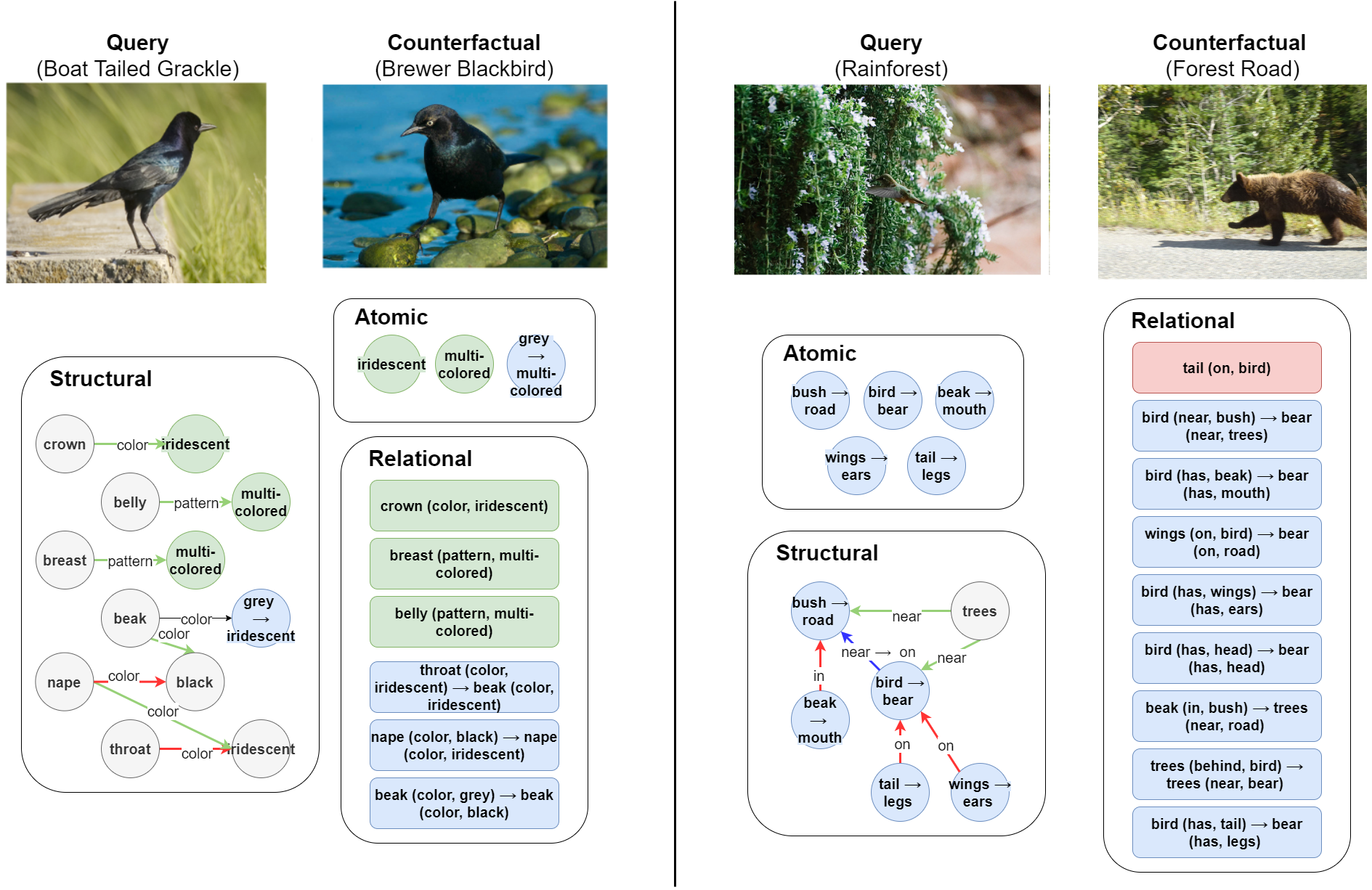}
    \caption{U-CECE expressivity tiers yielding the same counterfactual for different datasets (\textbf{left:} CUB example, \textbf{right:} VG-DENSE example).}
    \label{fig:qual_diverse_cases}
\end{figure*}

While the previous sections establish a quantitative standard for retrieval, the primary goal of U-CECE is to provide human-interpretable explanations. In this section, we transition from numerical metrics to a visual inspection of the counterfactuals generated at each expressivity tier. For the Structural level, we utilize the SGCE-GCN variant, as it consistently demonstrated the highest retrieval fidelity. By visualizing these transitions, we aim to evaluate how the mathematical minimality captured by different types of GED approximations aligns with logical, intuitive shifts in conceptual reasoning. In all figures, green indicates insertion, red deletion and blue replacement. Grey nodes are context nodes, that do not get edited.

Figure \ref{fig:qual_diverse_cases} provides a qualitative visual bridge to our earlier quantitative findings, illustrating how the three expressivity tiers interpret the same counterfactual transition. In the CUB example (Left), the L1 tier offers a general sense of the necessary attribute shifts (e.g., color changes) but lacks grounding, as it cannot specify which anatomical parts these attributes refer to. The L2 level serves as a "middle man," adding this missing context; however, it occasionally introduces confusion during replacements. For instance, the transition of "throat (color,iridescent)" to "beak (color,iridescent)" suggests a peculiar "transfer" of properties between parts rather than a logical state change. In contrast, the L3 tier provides the full conceptual picture: it clarifies that attributes are being added or modified on specific, existing parts without misrepresenting the parts themselves, accurately reflecting that black color is added to the beak while iridescent tones are mapped to the nape. This aligns with our observation that while the L3 tier may have higher raw edit counts ($\bar{e}_{total}$), its semantic fidelity holds because it avoids potential logical shortcuts taken by the lower tiers. For the VG-DENSE sample (Right), the limitations of the L2 tier become more pronounced. Its triplet-based replacements, such as "beak (in,bush)" transitioning to "tree (near,road)", can be difficult to decode when relational density is high. The Structural subgraph, however, explicitly maps the logic required to transition from  "Rainforest"$\rightarrow$"Forest Road": it swaps the bird for a bear (an entity more intrinsic to the target environment) and physically introduces a road where greenery previously existed. The inclusion of extra edges in the L3 tier refines the conceptual hierarchy (e.g., increasing tree density), consistent with L3 achieving the highest $P@1$ in dense settings. Interestingly, the L1 tier remains surprisingly useful for single-concept transitions, reinforcing our quantitative finding that often simple concept-based retrieval is a low-cost alternative for broad conceptual shifts.

\begin{figure}[t!]
    \centering
    \includegraphics[width=0.93\textwidth]{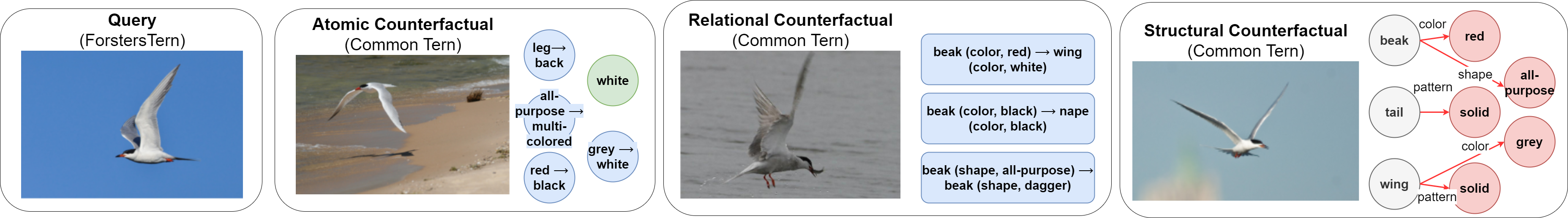}
    \caption{Qualitative comparison of diverging counterfactuals retrieved by U-CECE expressivity tiers for CUB example (same $\mathsf{Common Tern}$ target class). }
    \label{fig:qual_tiers_tern}
\end{figure}

In cases where the different expressivity tiers retrieve distinct counterfactual images for the same target class (Figure \ref{fig:qual_tiers_tern}), we gain further insight into the specific "reasoning" of each engine.
While all three retrieved counterfactuals are semantically valid representatives of the $\mathsf{Common Tern}$ target class, the L3 engine identifies a candidate that is arguably dominant in terms of pose and visibility alignment with the original query. 
The structural explanation provides a highly straightforward narrative for the transition: to move from a $\mathsf{Forsters Tern}$ to a $\mathsf{Common Tern}$, one must eliminate the "solid" pattern and specifically reduce the prominence of "red" in the beak. While this core logic is partially reflected in the lower tiers, most notably in the L1 tier's identification of a "red-to-black" shift, the L3 tier provides the most refined and anatomically consistent roadmap. This reinforces our earlier quantitative findings: while the L1 tier is consistent at finding a correct bird through general attribute matching, in this case,  the L3 tier succeeds in selecting the bird that preserves the query's underlying graph-based context, even if it was not the reference sample.

\begin{figure}[t!]
    \centering
    \includegraphics[width=0.93\textwidth]{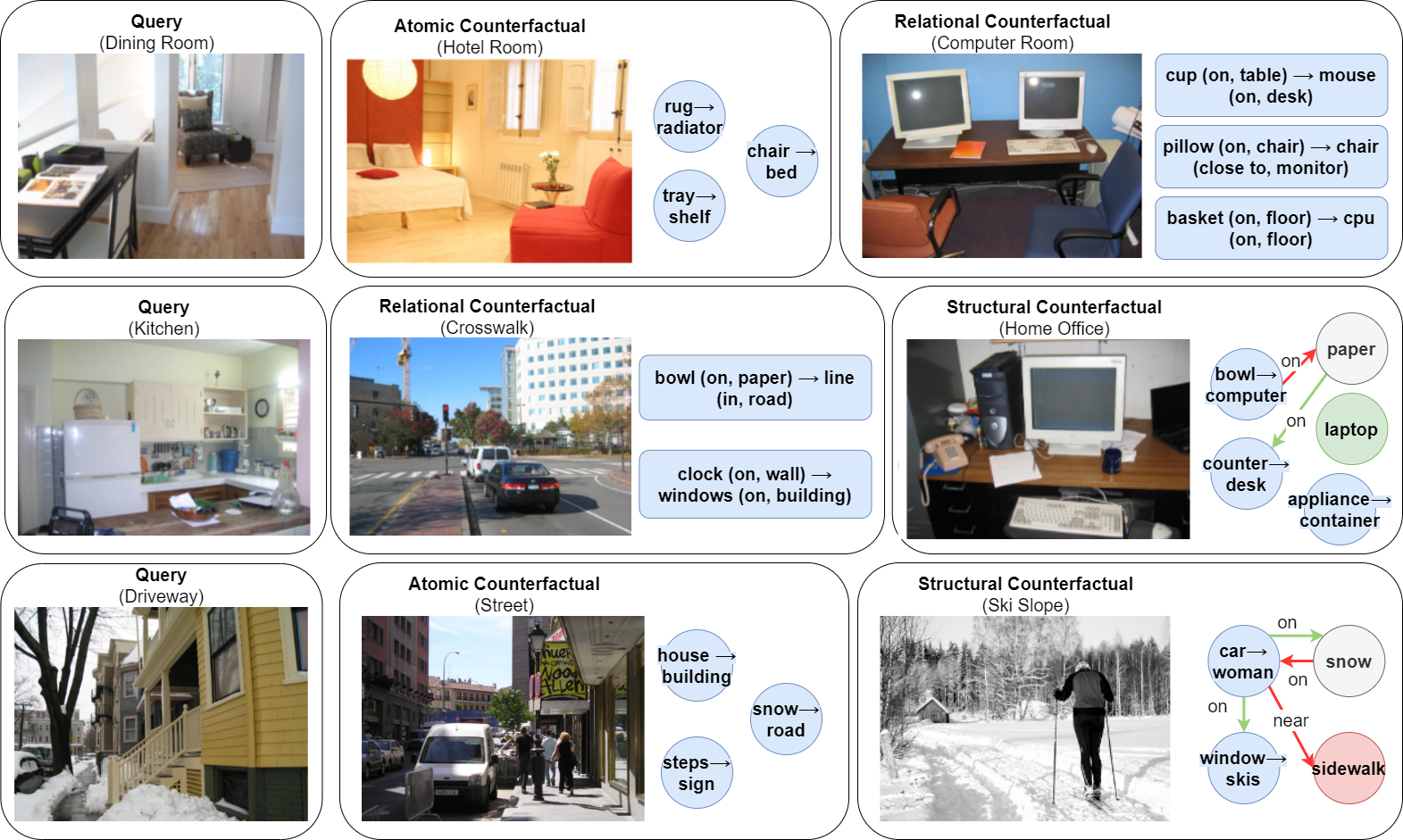}
    \caption{Qualitative comparison of diverging counterfactuals retrieved by U-CECE expressivity tiers for VG datasets (pairwise comparisons). A subset of edits is depicted for clarity purposes.}
    \label{fig:quals_by_pair_vg}
\end{figure}

In Figure \ref{fig:quals_by_pair_vg}, we present pairwise comparisons on VG to highlight the specific logic of each expressivity mode. For clarity, we focus on a subset of representative replacements rather than displaying the complete edit graphs as in previous sections. 
Row 1 compares L1 vs. L3 mode. In sparse VG-RANDOM scenes like the Dining Room, where annotations often include isolated nodes (e.g. $\mathsf{rug}$), the L1 tier suggests broad conceptual swaps like $\mathsf{chair}\rightarrow \mathsf{bed}$ to retrieve a Hotel Room. These edits provide a general category shift but lack topological grounding. In contrast, the L2 tier utilizes triplet context to identify objects with similar scene interactions, such as swapping a $\mathsf{basket}$ $\mathsf{on}$ $\mathsf{floor}$ for a $\mathsf{CPU}$ $\mathsf{on}$ $\mathsf{floor}$, thereby retrieving a Computer Room that feels more situationally grounded. Row 2 highlights the merit of structural expressivity in sparse scenarios. In the cluttered Kitchen query, the limitations of triplet-based retrieval become evident. Since the L2 tier over-indexes on specific triplets while ignoring the numerous isolated nodes typical of cluttered scenes, it suffers from semantic drift, matching a $\mathsf{bowl}$ $\mathsf{on}$ $\mathsf{paper}$ to a $\mathsf{line}$ $\mathsf{in}$ $\mathsf{road}$ and retrieving an irrelevant $\mathsf{crosswalk}$. The L3 engine, however, accounts for both interconnected and isolated nodes. It successfully identifies a Home Office as a semantically closer counterfactual, logically swapping the $\mathsf{counter}$ for a $\mathsf{desk}$ to preserve the "indoor tabletop" nature of the scene. Finally, row 3 compares L1 vs. L3 mode in relationally denser scenes (like the depicted snow scene from VG-DENSE). While the L1 tier simply swaps $\mathsf{house}$ for $\mathsf{building}$, it fails to capture how the depicted objects interact with the pervasive snow. The L3 engine re-purposes this interaction: it replaces the snow-covered $\mathsf{car}$ with a $\mathsf{woman}$ and her $\mathsf{skis}$ who is also interacting with the $\mathsf{snow}$. This transition illustrates an example where the L3 tier does not just list presence; it refines the conceptual hierarchy to reflect the environmental truth, consistent with its stronger agreement with the reference on dense-relational benchmarks.




\section{Testing Human Perception on U-CECE counterfactuals}
\label{sec:human}

While previous experiments and analysis address explanation minimality, retrieval quality, and engine efficiency, a systematic evaluation of the explanations themselves to the final recipients is required. 
This is particularly vital for the L3 variant which, as the most expressive tier by design, prioritizes feature interdependence.
To justify the selection of the L3 variant for our human evaluation on the CUB dataset - despite lower-expressivity tiers occasionally showing higher $P@1$ scores - we anchor our study in the proven pedagogical efficacy of this specific engine: as demonstrated in  \citet{dimitriou2023structure}, the structural representations produced by SGCE are exceptionally robust for machine teaching tasks, with human participants successfully learning to categorize complex species based exclusively on the structural edit subgraphs, even in the complete absence of visual imagery during the teaching phase. By extending these findings, our current human perception experiments evaluate whether the neural-based retrieval engine serves as a semantically sensible alternative to the reference GED. This is a vital distinction: while  GED provides a deterministic reference, the GNN is trained on thousands of pairs to internalize a latent manifold that represents the conceptual essence of a "Bird". Consequently,  these human trials aim to verify that even when the engine's top-ranked retrieval deviates from the reference GED, it remains within the same semantic neighborhood, preserving the logical integrity and explanatory power required for effective human-AI collaboration.

\subsection{Design of Human Survey}
We design three distinct survey formats to isolate different aspects of human judgment. Consistent with established visual XAI research \citep{vandenhende2022making, dervakos2023choose, dimitriou2023structure}, we select the CUB dataset for bird species classification, to have humans evaluate the generated explanations. This selection serves as a control for prior knowledge: by employing non-expert annotators, we ensure that similarity assessments are based on the provided visual counterfactuals rather than pre-existing taxonomical expertise. To ensure participants evaluate high-level semantic concepts rather than low-level image artifacts, they are provided with the following standardization criteria:
\begin{tcolorbox}[
    colback=pink!8,
    colframe=pink!45!black,
    fonttitle=\bfseries,
    sharp corners,
    boxrule=0.6pt,
    left=4pt,
    right=4pt,
    top=4pt,
    bottom=4pt
] \small
    "\textit{Assess whether the images are semantically equivalent by concentrating exclusively on the species and characteristics of depicted birds. Ignore any other elements, such as the surroundings, pose, or position of the birds}."
    \end{tcolorbox}
The survey experiments are: i) \textbf{Select}: Participants are shown the source image alongside both the reference (GED) and the retrieved (GNN) counterfactuals. They are asked to identify which of the two images is more appropriate as a counterfactual, or if they are conceptually equal given the source; ii) \textbf{Sem-Eq-Triplets}: Given the same source-GED-GNN setup, participants are asked if the two counterfactuals are semantically equivalent in the context of the source; iii) \textbf{\textbf{Sem-Eq-Pairs}:} To assess baseline semantic consistency without the influence of the "counterfactual task," participants evaluate the equivalence of the two counterfactual images in isolation. For brevity, we refer to the latter two together as \textbf{Sem-Eq-x} in our analysis.
\begin{figure}[t!]
    \centering
    \begin{minipage}[b]{0.48\textwidth}
        \centering
        \includegraphics[width=\linewidth]{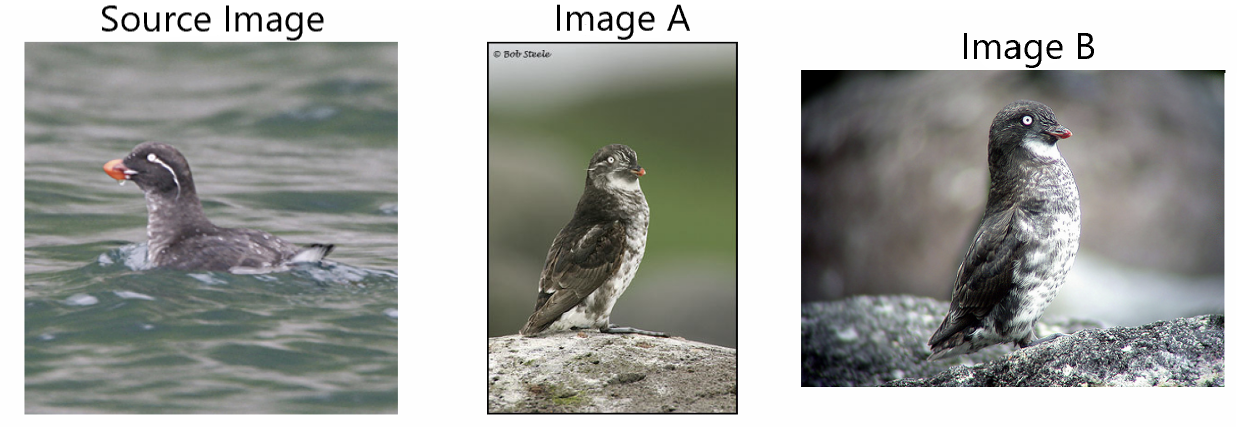}
        \caption*{$\mathsf{Auklet}$ classes ($\mathsf{Parakeet}$ $\to$ $\mathsf{Least}$)} 
    \end{minipage}
    \hfill 
    \begin{minipage}[b]{0.48\textwidth}
        \centering
        
        \includegraphics[width=\linewidth]{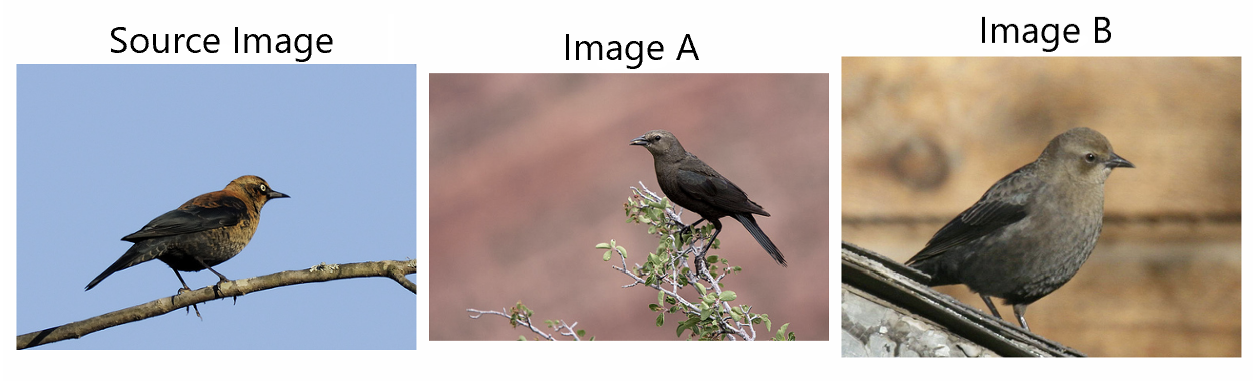}
        \caption*{$\mathsf{Blackbird}$ classes ($\mathsf{Rusty}$ $\to$ $\mathsf{Brewer}$)} 
    \end{minipage}
    
    \caption{Examples of visual form layouts showing two triplets of images used in the experiments. Each shows (Source, Image A (reference GED), Image B (U-CECE-Structural w/ GCN)).
    As noted, in the \textbf{Sem-Eq-Pairs} experiment, the source image was not provided, nor was it revealed that images A, B were alternative counterfactuals for a specific query.}
    \label{fig:select-screenshots_combined}
\end{figure}

Each survey comprises 30 randomly selected instances from two specific counterfactual class transitions: $\mathsf{Rusty Blackbird}\rightarrow \mathsf{Brewer Blackbird}$, and $\mathsf{Parakeet} \rightarrow \mathsf{Least Auklet}$. These pairs are selected based on prior claims that they are widely confused by image classifiers during explainability tasks \citep{vandenhende2022making}. For each instance, we gather multiple independent human annotations ($35$ for Select, $36$ for Sem-Eq-Triplets, and $28$ for Sem-Eq-Pairs). The visual layout of these forms is illustrated in Figure \ref{fig:select-screenshots_combined}, while comprehensive details regarding participant information and additional examples are available in App. \ref{app:human}.

To transform subjective survey responses into mathematically grounded data, we move beyond simple majority voting. Specifically, we apply a Chi-Square Goodness-of-Fit Test \citep{categorical-analysis} to every question. This allows us to segment our results into two distinct "buckets": i) Signal, i.e. questions where the null hypothesis (random guessing) is rejected ($p < 0.05$)  representing cases where human intent is clear and consensus is statistically significant, and ii) Noise, i.e. questions where responses are statistically indistinguishable from random noise ($p \geq 0.05$), which identify edge cases where semantic differences are too subtle or subjective for universal agreement. The results of these surveys are presented below.

Assessing the semantic similarity of bird instances proved challenging, with approximately 1/3 of the instances exhibiting high inter-rater variance (noise). Notably, our Sem-Eq-x experiments demonstrate high consistency across annotators, with 25 out of 30 instances categorized the same way. 
This overlap hints that the inclusion of the source image does \textbf{not} fundamentally alter the perceived semantic relationship between counterfactuals. In contrast, agreement diminishes in the \textbf{Select} experiment, likely because the forced-choice format introduces subjective variance absent in binary tasks. Ultimately, 11 instances remain consistent across all setups, while only 4 are entirely excluded from the primary analysis due to noise.

\begin{figure*}[t!] 
    \centering
    
    \begin{subfigure}[b]{0.35\textwidth}
        \centering
        \includegraphics[width=\textwidth]{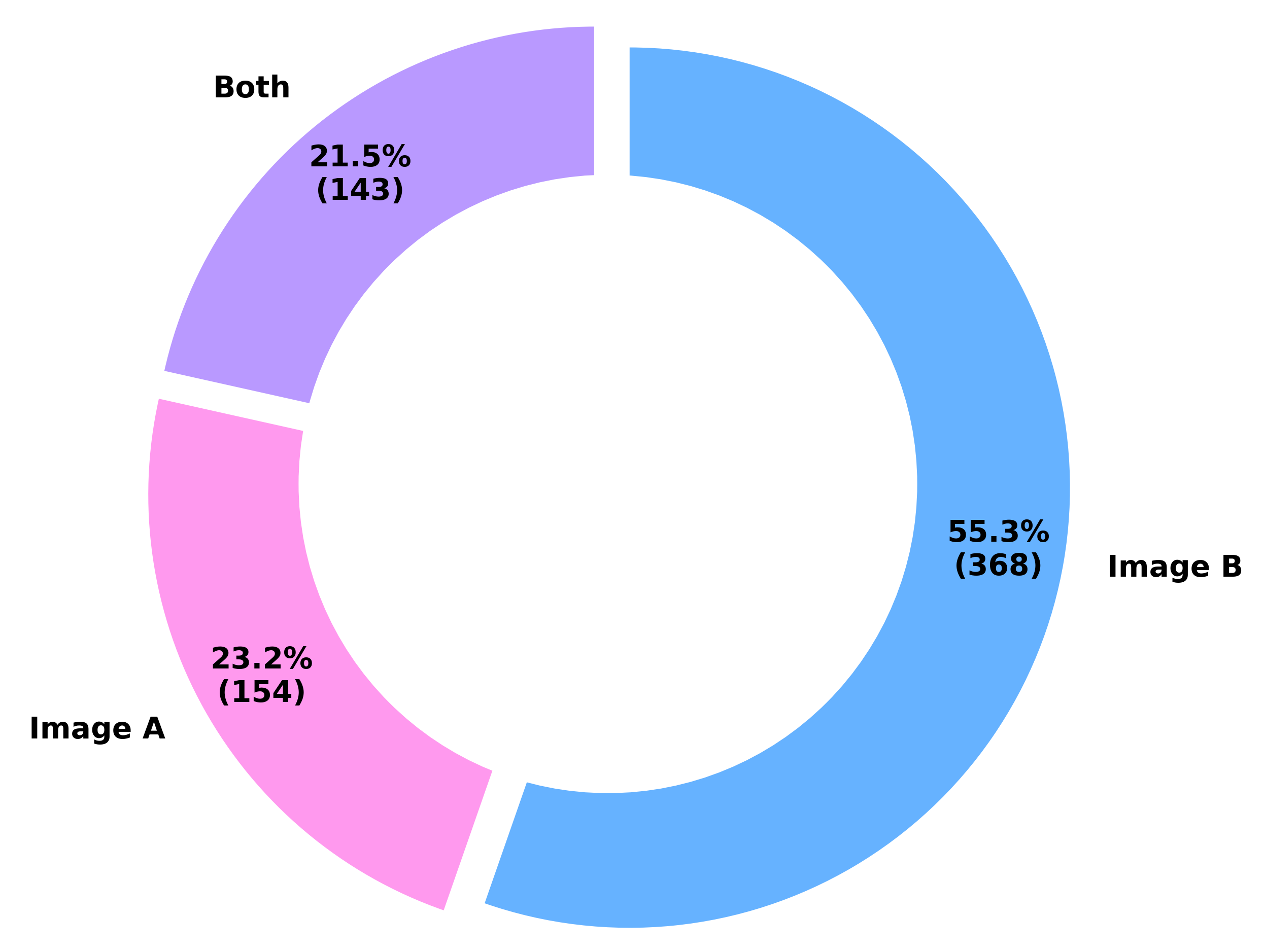}
        \caption{Select Exp: Overall}
        \label{fig:select-all}
    \end{subfigure}
    \hfill
    \begin{subfigure}[b]{0.32\textwidth}
        \centering
        \includegraphics[width=\textwidth]{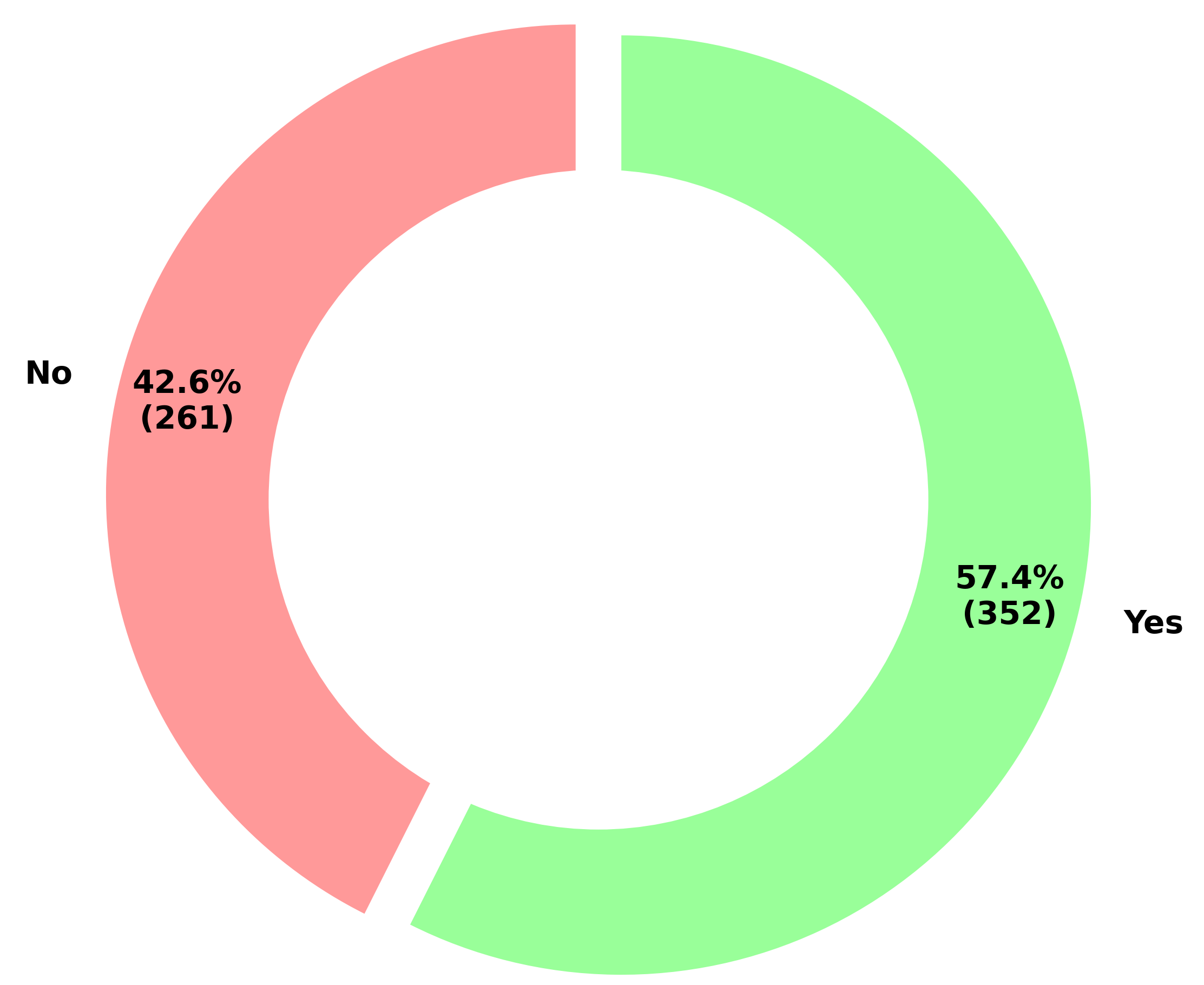}
        \caption{Triplets Exp: Overall}
        \label{fig:triplets-all}
    \end{subfigure}
    \hfill
    \begin{subfigure}[b]{0.3\textwidth}
        \centering
        \includegraphics[width=\textwidth]{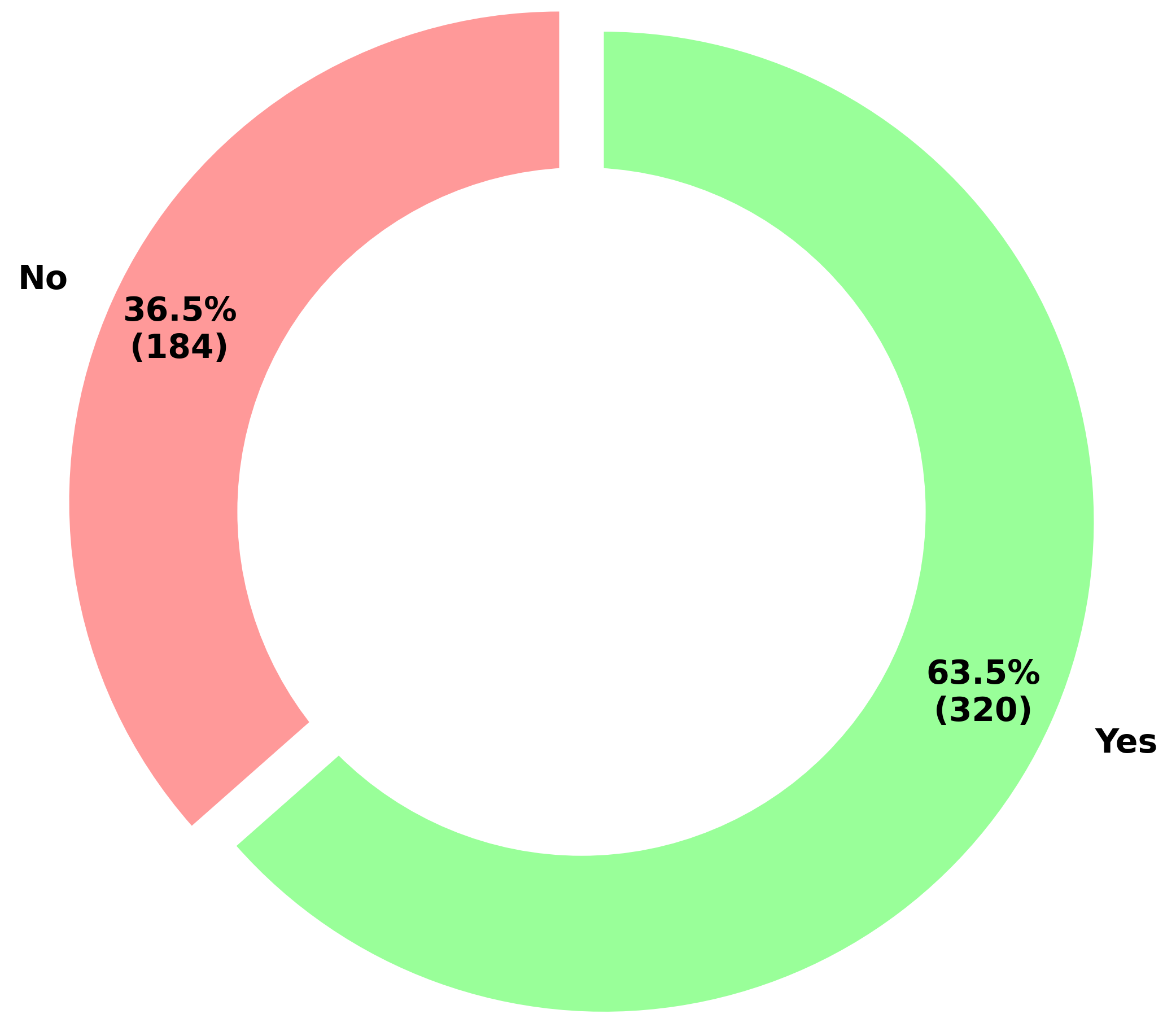}
        \caption{Pairs Exp: Overall}
        \label{fig:pairs-all}
    \end{subfigure}
    
    \vspace{0.5cm} 
    
    \begin{subfigure}[b]{0.32\textwidth}
        \centering
        \includegraphics[width=\textwidth]{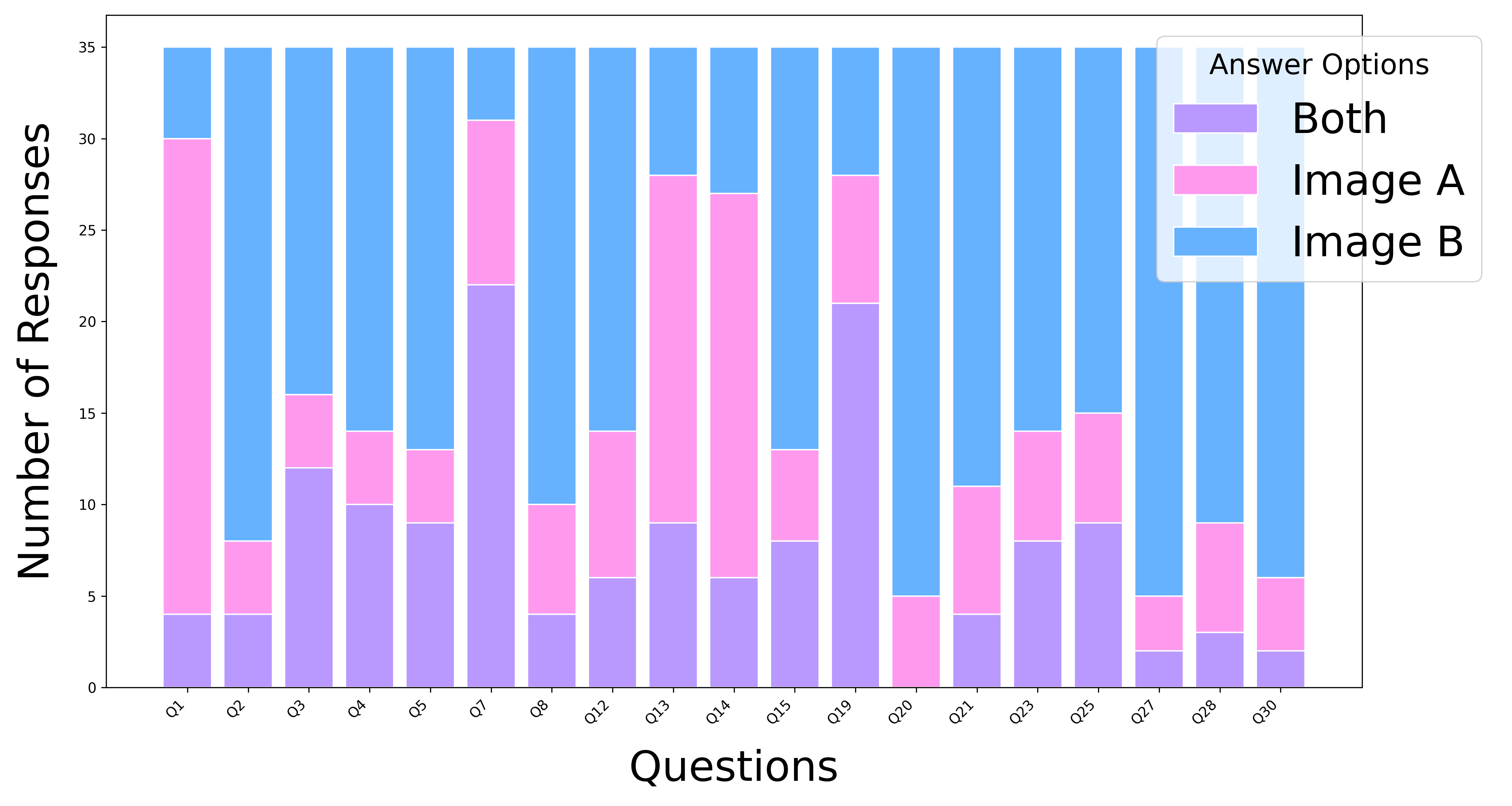}
        \caption{Select Exp: Per Question}
        \label{fig:select-qs}
    \end{subfigure}
    \hfill
    \begin{subfigure}[b]{0.32\textwidth}
        \centering
        \includegraphics[width=\textwidth]{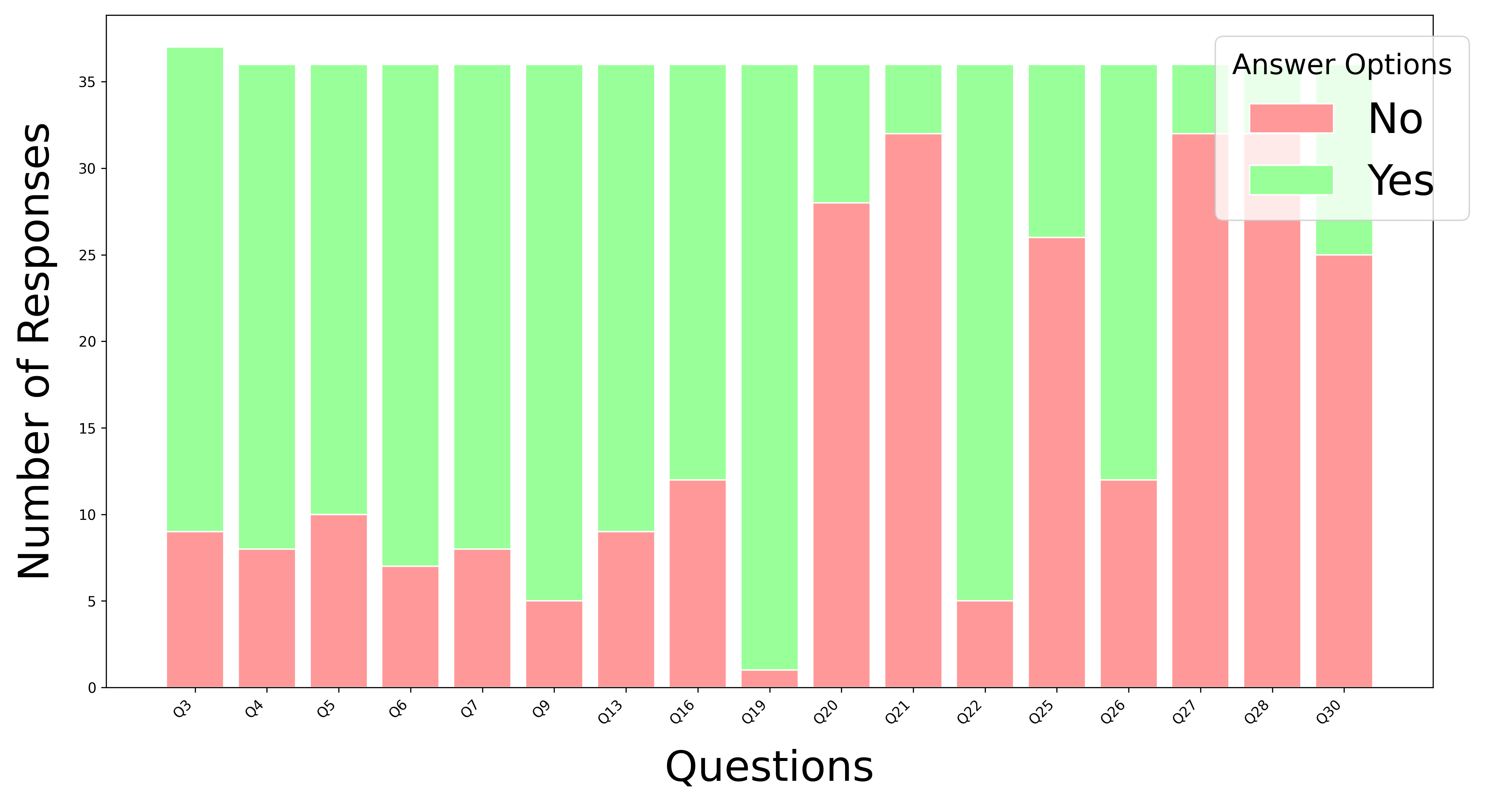}
        \caption{Triplets Exp: Per Question}
        \label{fig:triplets-qs}
    \end{subfigure}
    \hfill
    \begin{subfigure}[b]{0.32\textwidth}
        \centering
        \includegraphics[width=\textwidth]{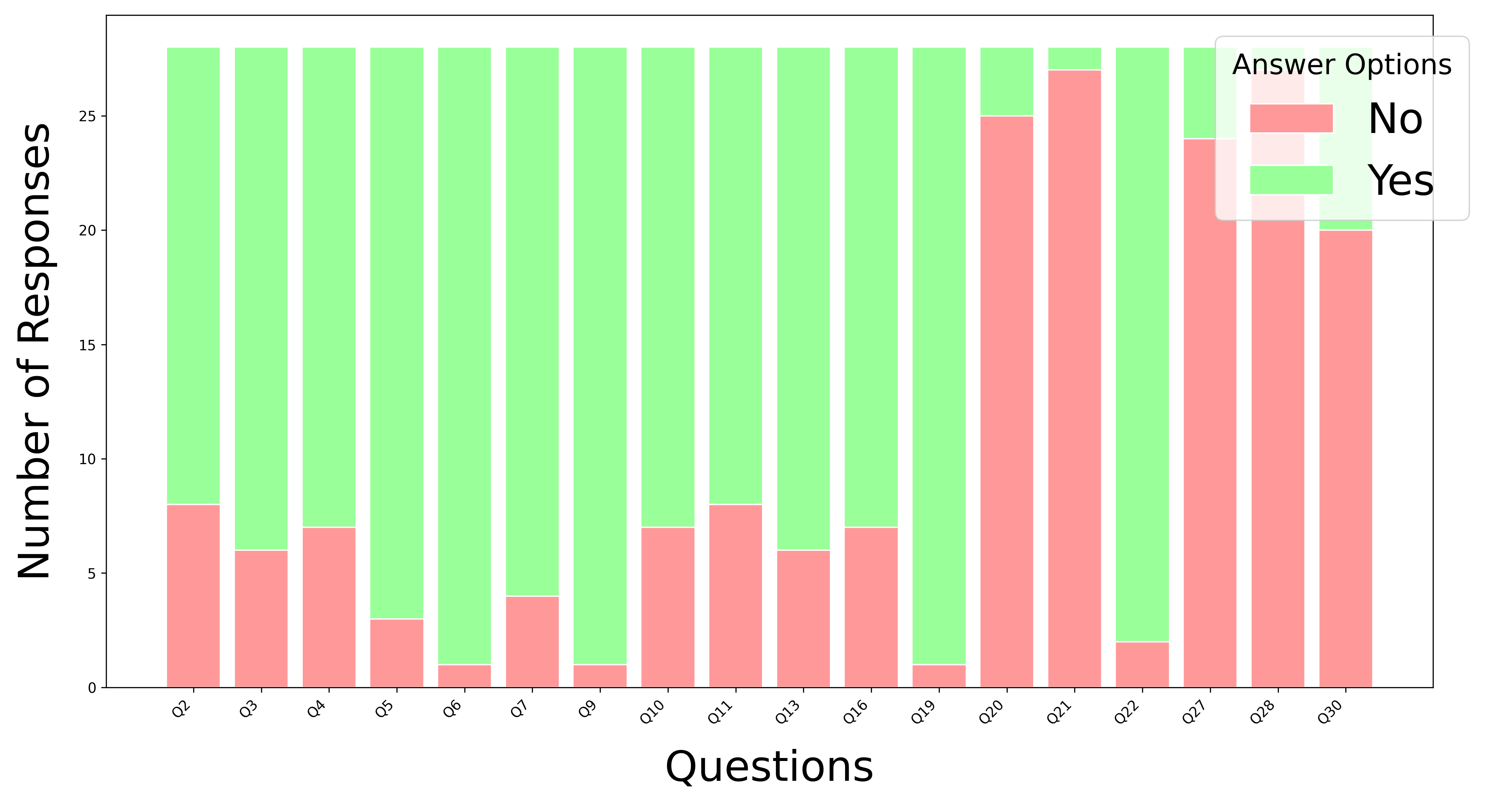}
        \caption{Pairs Exp: Per Question}
        \label{fig:pairs-qs}
    \end{subfigure}
    
    \caption{Human evaluation results (signal dataset). The top row displays the aggregate consensus for each experiment, while the bottom row details the distribution for each individual question. 
    }
    \label{fig:all-experiments}
\end{figure*}

The quantitative results of the human evaluation study are illustrated in Figure \ref{fig:all-experiments}. To focus on mathematically validated human consensus, these visualizations present the results from the \textit{signal dataset} ($p < 0.05$). Corresponding analysis detailing the high-entropy 'Noise' dataset are provided in Appendix \ref{app:human-noise}.

The results on signal data indicate that while the GNN(GCN)-retrieved counterfactuals are suboptimal under the reference GED, human evaluators predominantly perceive them as semantically preferable over the reference GED counterfactuals. First of all, the \textbf{Select} experiment demonstrates that the GCN-based retrieval is frequently preferred over the reference GED counterfactual. 
The GCN-generated counterfactuals are preferred in $55.3\%$ of responses, compared to only $23.2\%$ for the reference GED baseline (Figure \ref{fig:select-all}). Interestingly, the cognitive complexity introduced by comparing candidates against a third source image actually appears to favor GCN-retrieved instances. While this does not invalidate GED as a rigorous, deterministic evaluation criterion for structural minimality, it highlights a critical distinction in human-centric explainability: as the GCN projects structural topology and rich GloVe features into a continuous latent space, it performs a form of "semantic smoothing." This enables the neural network to construct holistic instance representations that blend semantics and structure, which may capture visual patterns not reflected in discrete edit costs. Ultimately, 
  \begin{wrapfigure}{r}{0.475\textwidth}
    \centering
    \includegraphics[width=\linewidth]{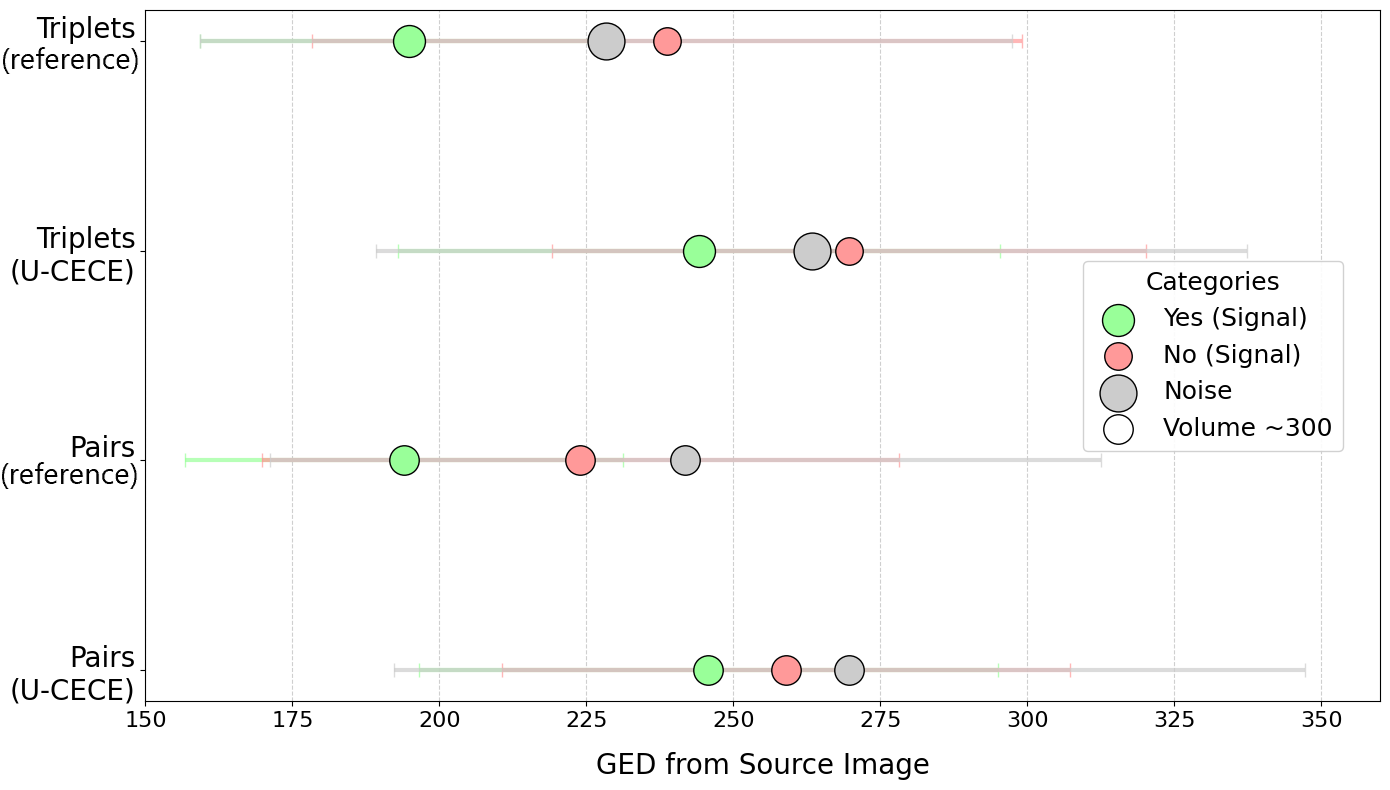}
    \caption{Distribution of GED across the Sem-Eq-x human perception surveys. Point markers represent the mean GED from the source image for reference and U-CECE (GCN) counterfactuals, categorized by participant consensus: \textit{Yes} (Signal), \textit{No} (Signal), and \textit{Noise}. Error bars denote standard deviation. The area of each marker is proportional to the total volume of user annotations in that specific category.}
    \label{fig:ged_comparisons_one_ax}
\end{wrapfigure}
this suggests that, in this setting, continuous neural representations can align more closely with human visual intuition than discrete edit costs.
Examples of cases where humans preferred U-CECE counterfactuals over reference GED are available in Appendix \ref{app:human-ged-discrepancy}.
Furthermore, in the \textbf{Sem-Eq-Triplets} setup, where the source image is present as a reference, $57.4\%$ of respondents affirm GCN equivalence (Figure \ref{fig:triplets-all}). Notably, in the \textbf{Sem-Eq-Pairs} experiment, which removes the source image, consensus on GCN equivalence increases to $63.5\%$ (Figure \ref{fig:pairs-all}). We interpret this by assuming that the cognitive load of triangulating visual similarities across three distinct images likely reinforces perceptual ambiguity. Instead, when evaluated in isolation, the retrieved and reference images are more readily recognized as sharing the same conceptual essence, suggesting that, within this CUB evaluation, the perceived semantic similarity between the retrieved and reference counterfactuals does not depend on presenting the source image. Crucially, these findings support the L3 (Structural) expressivity of U-CECE. The observed visual-equivalence judgments provide evidence that the GCN-based approximation remains semantically close to the reference GED retrievals in this CUB setting.
Finally, an analysis of the per-question distributions (Figures \ref{fig:triplets-qs}, \ref{fig:pairs-qs}) reveals a concentrated cluster of negative responses in the latter trials, which predominantly featured $\mathsf{Blackbird}$ classes. We hypothesize that the uniform, prominent black feathering of these instances made semantic assessment uniquely challenging, forcing non-expert participants to rely on irrelevant features, such as size or pose, to artificially differentiate the images.

To complement our behavioral analysis, we contextualize the human survey (Sem-Eq-x) results 
by comparing categorical responses to the distribution of pre-computed GEDs for both reference and GCN-retrieved counterfactuals (Figure \ref{fig:ged_comparisons_one_ax}). We aggregate responses from the noise dataset, where no statistical consensus is reached, as a baseline for semantic ambiguity.
This fine-grained analysis of the human experiments maps subjective  perception to a rigid structural metric, offering insight into what semantic preference entails for non-expert observers.
As expected, the GCN-retrieved results are mathematically suboptimal under the reference, yielding larger mean GED  compared to the reference across all categories. However, a clear pattern emerges within the GCN-retrieved group: instances where humans reached a "Yes" consensus consistently exhibit lower average GEDs than those labeled "No". This suggests that even when the GCN fails to find the reference minimum, its successful approximations (as perceived by humans) still reside in a tighter structural neighborhood than its failures.

The relationship between structural distance and human judgment shifts slightly depending on the experimental context. In the \textbf{Sem-Eq-Triplets} experiment, the noise category (represented by the grey marker with the highest inter-rater variance) sits squarely between the "Yes" and "No" signals, highlighting a zone where the structural distance from the source seems ambiguous. In contrast, in the \textbf{Sem-Eq-Pairs} experiment, the noise category is pushed further to the right, implying that without the source image as a reference, humans completing this survey require a higher degree of mathematical similarity to affirm semantic equivalence or not.

It is crucial to note that while the GCN acts as a "bird expert" capturing fine-grained structural cues, we assume that the participants (non-experts) likely utilized a coarser semantic filter. The relatively tight standard deviation (horizontal bars) for "Yes" responses in the reference categories suggests that when structural minimality is high, even non-experts provide highly consistent semantic labels, whereas the GNN's increased variance in the "No" and "Noise" regions reflects the difficulty of maintaining semantic coherence as structural distance grows.

\subsubsection{Human-LVLM Perception Alignment}

Foundational Models such as Large Vision-Language Models (LVLMs) have demonstrated impressive capabilities in visual reasoning tasks, even acting as judges prompted for evaluating vision-language settings  \citep{Chen2024MLLMasaJudgeAM}. In our case, we consider LVLMs to be an alternative to human evaluators, exploring their potential to obtain additional insights regarding our retrieved counterfactuals. To this end, we repeat the questions given to humans as prompts, explicitly formulated to ensure proper LVLM responses. We experiment with two cases: first, we perform a single-step prompting process, directly equivalent to the evaluation that humans performed. In the second case, we perform the reasoning-enhanced analyze-then-judge procedure proposed by \cite{Chen2024MLLMasaJudgeAM}, which first assists the LVLM into understanding the images under consideration, and then perform any querying regarding them as the next step.
Since LVLMs may hallucinate, we repeat the exact same evaluation experiment 10 times, simulating each LVLM session as a separate annotator, and keep the most frequent value for each answer choice. However, we observe that the used LVLM perfectly replicates its answers across runs, indicating confidence about its answers.

\begin{figure}[ht!]
    \centering
    \begin{minipage}{0.3\textwidth}
        \centering
        \includegraphics[width=0.87\textwidth]{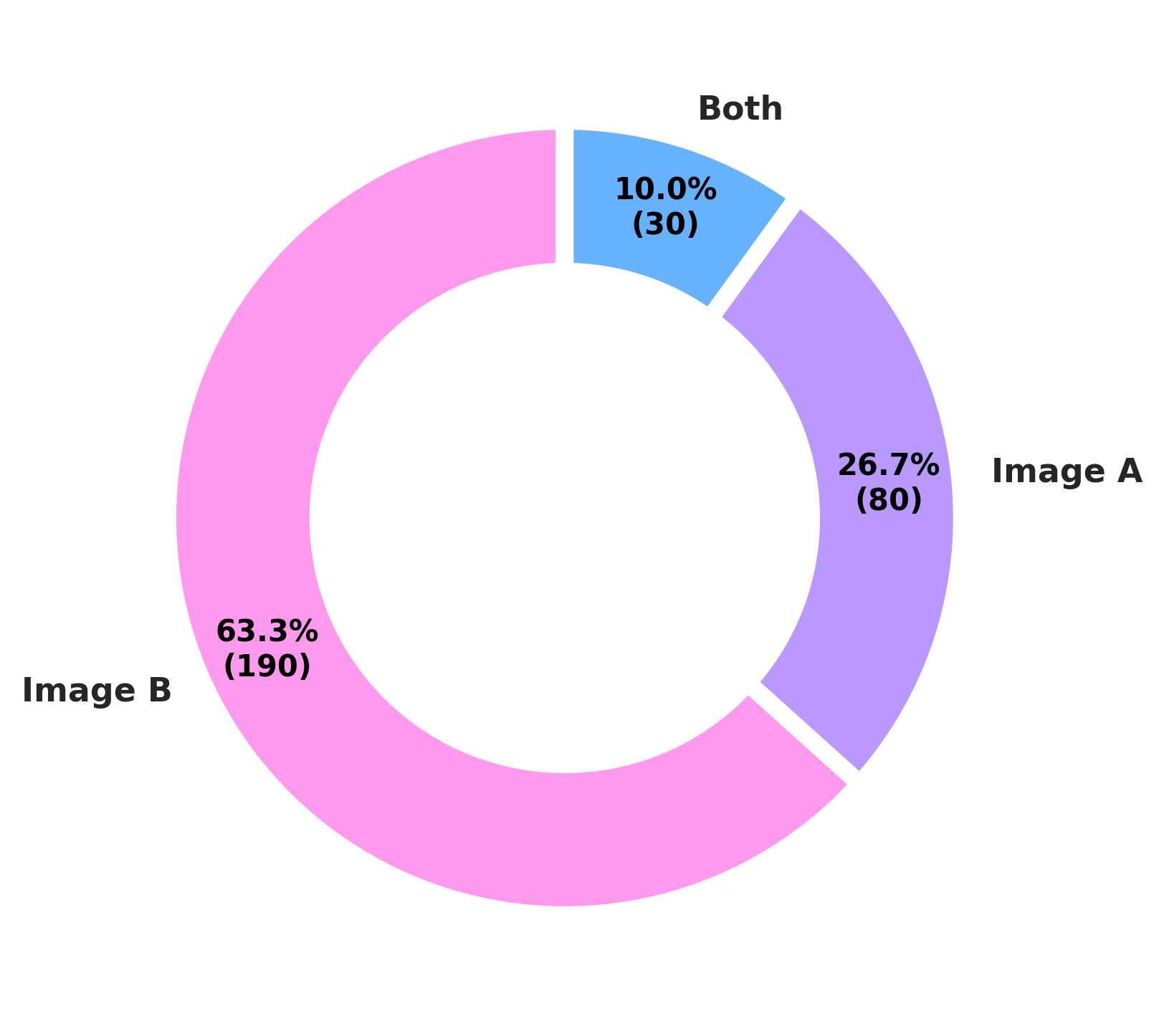}
    \end{minipage}%
    \hspace{0.01\textwidth} 
        \begin{minipage}{0.3\textwidth}
        \centering
        \includegraphics[width=0.8\textwidth]{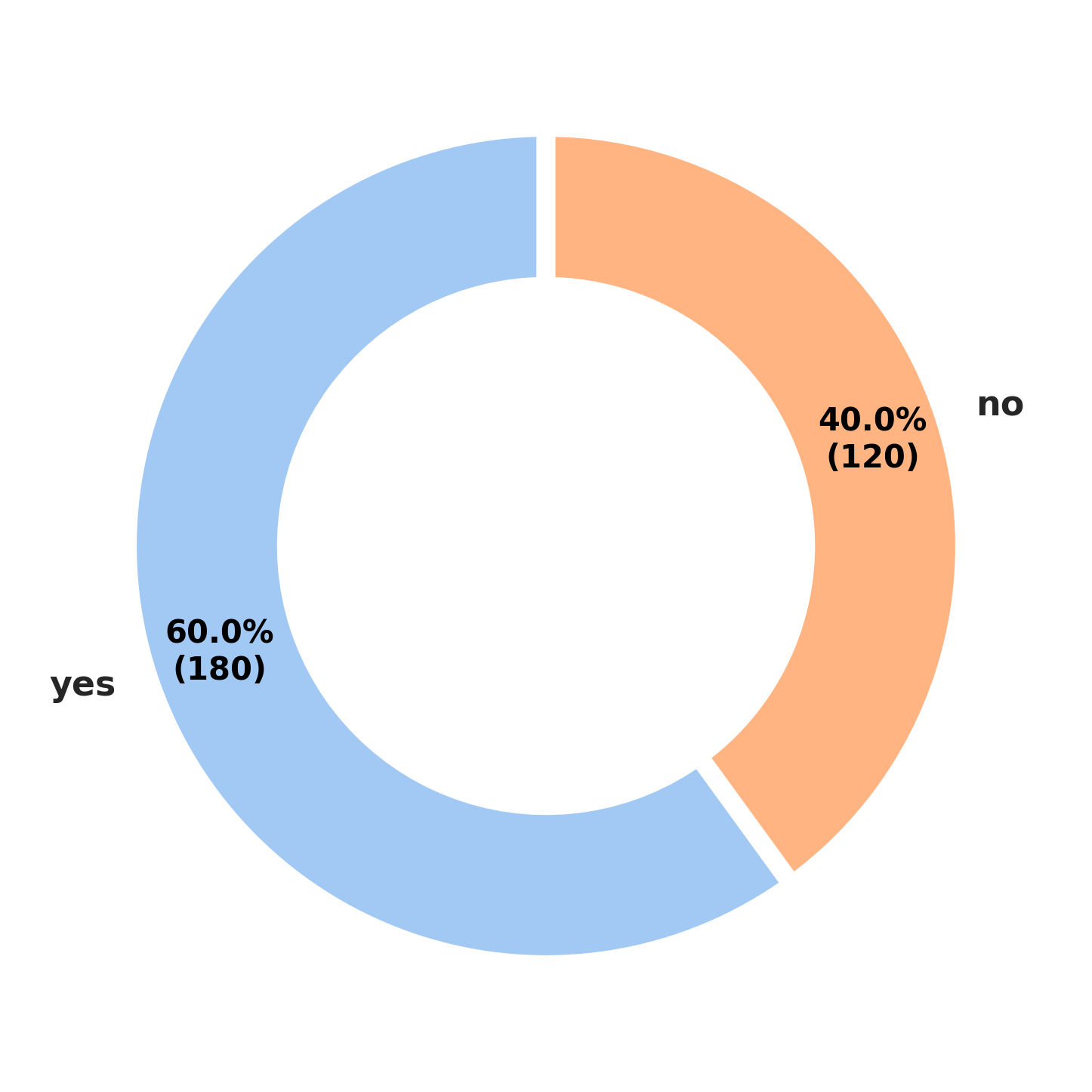}
    \end{minipage}
    \begin{minipage}{0.3\textwidth}
        \centering
        \includegraphics[width=0.8\textwidth]{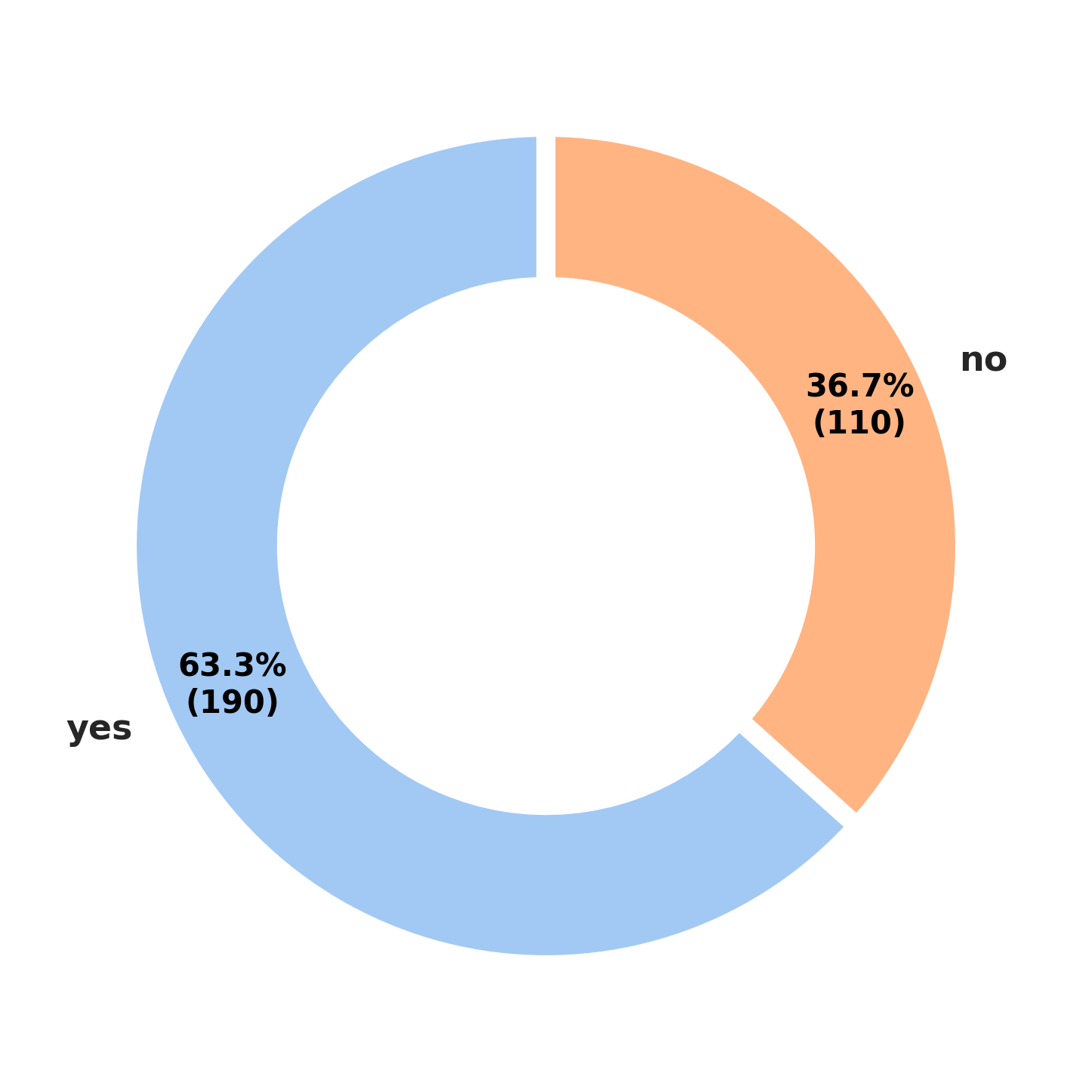}
    \end{minipage}%
    \hspace{0.01\textwidth} 

    \vspace{0.01\textwidth} 

    \begin{minipage}{0.3\textwidth}
        \centering
        \includegraphics[width=0.9\textwidth]{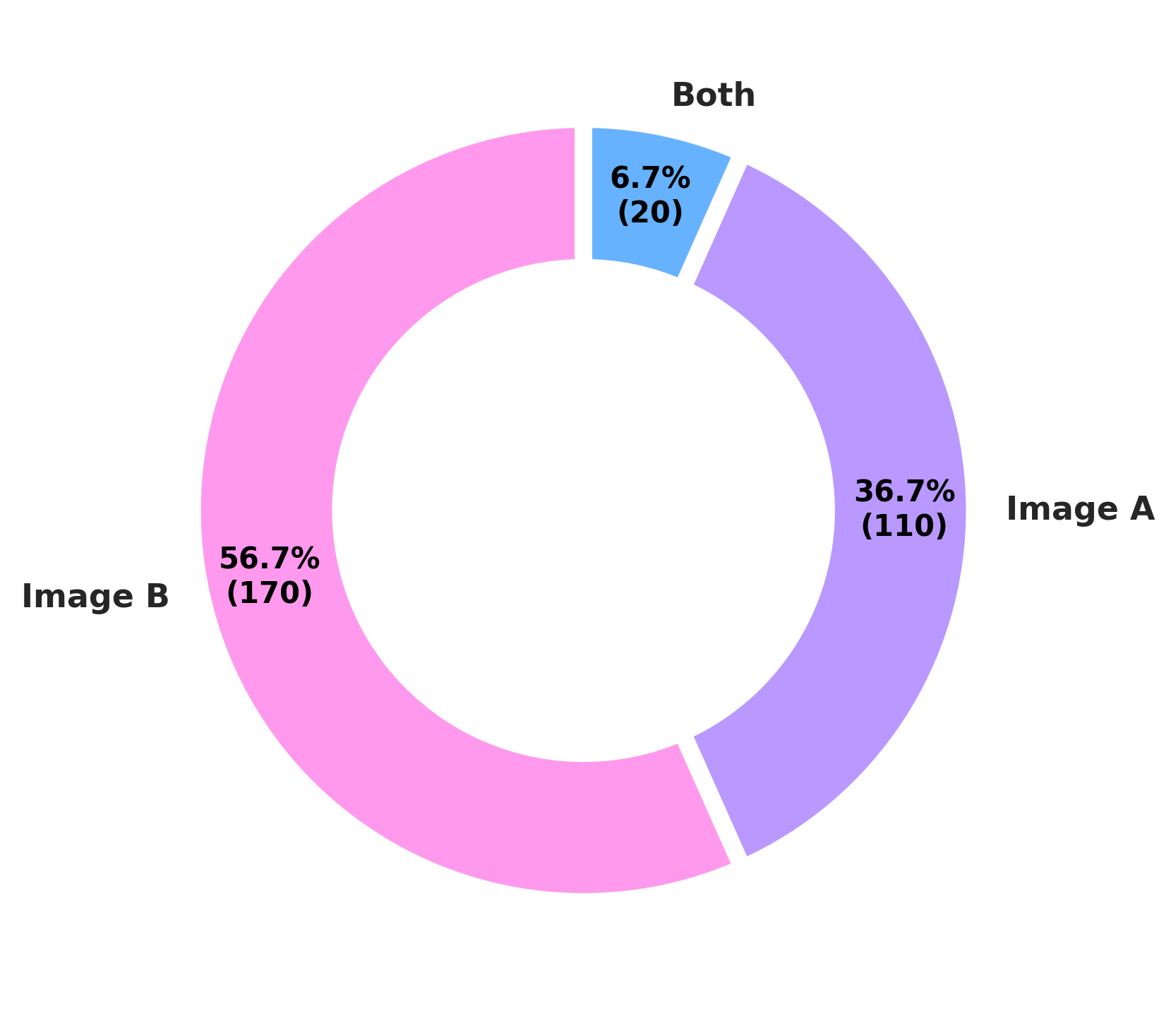}
    \end{minipage}%
        \hspace{0.01\textwidth}
    \begin{minipage}{0.3\textwidth}
        \centering
        \includegraphics[width=0.8\textwidth]{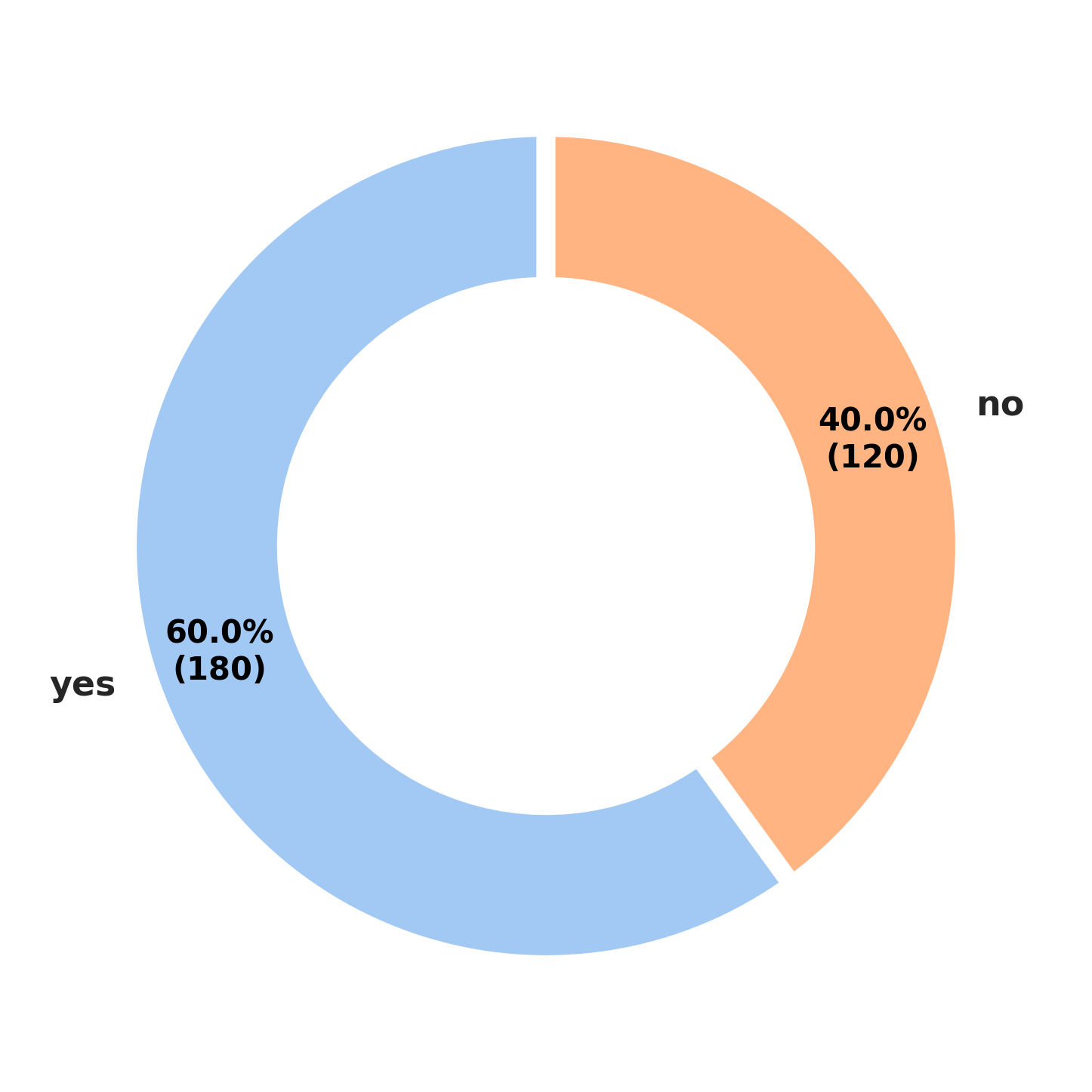}
    \end{minipage}
    \hspace{0.01\textwidth}
    \begin{minipage}{0.3\textwidth}
        \centering
        \includegraphics[width=0.8\textwidth]{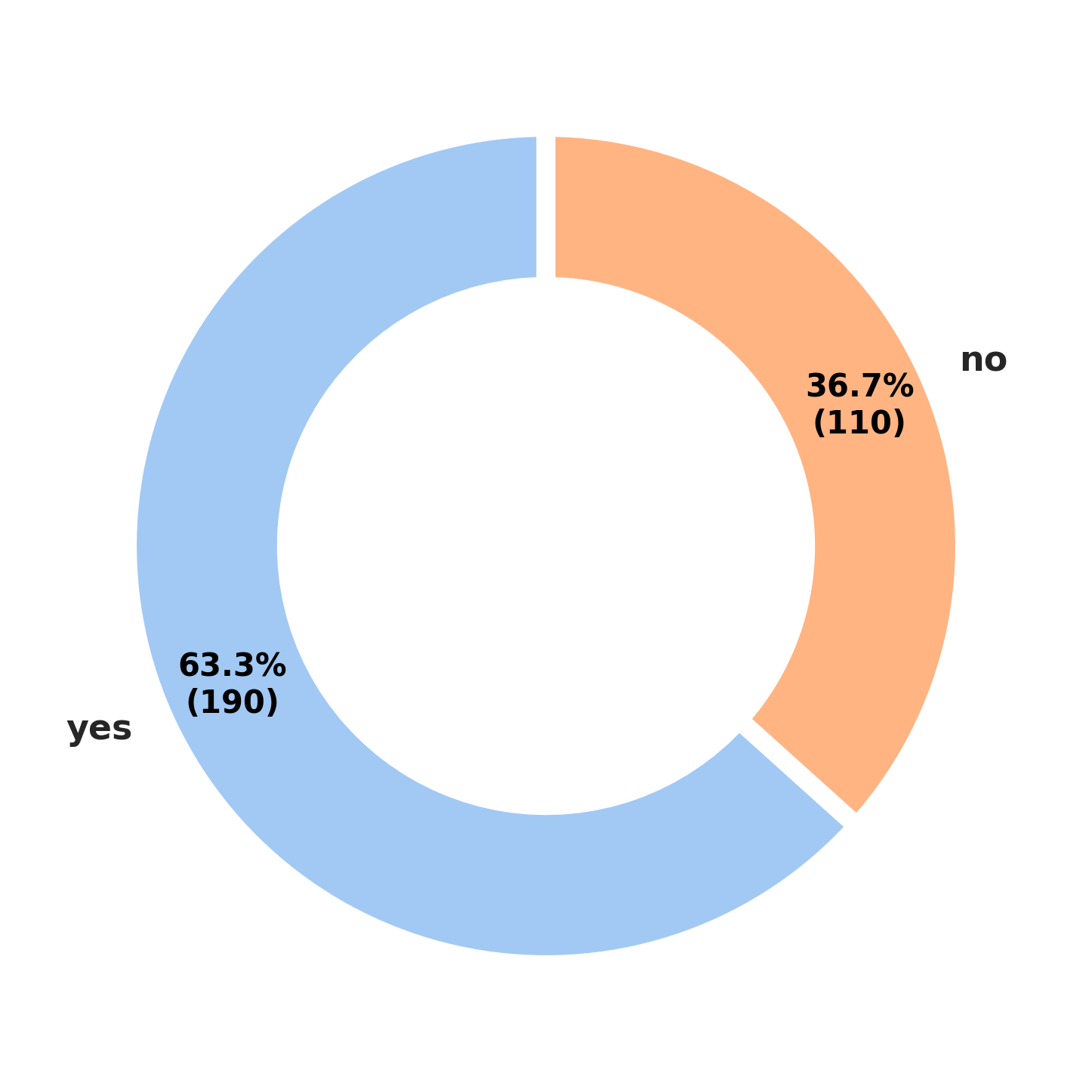}
    \end{minipage}%

    \caption{Results of LVLM-as-Judge. Top row for single step evaluation, second row for the analyze-then-judge procedure. The leftmost column corresponds to the \textbf{Select} experiment, the middle column represents the \textbf{Sem-Eq-Triplets} experiment, and the rightmost column corresponds to the \textbf{Sem-Eq-Pairs} experiment, all replicated for LVLM annotators.}
    \label{fig:lvlm-judge}
\end{figure}

\paragraph{Experimental setup.} The model used to perform the questionnaire judgment is ChatGPT 5.4 thinking, using the following default zero-shot prompt:
\begin{tcolorbox}[
    colback=gray!5,
    colframe=black!70,
    title={Annotation Prompt (single-step)},
    fonttitle=\bfseries,
    sharp corners,
    boxrule=0.6pt,
    left=6pt,
    right=6pt,
    top=6pt,
    bottom=6pt
] \small
You are going to evaluate a questionnaire regarding counterfactual explanations on images. Provide your annotations without relying on any prior knowledge about bird species, and simulate a reasoning process similar to that of an engineering student. Focus on salient semantic characteristics of the bird instances provided, which are likely to represent discriminative features of the bird class, and ignore superficial factors such as lighting, viewing angle, or bird positioning. Return your responses in the form of a dictionary: \texttt{\{question number: label\}}.
\end{tcolorbox}
For the analyze-then-judge experiment, the prompt is the following:
\begin{tcolorbox}[
    colback=gray!5,
    colframe=black!70,
    title={Annotation Prompt (two-step)},
    fonttitle=\bfseries,
    sharp corners,
    boxrule=0.6pt,
    left=6pt,
    right=6pt,
    top=6pt,
    bottom=6pt
] \small
Now, I want you to perform re-annotation on the questionnaire following a different reasoning process. You have first to analyze the image visually and as a second step, you have to decide the label upon your previous analysis. Once again, you are going to provide your annotations without reliance to any prior knowledge about bird species, while you have to simulate your way of thinking similar to an engineering student. You should rely on salient semantic characteristics of the bird instances provided (which most likely would represent discriminative features of the bird class) and ignore other superficial features such as lighting, angle, bird positioning etc. Return your responses in a dictionary: \texttt{\{question number: label\}}.
\end{tcolorbox}
\paragraph{Results of LVLM evaluation} 

are presented in Figure \ref{fig:lvlm-judge}, as an analogy to human evaluation. In all "Yes"/"No" experiments, the significantly higher "Yes" percentage indicates that the LVLM judges the retrieved counterfactual as semantically equivalent to the reference one in most evaluated cases. By also observing the leftmost column, the most frequent answer "B" reveals that selecting the GCN-retrieved instance is more preferable over the reference, reaching a consensus with human preferences (Figure \ref{fig:select-all}) and validating human-LVLM conceptual alignment. 

\begin{figure}[t!]
    \centering
    \begin{minipage}{0.3\textwidth}
        \centering
        \includegraphics[width=0.8\textwidth]{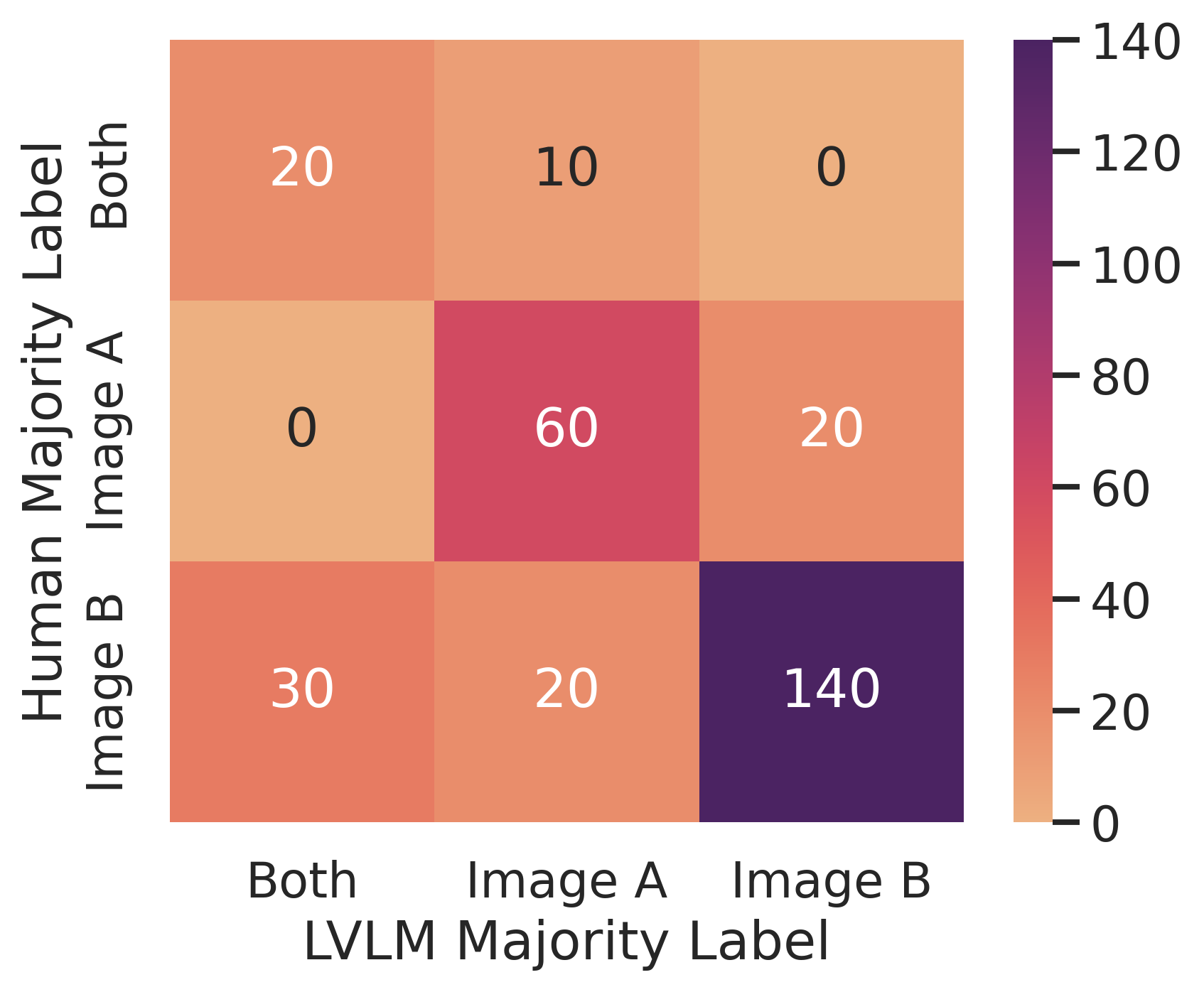}
        \caption*{\small Exact agreement: 0.7333 (73.33\%)
Cohen's kappa: 0.5285}
    \end{minipage}%
    \hspace{0.01\textwidth} 
    \begin{minipage}{0.3\textwidth}
        \centering
        \includegraphics[width=0.8\textwidth]{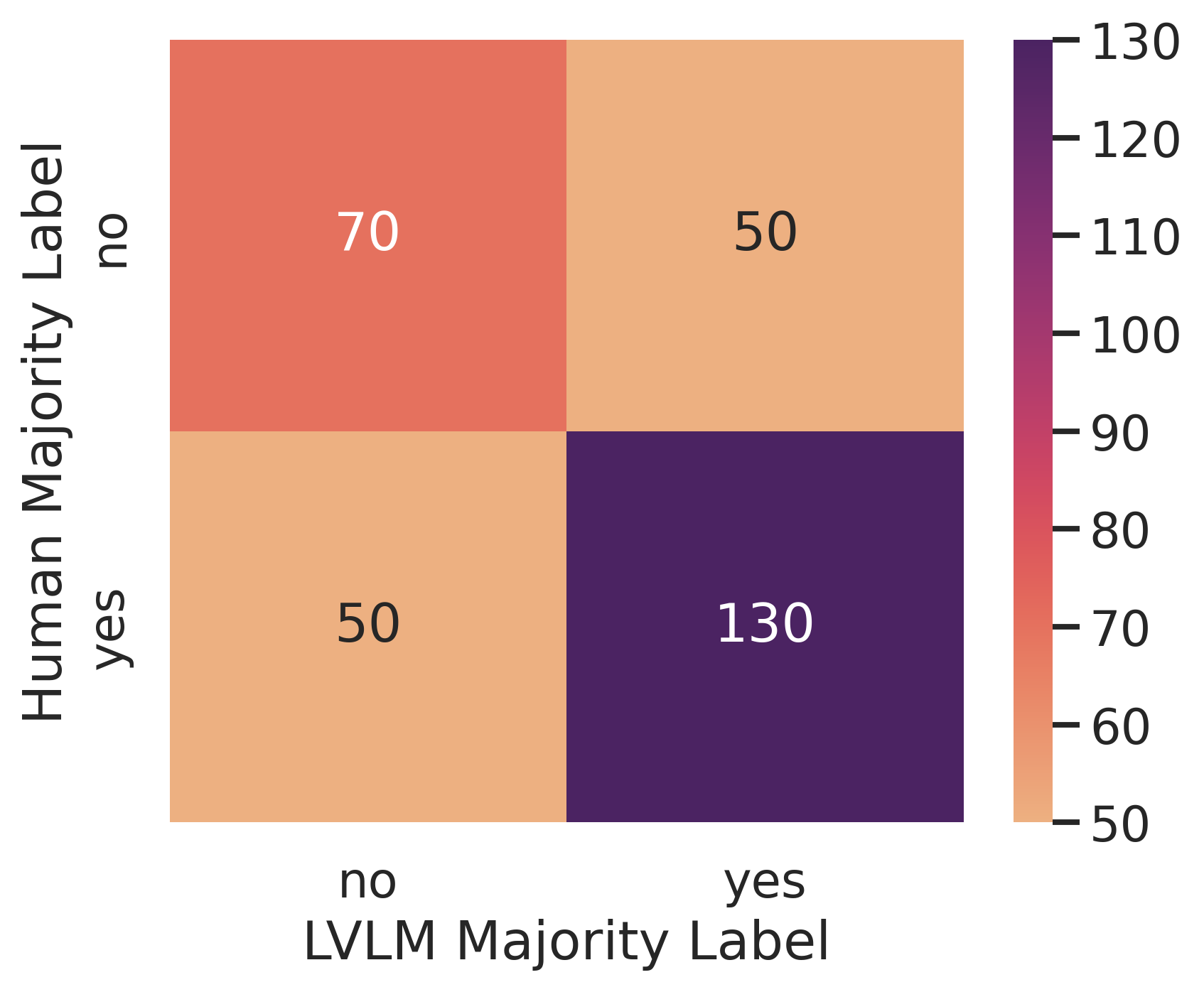}
        \caption*{\small Exact agreement: 0.7333 (73.33\%)
Cohen's kappa: 0.4258}
    \end{minipage}%
    \hspace{0.01\textwidth} 
    \begin{minipage}{0.3\textwidth}
        \centering
        \includegraphics[width=0.8\textwidth]{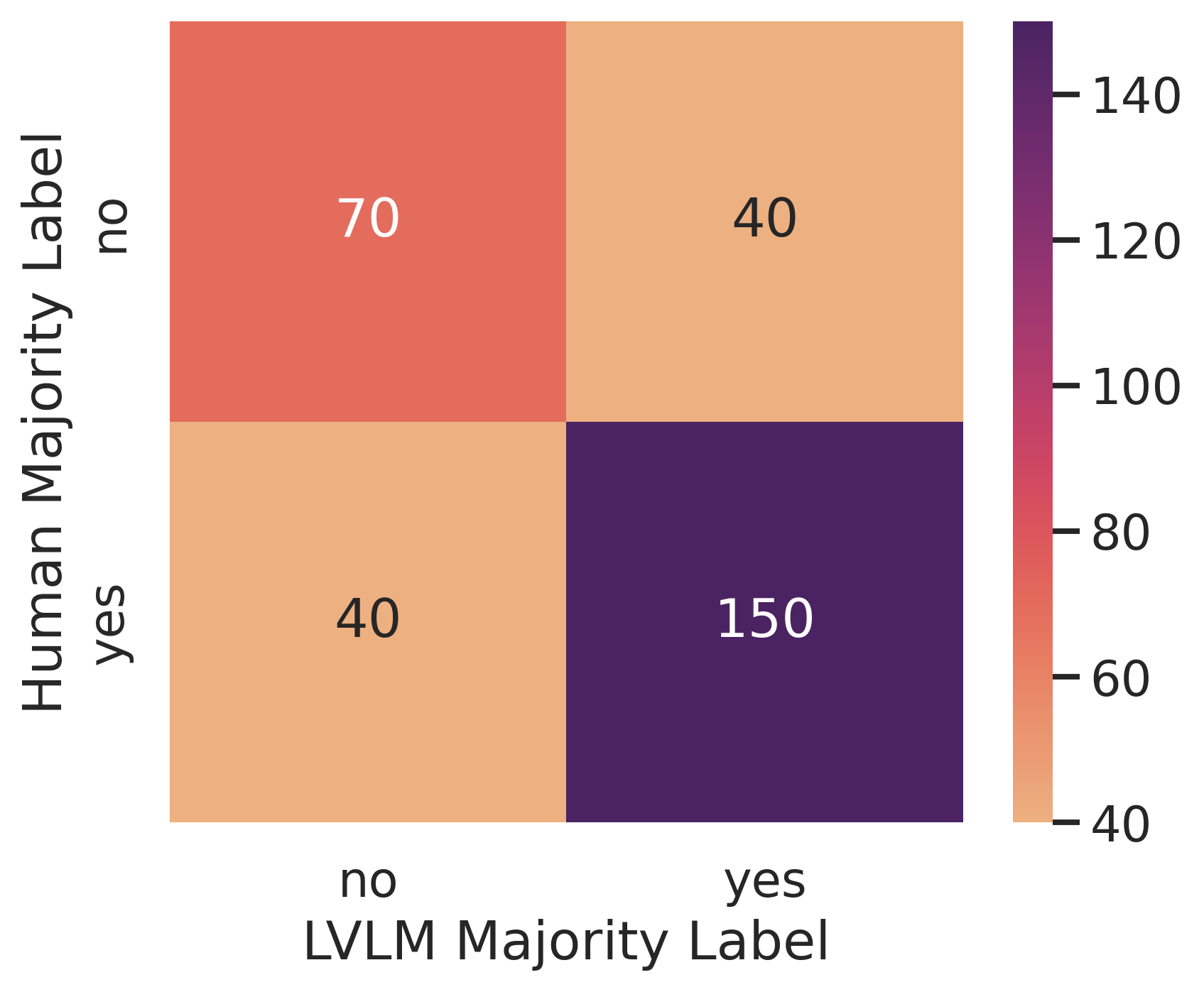}
        \caption*{\small Exact agreement: 0.6667 (66.67\%)
Cohen's kappa: 0.3056}
    \end{minipage}
    \vspace{0.01\textwidth} 

    \begin{minipage}{0.3\textwidth}
        \centering
    \includegraphics[width=0.8\textwidth]{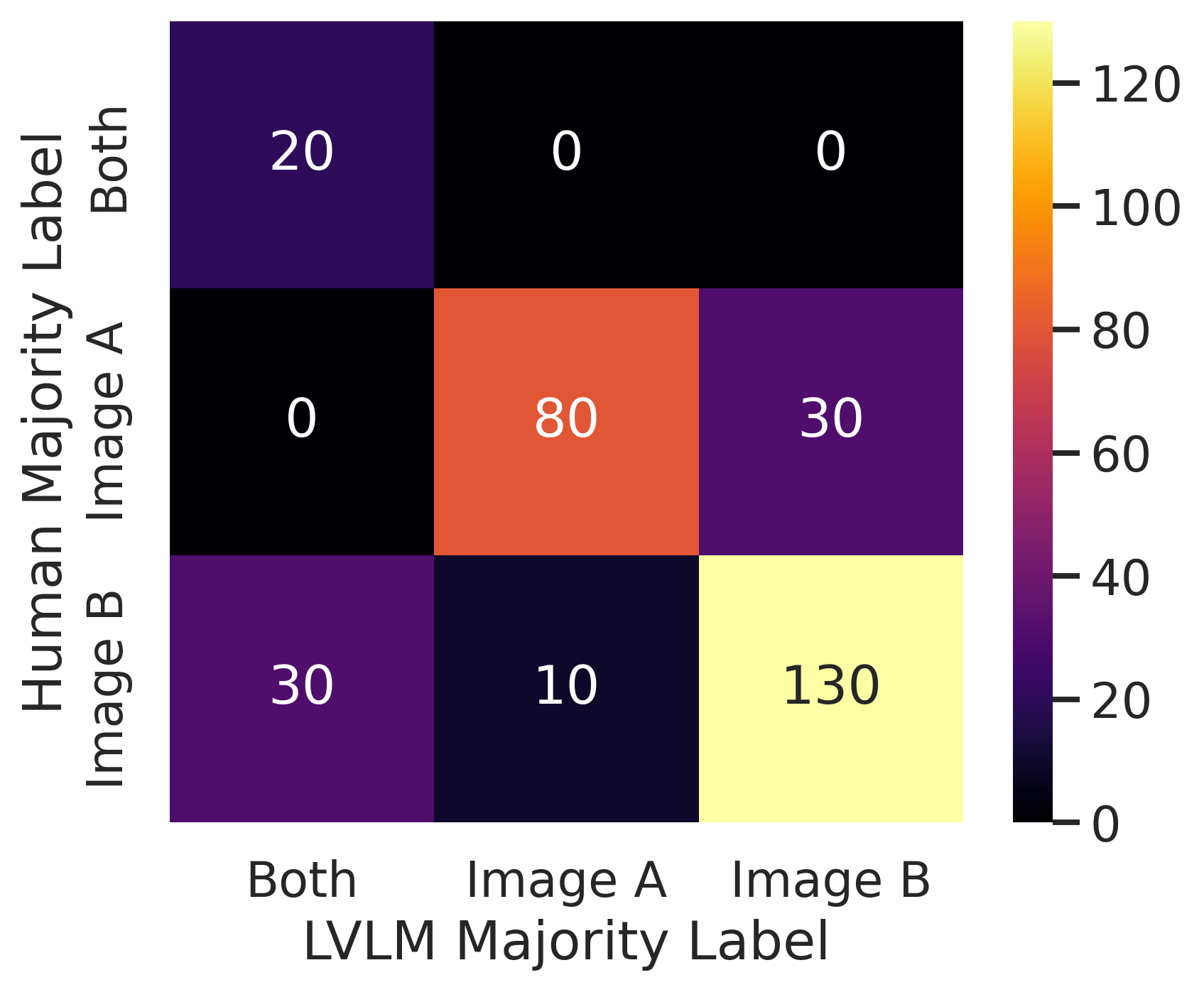}
    \caption*{\small Exact agreement: 0.7667 (76.67\%)
Cohen's kappa: 0.5954}
    \end{minipage}%
    \hspace{0.01\textwidth}
    \begin{minipage}{0.3\textwidth}
        \centering
        \includegraphics[width=0.8\textwidth]{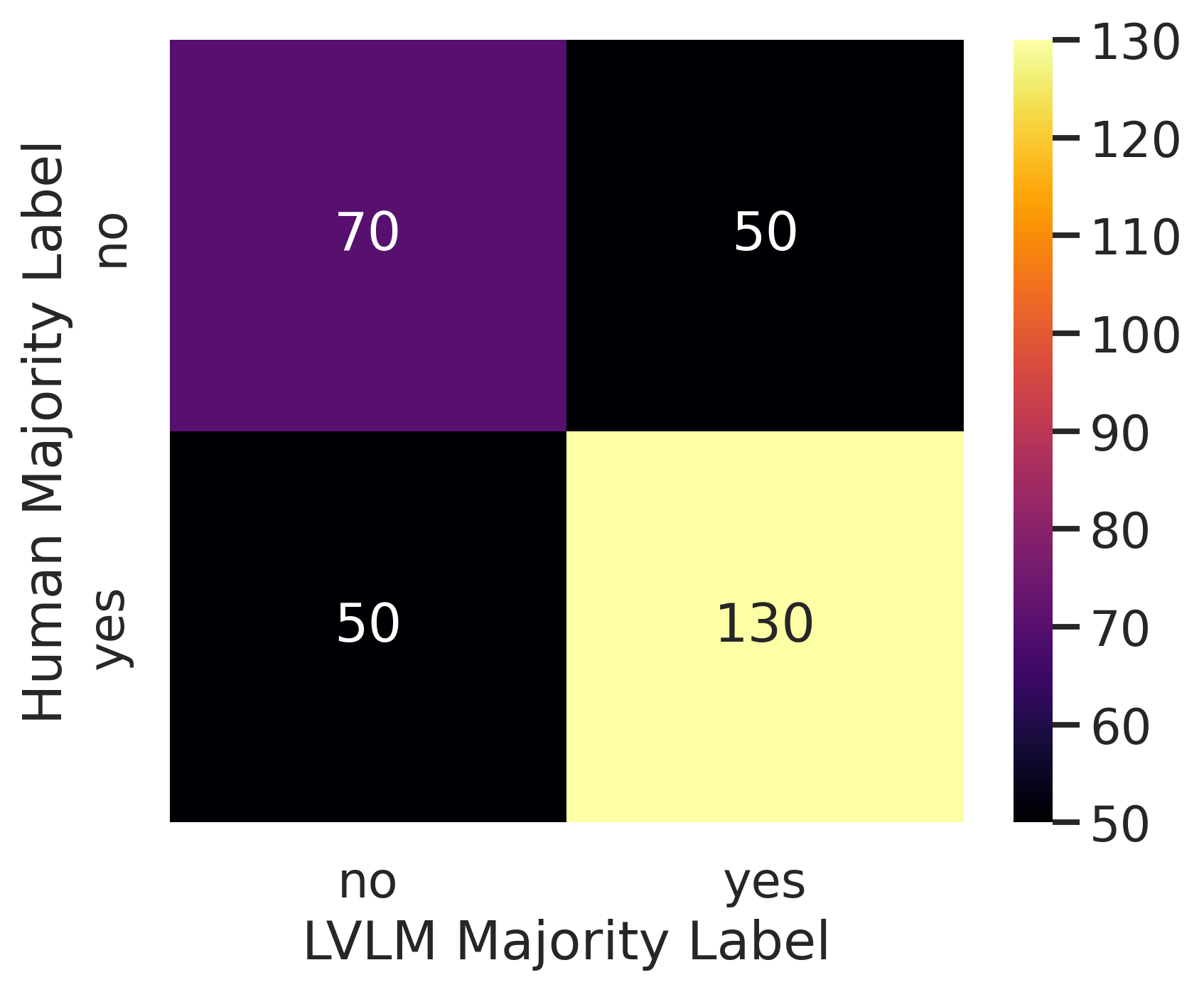}
        \caption*{\small Exact agreement: 0.7333 (73.33\%)
Cohen's kappa: 0.4258}
    \end{minipage}%
    \hspace{0.01\textwidth}
    \begin{minipage}{0.3\textwidth}
        \centering
        \includegraphics[width=0.8\textwidth]{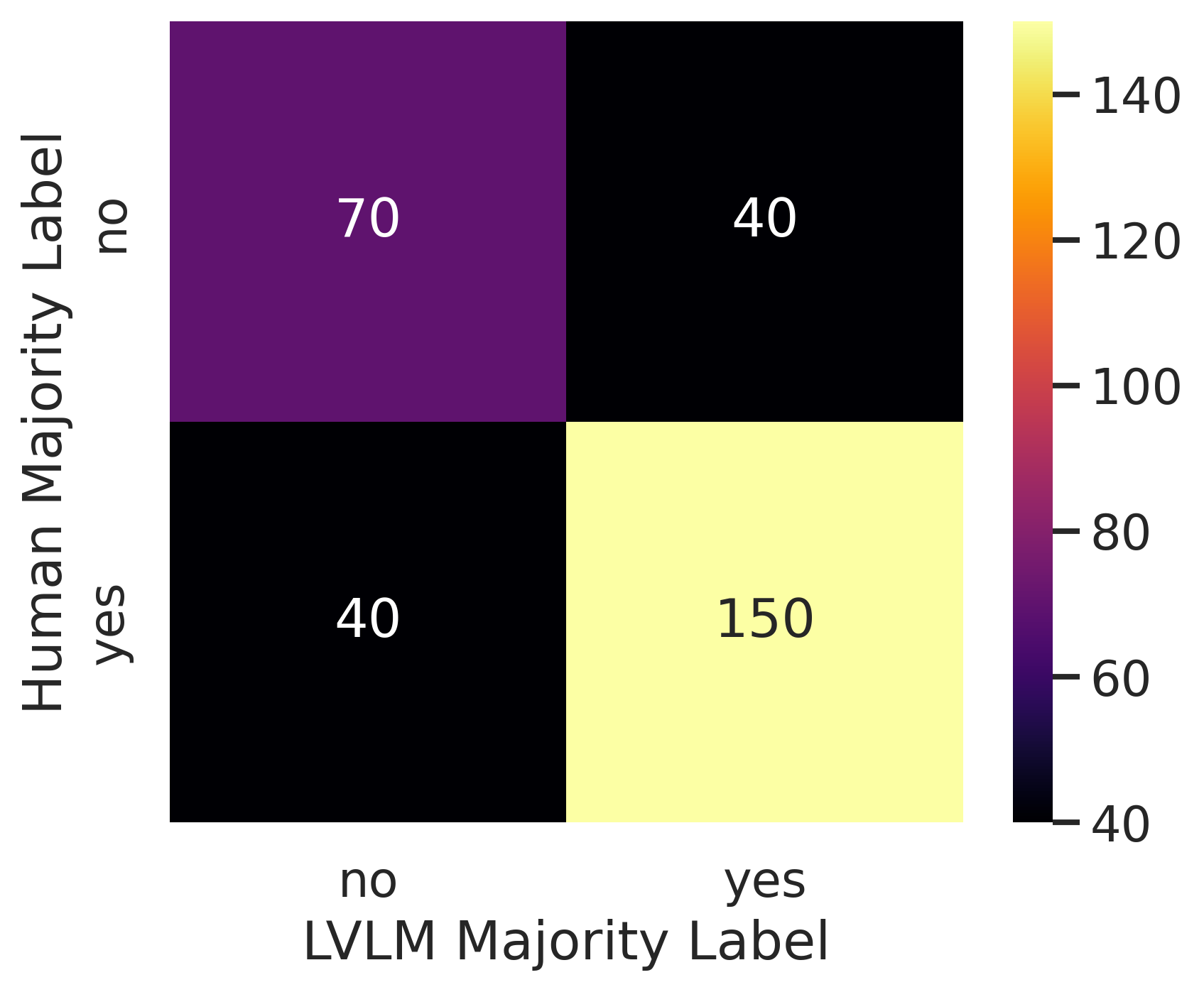}
        \caption*{\small Exact agreement: 0.6667 (66.67\%)
Cohen's kappa: 0.3056}
    \end{minipage}
    \caption{Human vs LVLM responses agreement confusion matrices directly corresponding to Fig. \ref{fig:lvlm-judge}.}
    \label{fig:human-vs-lvlm}
\end{figure}

\paragraph{Human-LVLM agreement} is presented in Figure \ref{fig:human-vs-lvlm}.
Across settings, human–LVLM agreement is consistently above chance yet moderate, with performance varying by format. Agreement is highest in the \textbf{Select} experiment, particularly with reasoning (76.67\% exact agreement, 
$\kappa$=0.5954), moderate in \textbf{Sem-Eq-Pairs} (73.33\%, $\kappa$=0.4258), and lowest in \textbf{Sem-Eq-Triplets} (66.67\%, 
$\kappa$=0.3056). Analyze-then-judge reasoning improves agreement only in the \textbf{Select} condition, suggesting that its benefit depends on task structure.

We attribute the observed discrepancies in LVLM responses to the varying cognitive demands and ambiguity of each task. The \textbf{Sem-Eq-Pairs} setting restricts the model to a binary contrast; while this simplicity facilitates higher raw agreement, the moderate $\kappa$ score reveals this alignment is largely a byproduct of the limited label distribution. By contrast, the \textbf{Sem-Eq-Triplets} setting expands the search space, requiring fine-grained discrimination that increases ambiguity and consequently lowers human–LVLM agreement. The \textbf{Select} setting yields the highest overall agreement and is the only scenario that materially benefits from explicit reasoning. Rather than forcing symmetric comparisons, it tasks the model with isolating the most prominent cue. This indicates that reasoning only helps when prioritizing distinct features, offering no advantage in environments that are overly constrained (Triplets) or highly ambiguous (Pairs). Ultimately, in our evaluation, the LVLM aligned closely with human judgment when isolating key discriminators, but faltered when coordinating multiple competing alternatives.
\section{Limitations}
\label{sec:limitations}
U-CECE is limited by the quality of the conceptual abstraction stage: errors in concept extraction, relation parsing, or graph construction can propagate to retrieval and explanation quality. Our experiments mitigate this by focusing on richly annotated datasets such as CUB and VG, but this likely overestimates robustness relative to noisier real-world settings.  The human study is restricted to a controlled bird domain and exhibits a noisy subset, while the LVLM-as-a-judge analysis shows only moderate human–LVLM agreement and is sensitive to task format. Although the U-CECE framework establishes conceptuality in counterfactual explanations and improves semantic faithfulness and structural minimality, broader issues remain.
Relatedly, U-CECE retrieves an existing instance from a different predicted class and extracts the conceptual differences between the two; it does not verify that applying those edits to the original input would itself flip the classifier. The retrieved edits therefore characterise the minimal conceptual difference between two decision regions rather than a validated pixel-level intervention, which would require a generative component as in \citep{spanos2025vcece}.
Finally, the critical distinction between merely \textit{explanatory} and truly \textit{causal} counterfactuals remains a chief issue, without which users might draw incorrect conclusions, leaving issues of real-world actionability, fairness, and robustness to abstraction noise for future work. 

\section{Future directions}


\paragraph{The human-machine explanatory gap}
Conceptual edits have been proven crucial for unraveling current explainability issues in different experimental setups, such as counterfactual image generation \citep{spanos2025vcece}. In this case, conceptualization acts as an indispensable tool for diagnosing the \textit{explanatory gap} between humans and visual classifiers. When explanations lack conceptual grounding, attempts to explain a black-box classifier in human-explainable terms become futile, as human and neural semantic levels widely differ. 
The recent shift towards diffusion-based counterfactual generation highlights this perfectly. Despite their unparalleled generative capacity, diffusion methods cannot bridge this gap on their own. The inability to accurately design salient, discriminative concepts stems not from the diffusion model's capabilities, but from the inherently misaligned signal provided by the classifier. Consequently, the more a classifier's semantics abstract away from human conceptualization, the more misaligned the guiding signal becomes, often causing diffusion models to hallucinate plausible but conceptually meaningless visual features. 
In essence, without a conceptually grounded signal, generating high-fidelity images yields limited interpretable substance. The work of \cite{spanos2025vcece} supports the claims presented in the current paper, setting new core desiderata in counterfactual image generation literature.

\paragraph{Robustness assessments of conceptual counterfactuals}
An important future avenue is the stress-test on the fragility of conceptual counterfactuals, attributing and quantifying sources of brittleness in the image itself, as well as during concept extraction, relational parsing, graph construction, and retrieval stages. 
Among these stages, concept extraction is particularly critical, as it directly dictates the quality of the final explanation. Notably, accurately extracting high-quality relations is substantially more difficult than isolating individual objects, as it requires capturing complex, often ambiguous visual interactions. Building on \cite{dimitriou2023structure}, future research should rigorously evaluate various relation extractors and end-to-end SGG pipelines to pinpoint failure modes triggered by poor, underspecified or incomplete data.
Specifically, future experimentation can focus on concept 'survival' across perturbation intensity and type and the discovery of bifurcation points where small input or concept changes lead to qualitatively different counterfactual explanations. Related findings will establish such analysis as a measure of ambiguity margins in derived explanations, while also exposing concept importance in human terms. At the same time, accompanying curation strategies will enhance the meaningfulness, trust and causal validity of conceptual counterfactual explanations.

\paragraph{Conceptual Uncertainty Estimation and Mitigation} Ambiguity analysis inspires the exploration of a wider research direction: the uncertainty estimation of conceptual counterfactuals, particularly when semantic evidence is noisy, weakly specified, or inherently ambiguous. Targeted stress tests are able to reveal whether uncertainty stems from annotation noise, class overlap, or genuine aleatoric ambiguity, offering discrete, semantic-aware measures on the data themselves. Since such conceptual uncertainty can be well-measured and calibrated, any ambiguity of consequent explanation systems can be better isolated and gauged, while also mitigation endeavors can be framed more accurately, allowing for ambiguity-aware retrieval, conceptual enrichment and data-specific curation. Ultimately, this would position conceptual counterfactuals not only as explanatory artifacts, but also as tools for uncertainty estimation and reduction.

\section{Conclusion}
\label{sec:conclusion}
In this paper, we introduced U-CECE, a unified multi-resolution framework for conceptual counterfactual explanations that bridges atomic, relational, and structural concept representations within a single pipeline. More broadly, our findings reinforce the value of conceptual counterfactuals as a human-centered alternative to low-level counterfactual explanations: by operating on semantically meaningful concepts rather than pixel-level edits, they produce explanations that are better aligned with how humans perceive, compare, and reason about visual changes. Within this perspective, the three expressivity levels of U-CECE make explicit a central trade-off in conceptual explainability. The atomic level provides fast and lightweight explanations, the relational level captures simple interactions at moderate cost, and the structural level delivers the most faithful representation of complex scenes by preserving full relational topology, albeit with higher computational demands. Our experiments on CUB and Visual Genome show that no single level is universally optimal; instead, explanatory structure should be selected according to the relational complexity of the data and the practical constraints of the application. At the highest expressivity tier, this trade-off is further refined through the complementary transductive and inductive modes: the former offers high-precision retrieval when sparse data and strict structural minimality are paramount, while the latter supports scalable and flexible deployment when data are abundant and computational constraints favor efficient generalization. Human evaluation on CUB further demonstrates that the retrieved structural counterfactuals are not only semantically robust, but often preferred over GED-based references, highlighting an important distinction between deterministic minimality and perceived explanatory quality. While GED is rightfully established as a foundational metric for the evaluation and optimization of conceptual counterfactuals, this divergence reveals its limitations. It underscores that human-in-the-loop validation remains indispensable, as the true suitability and cognitive resonance of an explanation is notoriously difficult to capture through purely quantitative metrics. The LVLM-as-a-judge results complement this picture by largely reproducing the same preference and equivalence trends, while also showing that human-LVLM alignment depends on task structure and is strongest when the task emphasizes salient semantic cue prioritization. Taken together, these findings position U-CECE not simply as a retrieval framework, but as a principled foundation for studying how semantic abstraction, relational structure, and computational efficiency jointly shape the quality of counterfactual explanations. Overall, U-CECE offers a practical, extensible, and human-aligned foundation for conceptual counterfactual explanation, one that supports flexible deployment today while opening a broader path toward richer, more faithful, and more cognitively grounded explainability in the future.

\subsubsection*{Acknowledgments}{This work was supported by the Hellenic Foundation for Research and Innovation (HFRI)  under the 5th Call for HFRI PhD Fellowships (Fellowship Number 19268).}


\bibliography{main}
\bibliographystyle{tmlr}

\appendix






\section{Experimental Details}
\label{app:train}
For the supervised GNN models, we utilize a single-layer architecture with a dimension of 2048, optimized using Adam (no weight decay or dropout) with a batch size of 32 over 50 epochs. The learning rate is set to 0.04 for GCN and 0.02 for GAT and GIN; GAT specifically employs 8 attention heads, while the $\epsilon$ parameter for GIN is non-learnable. In the unsupervised setting, training is performed in 30 epochs.
For the VG dataset, a learning rate of $10^{-4}$ is maintained for both stages. For the CUB dataset, we utilize a learning rate of $10^{-3}$.
The number of training pairs, denoted as $p$, is set following the heuristic $p \approx N/2$ from the SGCE paper, resulting in 70K and 50K pairs for datasets with 500 and 422 graphs, respectively. 
Finally, all experiments were executed on a single NVIDIA T4 GPU, ensuring a consistent and accessible computational baseline.

\section{Complementary Experimental Findings on Expressivity Modes
}
\label{app:expressivity-more}
\begin{table}[ht!]
\centering
\small
\renewcommand{\arraystretch}{1.4}
\caption{Retrieval Metrics ($k=2, 4$) across Expressivity Tiers. $P_{bin}$ denotes binary precision.
}
\label{tab:appendix-retrieval-metrics}
\begin{tabular}{llcccccc}
\toprule
\textbf{Dataset} & \textbf{Level} & \textbf{P@2} $\uparrow$ & \textbf{P@4} $\uparrow$ & \textbf{nDCG@2} $\uparrow$ & \textbf{nDCG@4} $\uparrow$ & \textbf{$P_{bin}$@2} $\uparrow$ & \textbf{$P_{bin}$@4} $\uparrow$ \\
\midrule
\multirow{3}{*}{\textbf{CUB}} 
    & Atomic     & \textbf{0.241} & \textbf{0.330} & \textbf{0.276} & \textbf{0.357} & \textbf{0.325} & \textbf{0.506} \\
    & Relational & 0.197 & 0.294 & 0.248 & 0.332 & 0.238 & 0.388 \\
    & Structural & 0.200 & 0.306 & 0.239 & 0.323 & 0.256 & 0.450 \\
\midrule
\multirow{3}{*}{\shortstack{\textbf{VG} \\ \textbf{DENSE}}} 
    & Atomic     & 0.278 & 0.306 & 0.297 & 0.375 & 0.338 & 0.418 \\
    & Relational & 0.208 & 0.261 & 0.262 & 0.344 & 0.242 & 0.318 \\
    & Structural & \textbf{0.292} & \textbf{0.358} & \textbf{0.333} & \textbf{0.407} & \textbf{0.370} & \textbf{0.492} \\
\midrule
\multirow{3}{*}{\shortstack{\textbf{VG} \\ \textbf{RANDOM}}} 
    & Atomic     & \textbf{0.256} & \textbf{0.302} & \textbf{0.293} & \textbf{0.372} & 0.296 & 0.388 \\
    & Relational & 0.099 & 0.143 & 0.201 & 0.290 & 0.110 & 0.182 \\
    & Structural & 0.246 & 0.295 & 0.290 & 0.369 & \textbf{0.300} & \textbf{0.424} \\
\bottomrule
\end{tabular}
\end{table}

As shown in Table \ref{tab:appendix-retrieval-metrics}, retrieval metrics increase significantly at higher $k$ values. Notably, the binary precision for locating the reference counterfactual within the top 2 or 4 samples approaches $50\%$, doubling the maximum $P@1$ ($\sim25\%$). Trends remain consistent across all datasets and methods.

\section{Human Survey}
\label{app:human}
\subsection{Details on Participants}

Prior to participation, informed consent was obtained from individuals to utilize their anonymized responses for research purposes. To preserve privacy, fine-grained demographic data was not recorded. However, the survey was distributed to a group of undergraduate and graduate students - age range 23 - 35, of both sexes. All participants share an educational background in software engineering. Crucially, none of the participants possessed prior domain expertise in ornithology. This lack of specialized knowledge was intentional; it ensures that their ability to evaluate the fine-grained bird counterfactuals relied entirely on the structural and visual explanations provided by the framework, rather than any pre-existing subject familiarity.

\subsection{Form Design and Instructions}

The human perception study was conducted using the Google Forms platform. This choice of medium allowed participants to leverage standard browser utilities, such as zooming in on high-resolution images, to carefully inspect the fine-grained visual characteristics of the birds before making their assessments. The study was divided into three distinct experimental setups, aligning with the analysis on the main paper: \textbf{Select}, \textbf{Sem-Eq-Triplets}, and \textbf{Sem-Eq-Pairs}. Each form began with a specific set of instructions to align the participants' focus purely on semantic equivalence (species and anatomical traits) rather than environmental noise.

"Triplets-Select" Initial Instruction:
    \textit{"The first image is the source, while the second and the third are proposed counterfactual images of the source. Focus on the species and the characteristics of the counterfactuals (Image A and Image B) and ignore other image details, bird pose or position. Which of the two counterfactuals, Image A or Image B, is the most appropriate to the source? If both are equally good counterfactuals, answer 'Both'."} 

For each instance, participants answered the question: \textbf{"Which of the two counterfactuals, Image A or Image B, is the most appropriate to the source?"} using a multiple-choice format (Image A, Image B, Both).

\begin{wrapfigure}{r}
{0.45\textwidth}
  \vskip -0.2in \centering
\includegraphics[width=\linewidth]{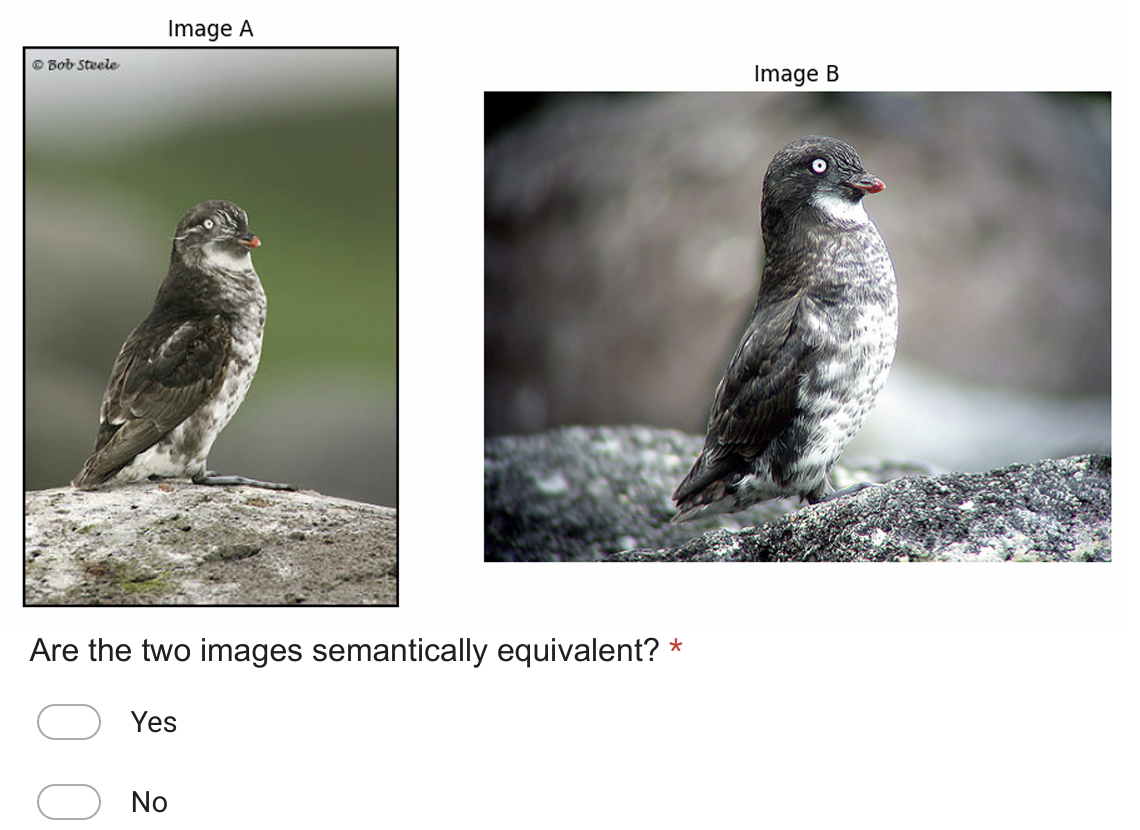} 
    \caption{An example of the \textbf{Pairs} experimental layout. The Source Image is removed to test unanchored semantic equivalence.}
    \label{fig:app_pairs_example}
    \vskip -0.2in
\end{wrapfigure}

"Sem-Eq-Triplets" Initial Instruction:
    \textit{"The first image is the source, while the second and the third are proposed counterfactual images of the source. Are the proposed counterfactuals (second and third image) semantically equivalent? Focus on whether the counterfactuals are of the same species and have similar characteristics."}

Participants were asked to answer \textbf{"Are the proposed counterfactuals (Image A and Image B) semantically equivalent?"} with a simple \textit{Yes} or \textit{No}.

"Sem-Eq-Pairs" Initial Instruction:
    \textit{"Given two images, assess whether they are semantically equivalent by concentrating exclusively on the species and characteristics of the birds depicted. Ignore any other elements, such as the surroundings, pose, or position of the birds."}

Participants answered the  question: \textbf{"Are the two images semantically equivalent?"} (\textit{Yes} / \textit{No}).

\subsection{Noise Results}
\label{app:human-noise}

\begin{wrapfigure}{r}{0.38\textwidth}
    \centering
\vskip -0.2in
    \includegraphics[width=0.9\linewidth]{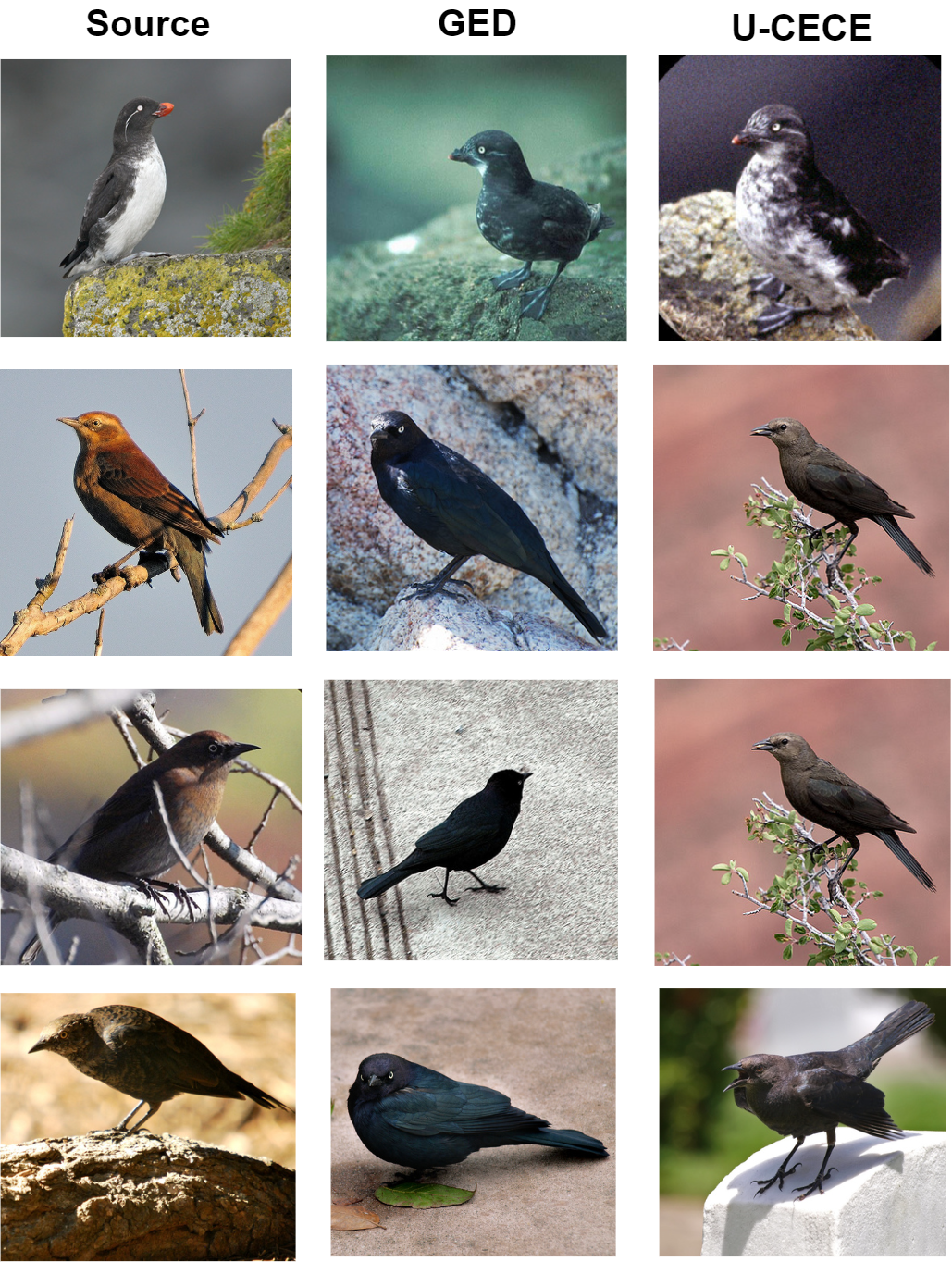}
    
    \caption{Examples of strong human preference of U-CECE over reference GED counterfactuals.} 
    \label{fig:human_preference}
    \vskip -0.3in
\end{wrapfigure}

Analysis of the human evaluation results for the Noise dataset across all experimental tiers reveals a lack of consensus, with response distributions approximating a random choice baseline. In the Select experiment, participants displayed near-equal preference between Image A (34.0\%), Image B (32.7\%), and the "Both" option (33.2\%), mirroring the $1/3$ probability of a random selector. This trend persists in the binary tiers, where "Yes" votes reached only 50.6\% for Pairs and 52.6\% for Triplets, effectively hovering at the $50\%$ chance threshold. These data validate our categorization of these instances as semantically ambiguous "noise," confirming that for these specific samples, the neural-based U-CECE-Structural retrievals and the reference GED counterfactuals are perceived as equally plausible (or equally suboptimal) from a human perspective.

\subsection{Human Preference of U-CECE over GED counterfactuals}
\label{app:human-ged-discrepancy}

The quantitative evaluation via GED and the human surveys are complementary by construction, and therefore in this section we aim to analyze when the two do not coincide.
To characterize where they come apart, we analyzed all queries where annotators preferred U-CECE over the reference counterfactual at least 5x over. For each instance, we recorded the rank of U-CECE within the GED-ordered candidate pool and the corresponding marginal edit-cost penalty. The findings indicate that the disparity is generally minimal: our proposed explanation typically ranks within the top decile of the candidate pool, incurring an edit-cost penalty small enough to be comparable to the approximation error inherent in the GED calculation itself. This behavior is consistent with the structural characteristics of the CUB dataset; the graphs predominantly consist of near-uniform stars exhibiting similar size and density. Consequently, structural costs remain relatively static across candidates, and the identification of the minimizer depends primarily on minor variations in attribute substitution penalties. In this regime, the GED objective is largely indifferent among several top candidates, and human preference acts as a tie-breaking mechanism. Beyond this majority, a smaller, long-tail distribution of cases exhibits more pronounced divergence, highlighting specific limitations within the cost model.

The four cases illustrated in Figure \ref{fig:human_preference} were selected to represent this spectrum of divergence. The three Blackbird examples (bottom 3 rows, ranks 2-5) represent near-ties. In these instances, our retrieved counterfactual and the reference counterfactual exhibit small differences in edit cost; adjudicating human preference between two similar counterfactuals exceeds the discriminative capacity of the GED formulation. The Auklet case (row 1) illustrates a significant divergence: our CE ranks lower under the GED metric yet is strongly preferred by human annotators. 
This specific instance highlights a critical assumption inherent in the cost model. 
Because color and pattern substitution costs are derived from a lexical hierarchy rather than a perceptual space, pronounced tonal discrepancies that are immediately obvious to human observers are not adequately penalized. Despite this, even this worst-case divergence remains within the top quartile of the GED ordering. Consequently, this divergence establishes the limitations of the reference objective as a proxy for human preference without invalidating its utility as a retrieval target.

\end{document}